\documentclass{nature}

\usepackage{booktabs}
\usepackage{multirow}
\usepackage{graphicx}  
\usepackage{makecell}
\usepackage{xurl}
\usepackage{hyperref}
\hypersetup{hidelinks}

\usepackage{lineno}

\usepackage{setspace}

\usepackage{amsfonts}
\usepackage{amssymb}
\usepackage{amsmath}

\usepackage{float}
\usepackage{placeins}
\usepackage{graphicx}

\usepackage[export]{adjustbox}

\usepackage[normal,normal,bf]{caption}

\newcounter{mybodyfigure}
\newcounter{myedfigure}

\newcommand{\stepbodyfigure}{\refstepcounter{mybodyfigure}}

\makeatletter
\g@addto@macro\caption@prepareslc{%
  \renewcommand{\stepbodyfigure}{\caption@l@stepcounter{mybodyfigure}}}
\makeatother

\newcommand{\stepedfigure}{\refstepcounter{myedfigure}}

\makeatletter
\g@addto@macro\caption@prepareslc{%
  \renewcommand{\stepedfigure}{\caption@l@stepcounter{myedfigure}}}
\makeatother

\usepackage[super,comma,sort&compress]{natbib}

\usepackage{hyperref}
\usepackage{verbatim}

\usepackage[normalem]{ulem}
\usepackage{rotating} 
\usepackage{soul}

\usepackage{color} 

\usepackage[colorinlistoftodos]{todonotes}

\makeatletter
\newsavebox\myboxA
\newsavebox\myboxB
\newlength\mylenA

\newcommand*\xoverline[2][0.75]{%
    \sbox{\myboxA}{$\m@th#2$}%
    \setbox\myboxB\null
    \ht\myboxB=\ht\myboxA%
    \dp\myboxB=\dp\myboxA%
    \wd\myboxB=#1\wd\myboxA
    \sbox\myboxB{$\m@th\overline{\copy\myboxB}$}
    \setlength\mylenA{\the\wd\myboxA}
    \addtolength\mylenA{-\the\wd\myboxB}%
    \ifdim\wd\myboxB<\wd\myboxA%
       \rlap{\hskip 0.5\mylenA\usebox\myboxB}{\usebox\myboxA}%
    \else
        \hskip -0.5\mylenA\rlap{\usebox\myboxA}{\hskip 0.5\mylenA\usebox\myboxB}%
    \fi}
\makeatother

\title{ALICE: Learning a General-Purpose Pathology Foundation Model from Vision, Vision-Language, and Slide-Level Experts} 

\author{Jiawen Li$^{1*}$, Tian Guan$^{1*}$, Huijuan Shi$^{3*}$, Xitong Ling$^{1}$, Mingxi Fu$^{1}$, Anjia Han$^{3+}$, Chao He$^{2+}$, Yonghong He$^{1,4+}$}

\begin{document}

\maketitle

\begin{affiliations}
 \item Institute of Biopharmaceutical and Health Engineering, Tsinghua Shenzhen International Graduate School, Tsinghua University, Shenzhen, China
 \item Department of Engineering Science, University of Oxford, Oxford, UK
 \item Department of Pathology, The First Affiliated Hospital of Sun Yat-sen University, Guangzhou, China
 \item Medical Optical Technology R\&D Center, Research Institute of Tsinghua, Pearl River Delta, Guangzhou, China
 \item[] $^{*}$Contributed equally
 \item[] 
 \item[] $^{+}$\textbf{Corresponding Authors:} \\
 Anjia Han (hananjia@mail.sysu.edu.cn), Chao He (chao.he@eng.ox.ac.uk) \\
 Yonghong He (heyh@sz.tsinghua.edu.cn)
 
\end{affiliations}

\begin{abstract}

\textit{\large Abstract} \\

Foundation models are reshaping computational pathology, yet their capabilities remain shaped by pretraining objectives, data sources, and spatial scales, fragmenting complementary expertise across separate backbones. Here we present ALICE, a unified foundation model trained through multi-stage agglomerative distillation that sequentially distills eight vision-only, vision–language, and slide-level teacher models into dedicated modules of a single backbone. ALICE is pretrained on 24,985,184 tile-level pathology images and 155,604 high-resolution images, and evaluated across 21 task scenarios, 96 downstream tasks, and 48 data sources, spanning region-of-interest tissue analysis, vision–language multimodal evaluation, and whole-slide clinical assessment. In all three evaluation settings, ALICE achieved the best average rank among task-matched pathology foundation models. These results demonstrate that agglomerative distillation can consolidate complementary capabilities from specialized models into a unified backbone for broad computational pathology applications. The model is available at https://github.com/WonderLandxD/ALICE.

\end{abstract}

\newpage


\section{Introduction}

Histopathological assessment of tissue remains the cornerstone of clinical oncology, forming the basis for diagnostic, prognostic, and treatment decisions across many cancer types \cite{song2023artificial}. Digitizing slides into whole-slide images (WSIs) has created unprecedented opportunities to develop computational tools that enhance pathology analysis at scale \cite{zheng2026lazyslide,lu2021data}. Prior deep learning methods have shown potential for specific tasks such as tumor detection, molecular biomarker prediction, and survival analysis \cite{jiang2024transformer,campanella2019clinical,tsai2023histopathology,zhu2023accurate}. However, most of these models are trained from scratch on task-specific, typically small labeled cohorts, limiting their performance and generalization capabilities. Large-scale foundation models pretrained on massive histopathology datasets through self-supervised or multimodal learning have fundamentally changed this paradigm, providing transferable vision representations adaptable to a wide range of downstream clinical tasks with substantially reduced annotation requirements \cite{chen2024towards,lu2024visual,yan2025pathorchestra,xu2024whole}. Consequently, a growing number of pathology foundation models (PFMs) have been developed, demonstrating substantial potential across diverse clinical applications.

Despite their common goal of learning general-purpose representations, existing PFMs are developed under different pretraining paradigms, leading to distinct but fragmented capabilities \cite{neidlinger2025benchmarking,tizhoosh2026rethinking}. For instance, vision-only models, such as UNI \cite{chen2024towards} and Virchow \cite{vorontsov2024foundation}, rely on self-supervised learning to capture dense morphological representations, but they lack explicit alignment with pathological concepts and language-level semantics. Vision-language models, such as CONCH \cite{lu2024visual} and MUSK \cite{xiang2025vision}, rely on multimodal contrastive or generative learning to align visual features with textual semantics, but they underperform on fine-grained visual discrimination tasks \cite{campanella2025clinical}. Slide-level models, such as GigaPath \cite{xu2024whole} and TITAN \cite{ding2025multimodal}, rely on whole-slide context or weak clinical signals to capture global tissue information, but their practical transferability remains constrained, as they can be difficult to fine-tune or insufficiently expressive for broad downstream use. Therefore, no existing model consistently covers the full spectrum of morphological, semantic, and slide-level capabilities required across pathology tasks.

A natural way to address this fragmentation is to use a unified architecture to integrate these capabilities. In the general computer vision community, agglomerative models \cite{heinrich2025radiov2,zhu2026efficient} have shown that experts trained with distinct objectives can be distilled into a unified backbone, treating pretrained models as complementary knowledge sources rather than isolated alternatives. In computational pathology, GPFM \cite{ma2026generalizable} has provided a first step by showing that unified knowledge distillation can improve the generalization of PFMs across diverse clinical tasks. However, pathology expertise spans local morphology, language-aligned diagnostic concepts, and whole-slide clinical context. Existing distillation frameworks primarily strengthen general visual representations but do not explicitly organize these heterogeneous sources of expertise by modality, scale, and level of clinical abstraction. Thus, to build a truly general-purpose PFM, a more structured aggregation strategy is required to absorb complementary expert knowledge step by step into a single backbone.

\begin{figure*}[htbp]
    \centering
    \includegraphics[width=0.99\linewidth]{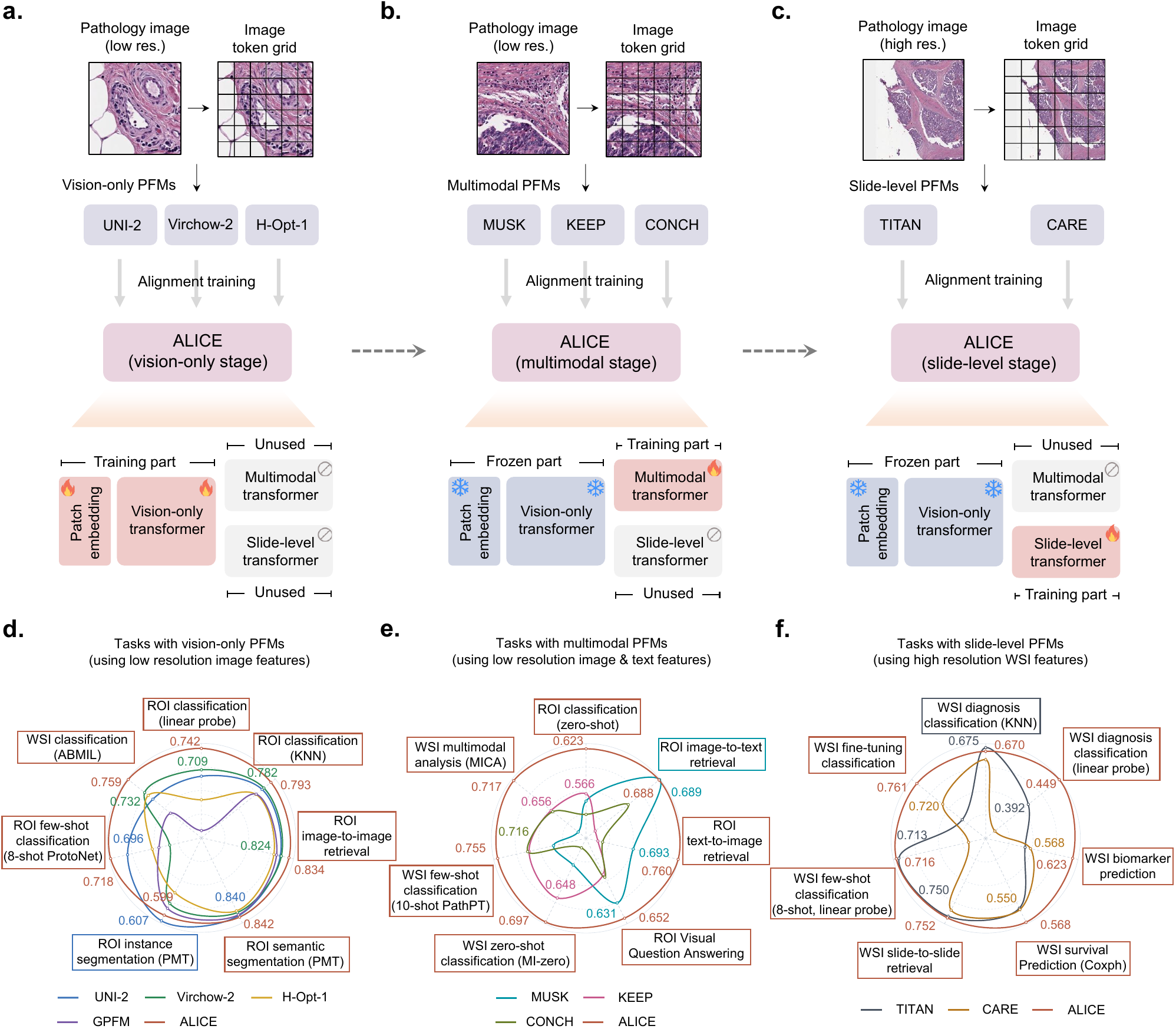}
    \caption{\textbf{Overview of ALICE training and evaluation.} \textbf{a-c,} Multi-stage agglomerative distillation framework for ALICE. \textbf{a,} In the vision-only stage, ALICE learns morphology-oriented representations from pathology tile images by aligning with three vision-only PFMs, UNI-2, Virchow-2, and H-Opt-1. The patch embedding module and vision-only transformer are trained, whereas the multimodal and slide-level transformers are inactive. \textbf{b,} In the multimodal stage, the pre-trained visual encoder is frozen, and ALICE learns language-aligned representations from three multimodal PFMs, MUSK, KEEP, and CONCH. \textbf{c,} In the slide-level stage, ALICE extends its representation to whole-slide analysis by aligning high-resolution pathology image features with two slide-level PFMs, TITAN and CARE. The slide-level transformer is trained while the visual backbone remains frozen. \textbf{d-f,} Benchmark evaluation of ALICE against task-matched PFMs across seven vision-only, seven vision-language, and seven slide-level task scenarios. Radar plots show average performance for each task scenario.}
    \label{fig:main_figure_1}
\end{figure*}

Here, we introduce Agglomerative Learning via Integrated Computational pathology Embedding (ALICE), a general-purpose histopathology foundation model that provides a unified representation for diverse computational pathology tasks spanning region-of-interest (ROI) tissue analysis, vision-language alignment, and whole-slide clinical analysis. ALICE is trained through a multi-stage agglomerative distillation strategy using 24,985,184 low-resolution and 155,604 high-resolution histopathology images, with eight pathology foundation models trained under different data sources, learning objectives, and spatial scales serving as teachers (\textbf{Figure \ref{fig:main_figure_1}a-c}). Specifically, ALICE first learns morphology-oriented visual representations from three vision-only models, UNI-2 \cite{chen2024towards}, Virchow-2 \cite{zimmermann2024virchow2}, and H-Opt-1 \cite{scalbert2026abstract}. It then incorporates language-aligned representations from three vision-language models, CONCH \cite{lu2024visual}, MUSK \cite{xiang2025vision}, and KEEP \cite{zhou2026knowledge}, enabling tissue patterns to be associated with diagnostic concepts expressed in natural language. Finally, ALICE learns whole-slide contextual representations from two slide-level models, TITAN \cite{ding2025multimodal} and CARE \cite{zhang2026care}, extending its applicability from local tissue analysis to direct WSI-level analysis.

To evaluate ALICE, we establish a comprehensive benchmark organized around three complementary evaluation settings: vision-only, vision-language multimodal, and slide-level analysis. The benchmark spans 21 task scenarios, comprising 96 downstream tasks across 48 data sources (\textbf{Figure \ref{fig:main_figure_1}d–f}). Across all three evaluation settings, ALICE achieved the best average rank among task-matched PFMs, exceeding the second-best model by 1.79, 6.39, and 3.04 percentage points. Compared with the average performance of all remaining models, ALICE further improved performance by 3.10, 7.41, and 4.00 percentage points, respectively. These results demonstrate that structured agglomerative distillation can consolidate morphology-oriented, language-aligned, and whole-slide contextual expertise into a single framework, establishing ALICE as a broadly applicable foundation model for digital pathology.


\section{Results}

\subsection{ALICE outperforms vision-only PFMs in visual feature transfer}

\begin{figure*}[htbp]
    \centering
    \includegraphics[width=0.99\linewidth]{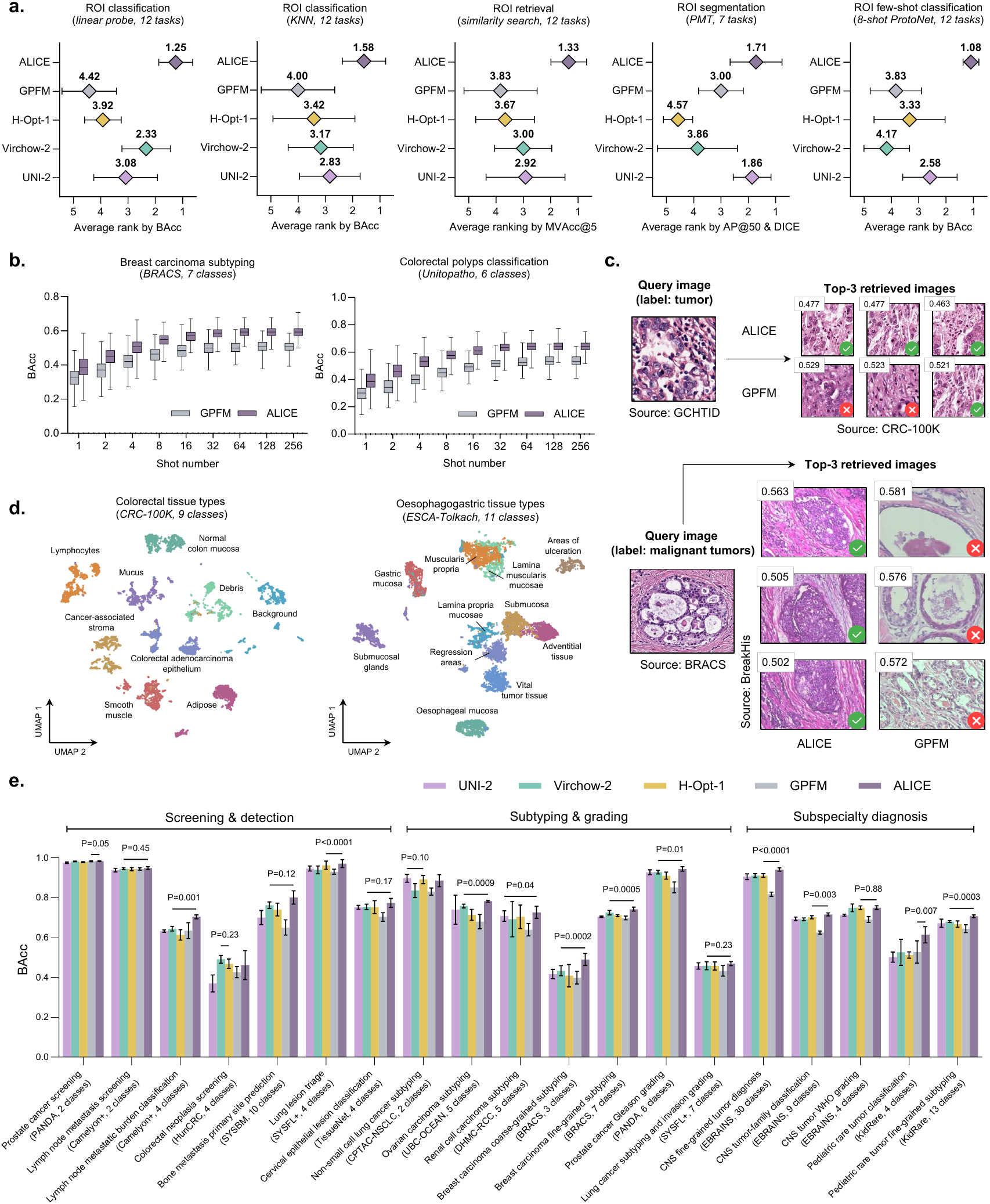}
    \caption{\textbf{Vision-only feature transfer performance of ALICE.} Caption on next page.}
    \label{fig:main_figure_2}
\end{figure*}
\begin{figure*}
  \caption*{(Previous page.) \textbf{Figure 2: Vision-only feature transfer performance of ALICE.} \textbf{a.} Average ranks of ALICE and four vision-only pathology foundation models across ROI-level transfer tasks, including linear-probe classification, KNN classification, image-to-image retrieval, segmentation using Plain Mask Transformer (PMT), and 8-shot few-shot classification with ProtoNet. \textbf{b.} ProtoNet few-shot ROI classification performance of ALICE and GPFM across different shot numbers on breast carcinoma subtyping (BRACS) and colorectal polyp classification (Unitopatho). \textbf{c.} Representative cross-source image-to-image retrieval examples comparing ALICE and GPFM. Query images and the top three retrieved images are shown with similarity scores; green checkmarks and red crosses indicate diagnostically matched and mismatched retrievals, respectively. \textbf{d.} UMAP visualization of ALICE embeddings for colorectal tissue typing in CRC-100K and oesophagogastric tissue typing in ESCA-Tolkach. \textbf{e.} Weakly supervised WSI classification using frozen vision-only features and an ABMIL classifier across 19 clinical tasks grouped into screening and detection, subtyping and grading, and CNS subspecialty diagnosis.}
\end{figure*}

We first evaluated ALICE in a vision-only feature transfer setting, a common deployment scenario in computational pathology in which ROIs or low-resolution tiles from WSIs are encoded into embeddings for downstream diagnostic, retrieval, and segmentation tasks \cite{chen2024towards}. We used each PFM as a frozen feature extractor and assessed the resulting embeddings either directly or with lightweight task-specific adapters. ALICE was compared with UNI-2, Virchow-2, H-Opt-1, and GPFM across seven representative evaluation scenarios, including ROI classification, image-to-image retrieval, ROI semantic and instance segmentation, few-shot classification, and weakly supervised WSI diagnostic classification using multiple instance learning (MIL).

At the ROI level, ALICE showed the strongest overall transferability across diverse evaluation protocols (\textbf{Figure \ref{fig:main_figure_2}a}). ALICE achieved the best average ranks under both linear probing and KNN classification settings, with average ranks of 1.25 and 1.58, respectively. ALICE also ranked first in image-to-image retrieval, with an average rank of 1.33 based on MVAcc@5. Although ALICE performed slightly below UNI-2 in ROI instance segmentation under the PMT setting (0.599 versus 0.607, AP@50), it still achieved the best overall rank across the seven segmentation tasks. Detailed per-dataset ROI classification, retrieval and segmentation results are provided in \textbf{Extended Data Figure~\ref{fig:extended_data_figure_2}} and \textbf{Extended Data Table \ref{tab:roi_lp}-\ref{tab:roi_seg}}. The advantage of ALICE was also particularly pronounced in data-efficient and non-parametric transfer settings. In few-shot classification, ALICE achieved an overall average rank of 1.08 across 12 ROI tasks, outperforming the second-best model by 1.50 rank positions, and consistently surpassed GPFM across all shot numbers in both breast carcinoma subtyping and colorectal polyp classification (\textbf{Figure \ref{fig:main_figure_2}b}, detailed results can be found in \textbf{Extended Data Figure~\ref{fig:extended_data_figure_3}} and \textbf{Extended Data Table~\ref{tab:roi_few_shot_protonet_aggc_5classes}-\ref{tab:roi_few_shot_protonet_unitopatho_6classes}}). In cross-source image-to-image retrieval, ALICE more reliably retrieved morphologically and diagnostically matched tissue regions than GPFM (\textbf{Figure \ref{fig:main_figure_2}c}). UMAP visualization further showed clear class-wise separation of ALICE features in colorectal and oesophagogastric tissue typing tasks (\textbf{Figure \ref{fig:main_figure_2}d}).

To investigate whether the low-resolution image features extracted by ALICE can support a standard weakly supervised slide-level diagnostic workflow, we further used ABMIL \cite{ilse2018attention} as a feature aggregator for WSI-derived tile images. We evaluated this setting across 19 clinical WSI classification tasks spanning early cancer detection, tumor typing and grading, and subspecialty diagnosis of the central nervous system. ALICE achieved best performance among the evaluated PFMs on 17 tasks and significantly outperformed the second-best model on 12 tasks (\textbf{Figure \ref{fig:main_figure_2}e} and \textbf{Extended Data Table \ref{tab:wsi_mil}}). These results indicate that the advantages of ALICE are not limited to ROI-level representation learning, but can also be translated into improved weakly supervised WSI classification when combined with the standard MIL aggregator.

\subsection{ALICE outperforms multimodal PFMs with text and vision features}

\begin{figure*}[htbp]
    \centering
    \includegraphics[width=0.99\linewidth]{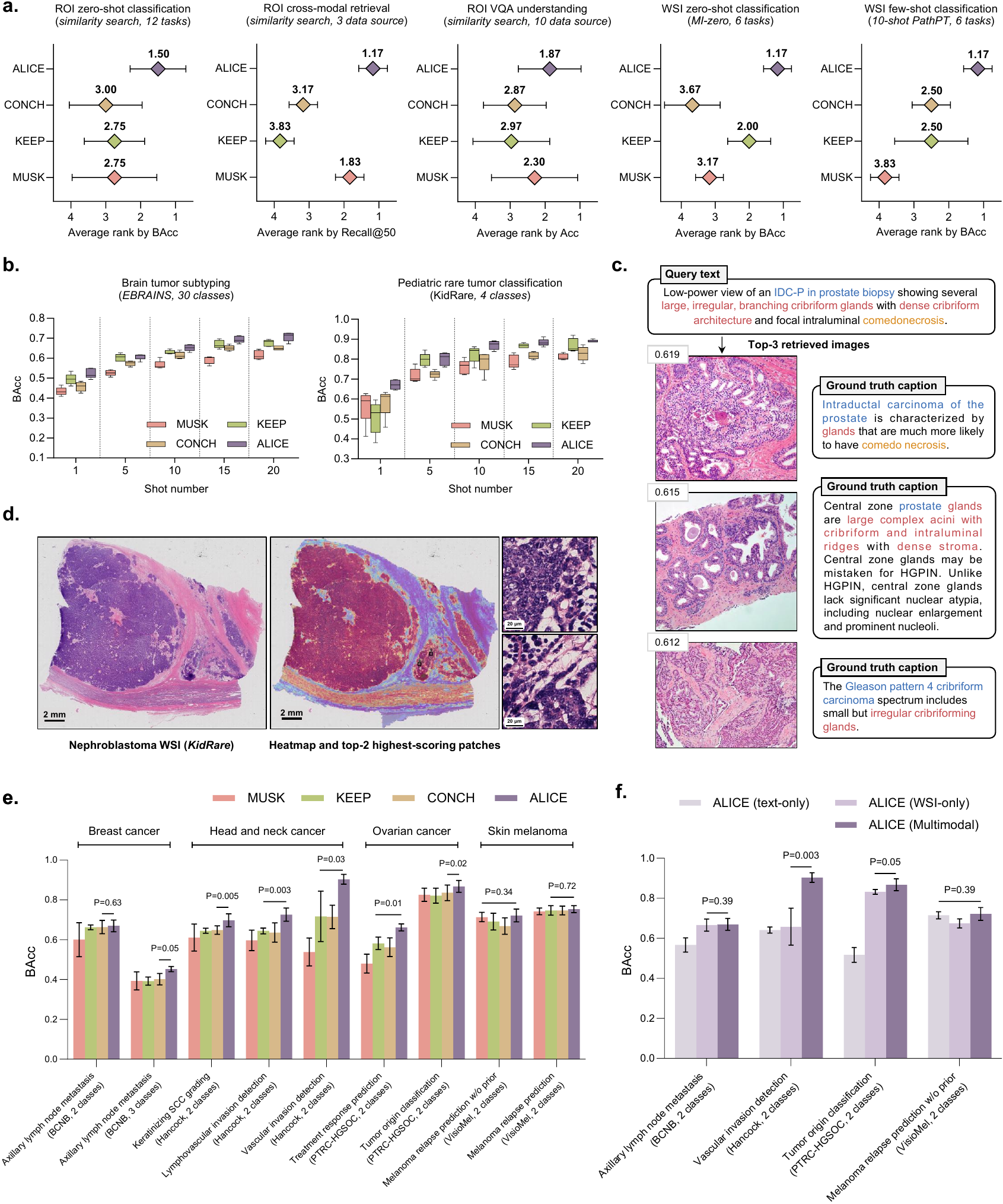}
    \caption{\textbf{Multimodal feature transfer performance of ALICE.} Caption on next page.} 
    \label{fig:main_figure_3}
\end{figure*}
\begin{figure*}
  \caption*{(Previous page.) \textbf{Figure 3: Multimodal feature transfer performance of ALICE.} \textbf{a.} Average ranks of ALICE and three multimodal pathology foundation models across multimodal evaluation settings, including ROI zero-shot classification, ROI cross-modal retrieval, ROI visual question answering understanding, WSI zero-shot classification using MI-Zero, and WSI few-shot classification using 10-shot PathPT. \textbf{b.} WSI PathPT few-shot classification performance of ALICE, MUSK, KEEP, and CONCH across different shot numbers on brain tumor subtyping (EBRAINS) and pediatric rare tumor classification (KidRare). \textbf{c.} Representative text-to-image retrieval example using a prostate pathology text description. The top three retrieved images are shown with similarity scores and corresponding ground-truth captions. \textbf{d.} Representative WSI-level prediction example for nephroblastoma from KidRare, showing the original WSI, attention heatmap, and the top two highest-scoring patches. \textbf{e.} Multimodal WSI clinical analysis using MICA adaptors across nine tasks spanning breast cancer, head and neck cancer, ovarian cancer, and skin melanoma. \textbf{f.} Modality ablation of ALICE on representative clinical WSI tasks, comparing text-only, WSI-only, and multimodal variants.
}
\end{figure*}

Pathology interpretation is inherently multimodal, requiring the integration of visual morphology with diagnostic terminology, textual descriptions, and clinical information \cite{acosta2022multimodal}. By aligning low-resolution histopathology image features with textual representations and transferring them to downstream tasks, multimodal PFMs can support a broader range of diagnostic workflows than vision-only models, including zero-shot classification, cross-modal retrieval, visual question answering (VQA), and text-guided WSI analysis. We compared ALICE with MUSK, KEEP, and CONCH, and benchmarked it across seven representative multimodal task scenarios: ROI zero-shot classification, ROI image-to-text and text-to-image retrieval, ROI VQA, WSI zero-shot classification, WSI few-shot classification, and multimodal WSI clinical prediction.

In the ROI setting, ALICE achieved the best overall performance (\textbf{Figure \ref{fig:main_figure_3}a}), ranking first across 12 zero-shot classification tasks with an average rank of 1.50 based on balanced accuracy. It also achieved the best average rank across three data sources in image-to-text and text-to-image retrieval, with an average rank of 1.17 based on recall@50. Furthermore, ALICE achieved the best average rank in ROI VQA, outperforming competing models in 6 of 10 tasks and reaching an overall average rank of 1.87. These results demonstrate that ALICE learns coordinated visual and textual representations that generalize to language-guided pathological interpretation. More detailed ROI vision-language transfer results are shown in \textbf{Extended Data Figure~\ref{fig:extended_data_figure_4}} and \textbf{Extended Data Table~\ref{tab:roi_zero_shot}-\ref{tab:pathmmu_vqa}}.

The strengths of ALICE were also evident in slide-level multimodal transfer. In six WSI zero-shot classification tasks using MI-Zero \cite{lu2023visual}, ALICE achieved the best average rank of 1.17 based on balanced accuracy (See detailed results in \textbf{Extended Data Table \ref{tab:zero_shot_comparison}}). In the 10-shot PathPT \cite{he2026boosting} setting, ALICE also outperformed the baseline models, achieving an average rank of 1.17 across six few-shot classification tasks. For example, in brain tumor subtyping and pediatric rare tumor classification, ALICE achieved competitive or better average performance across sample sizes (Figure \ref{fig:main_figure_3}b), demonstrating the effectiveness of its multimodal representations in data-limited slide-level diagnostic settings. More WSI few-shot classification results using PathPT can be found in \textbf{Extended Data Figure~\ref{fig:extended_data_figure_5}} and \textbf{Extended Data Table~\ref{tab:wsi_few_shot_pathpt_ebrains}-\ref{tab:wsi_few_shot_pathpt_cptac_nsclc}}.

Qualitative analysis further confirmed the multimodal alignment capability learned by ALICE. In text-to-image retrieval, the histological regions retrieved by ALICE were morphologically consistent with the query descriptions and corresponding ground truth labels, including prostate lesions with cribriform structures and comedo necrosis-related features (\textbf{Figure \ref{fig:main_figure_3}c}). In WSI-level prediction, by mapping patch-level target-class scores back onto the original WSI, the evidence heatmap for nephroblastoma highlighted diagnostically significant tumor regions, and the highest-scoring regions showed cellular and structural patterns consistent with the slide-level prediction (\textbf{Figure \ref{fig:main_figure_3}d}). These examples demonstrate that ALICE can link textual pathology concepts with local visual evidence. More representative ROI text-to-image, image-to-text, and WSI heatmaps are shown in \textbf{Extended Data Figure~\ref{fig:extended_data_figure_6}}, \textbf{\ref{fig:extended_data_figure_7}}, and \textbf{\ref{fig:extended_data_figure_8}}.

Clinical WSI analysis often extends beyond morphology-based slide classification, as many clinically relevant endpoints require joint interpretation of histological evidence and disease-specific clinical context. Tasks such as vascular invasion detection \cite{dorrich2025multimodal} and treatment response prediction \cite{chowdhury2023proteogenomic} may involve diagnostic clues that cannot be fully captured by image morphology alone. Therefore, we further evaluated ALICE using the MICA \cite{li2026ai} adapter on nine multimodal clinical WSI analysis tasks covering breast cancer, head and neck cancer, ovarian cancer, and cutaneous melanoma (\textbf{Figure \ref{fig:main_figure_3}e} and \textbf{Extended Data Table \ref{tab:wsi_multimodal_modality_comparison_appendix}}). ALICE achieved strong performance across these tasks, including axillary lymph node metastasis classification, keratinizing squamous cell carcinoma grading, lymphovascular and vascular invasion detection, ovarian cancer treatment response prediction, tumor origin classification, and melanoma recurrence prediction. To examine the contribution of each modality, we compared text-only, WSI-only, and multimodal versions of ALICE on representative tasks (\textbf{Figure \ref{fig:main_figure_3}f}). The multimodal version achieved the highest performance across all evaluation settings, particularly in vascular invasion detection and tumor origin classification. These results demonstrate that ALICE can use clinical text representations as supplementary information to WSI morphology, providing more robust multimodal slide-level inference for complex clinical diagnostic tasks.

\subsection{ALICE outperforms slide-level PFMs with high-resolution WSI features}

Unlike ROI-level or tile-level feature transfer, slide-level pathology foundation models aim to encode the global morphology, spatial organization, and tissue heterogeneity of whole-slide images into compact slide representations \cite{xu2024whole,ding2025multimodal,wang2024pathology,xu2025multimodal}. This setting is particularly important for clinical tasks in which diagnostic evidence is distributed across large tissue areas or depends on the overall architectural context of the slide, such as biomarker prediction and survival analysis. We compared ALICE with two representative slide-level PFMs, TITAN and CARE, across seven representative WSI task scenarios, including WSI diagnostic classification with KNN and linear probing, WSI few-shot classification, slide-to-slide retrieval, biomarker prediction, survival analysis, and task-specific fine-tuning.

\begin{figure*}[tbp]
    \centering
    \includegraphics[width=0.99\linewidth]{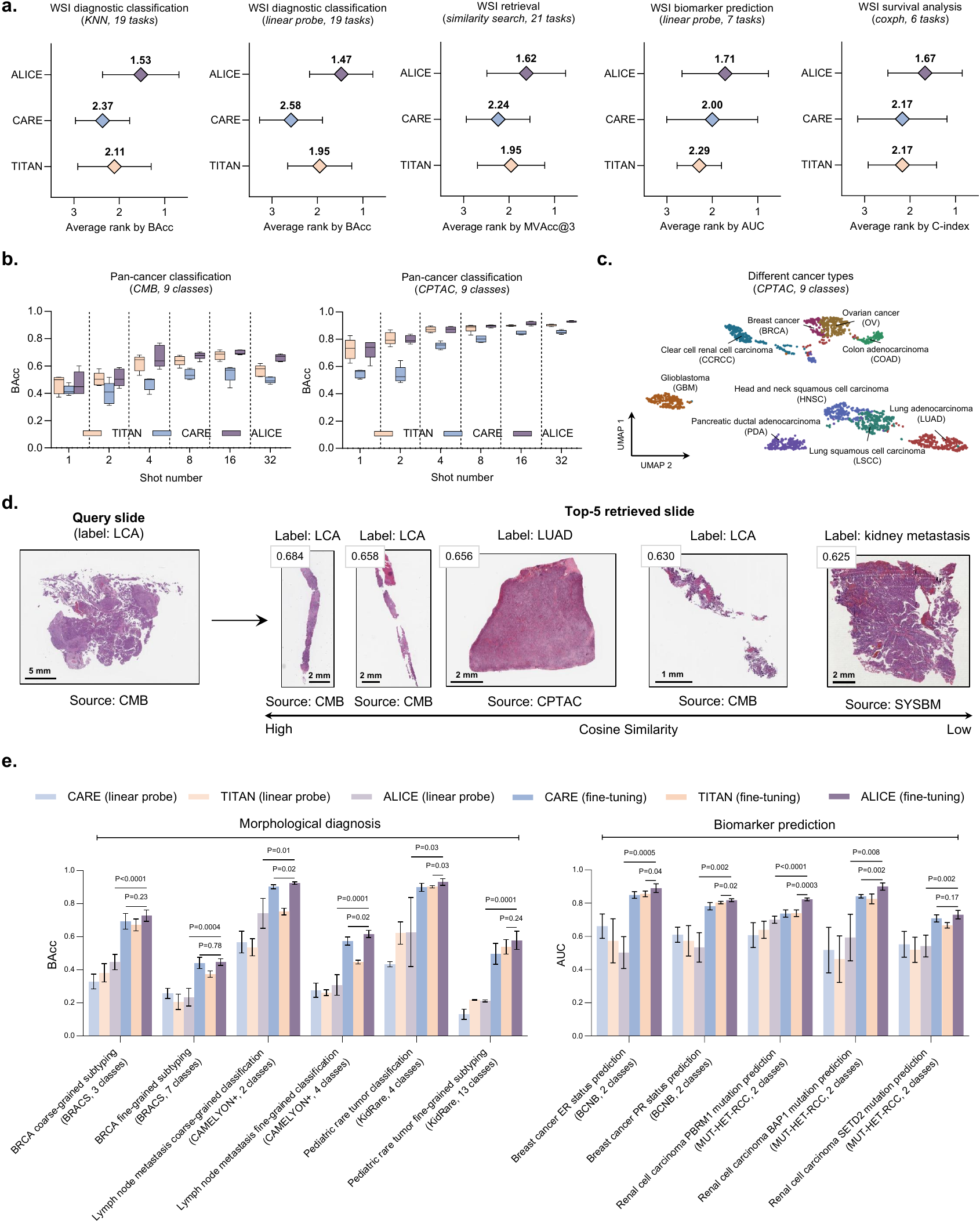}
    \caption{\textbf{Slide-level feature transfer performance of ALICE.} Caption on next page.}
    \label{fig:main_figure_4}
\end{figure*}

ALICE achieved the strongest overall performance across slide-level transfer tasks (\textbf{Figure \ref{fig:main_figure_4}a}). In WSI diagnostic classification, ALICE achieved the best average rank across 19 tasks under both KNN and linear probing settings, with average ranks of 1.53 and 1.47 based on balanced accuracy, respectively. ALICE also achieved the best average rank in slide-to-slide retrieval across 21 tasks (1.62, MVAcc@3), WSI biomarker prediction across seven tasks (1.71, AUC), and WSI survival analysis across six tasks (1.67, C-index). These results indicate that WSI features extracted by ALICE generalize across both morphology-driven and clinically oriented slide-level prediction tasks. Detailed slide-level diagnostic, biomarker and survival results are provided in \textbf{Extended Data Figure~\ref{fig:extended_data_figure_9}} and \textbf{Extended Data Table \ref{tab:knn_balacc_clinical_diagnosis_classcount_4dec}}-\textbf{\ref{tab:lp_ft_aligned_s5_pat30_balacc_clinical_diagnosis_classcount_4dec}}, \textbf{\ref{tab:lp_ft_aligned_s5_pat30_auc_biomarker_prediction_classcount_4dec}}-\textbf{\ref{tab:wsi_surv_cindex}}.

The advantage of ALICE was further supported by data-efficient and fixed-feature analyses. In few-shot pan-cancer classification, ALICE consistently achieved strong balanced accuracy across different shot numbers on both CMB and CPTAC, suggesting that its slide-level representations remain effective when only limited labeled slides are available (\textbf{Figure \ref{fig:main_figure_4}b}). UMAP visualization on CPTAC showed clear separation among nine cancer types, including breast cancer, clear cell renal cell carcinoma, colon adenocarcinoma, glioblastoma, head and neck squamous cell carcinoma, lung adenocarcinoma, lung squamous cell carcinoma, ovarian cancer, and pancreatic ductal adenocarcinoma (\textbf{Figure \ref{fig:main_figure_4}c}). In slide-to-slide retrieval, ALICE identified slides with identical or closely related diagnostic labels from three data sources, further supporting the ability of its high-resolution WSI features to capture global diagnostic similarity (\textbf{Figure \ref{fig:main_figure_4}d}). More representative slide-to-slide retrieval examples are shown in \textbf{Extended Data Figure~\ref{fig:extended_data_figure_10}}. Detailed few-shot slide-level classification can be found in \textbf{Extended Data Table \ref{tab:slide_few_shot_wsi_comparison}}.

\begin{figure*}
  \caption*{(Previous page.) \textbf{Figure 4: Slide-level feature transfer performance of ALICE.} \textbf{a.} Average ranks of ALICE and two slide-level pathology foundation models across WSI-level transfer tasks, including diagnostic classification with KNN and linear probing, slide-to-slide retrieval, biomarker prediction, and survival analysis. \textbf{b.} Linear probe few-shot pan-cancer classification performance of ALICE, TITAN, and CARE across different shot numbers on CMB and CPTAC. \textbf{c.} UMAP visualization of ALICE slide-level embeddings for nine cancer types from CPTAC. \textbf{d.} Representative slide-to-slide retrieval example using ALICE slide-level embeddings. A lung cancer adenocarcinoma (LCA) query slide from CMB is used to retrieve the top five most similar WSIs across CPTAC, CMB, and SYSBM data sources. \textbf{e.} Linear probe and fine-tuning performance of slide-level PFMs on morphological diagnosis and biomarker prediction tasks.}
\end{figure*}

Beyond frozen feature transfer, clinically specific WSI tasks often require adapting slide-level features to fine-grained morphological patterns and molecularly associated visual signals \cite{ding2025multimodal}. We therefore assessed whether high-resolution WSI features extracted by ALICE could be effectively specialized through task-specific fine-tuning (\textbf{Figure \ref{fig:main_figure_4}e} and \textbf{Extended Data Table \ref{tab:lp_vs_ft_diagnosis_results_setting_rows}-\ref{tab:lp_vs_ft_biomarker_results_setting_rows}}). Specifically, we jointly fine-tuned all slide-level branch parameters together with the downstream task-specific adapter. The evaluation included morphological diagnosis tasks, covering breast carcinoma subtyping, lymph node metastasis classification, and pediatric rare tumor classification, as well as biomarker prediction tasks, including breast cancer ER and PR status prediction and PBRM1, BAP1, and SETD2 mutation prediction in renal cell carcinoma. Fine-tuned ALICE achieved the best performance across the evaluated diagnostic and biomarker prediction settings, with notable gains in fine-grained tumor subtyping, lymph node metastasis burden classification, rare tumor subtyping, and renal cell carcinoma mutation prediction. These findings indicate that ALICE learns high-resolution WSI representations with strong adaptation capacity, enabling task-specific fine-tuning to extract clinically relevant slide-level signals more effectively than existing slide-level PFMs.

\subsection{Discussion}

In this paper we introduced ALICE, a unifed PFM that combines the complementary strengths of existing models into a single backbone. Although recent PFMs have substantially expanded the capabilities of computational pathology, most remain specialized by design: some excel at capturing local tissue morphology \cite{chen2024towards,vorontsov2024foundation,scalbert2026abstract}, others at aligning histology with diagnostic language \cite{lu2024visual,xiang2025vision,zhou2026knowledge}, and others at modeling whole-slide context \cite{ding2025multimodal,xu2024whole,zhang2026care,wang2024pathology}. This specialization creates a fragmented landscape in which different clinical tasks often require different pretrained models. ALICE addresses this challenge through multi-stage agglomerative distillation, progressively integrating morphology-oriented, language-aligned, and slide-level expertise into a unified representation framework. Across vision-only, multimodal, and slide-level evaluations, ALICE showed broad transferability across ROI-level tissue analysis, language-guided pathology tasks, and high-resolution WSI representation learning. These findings suggest that the diverse capabilities of current PFMs can be consolidated into a single broadly applicable model, providing a more unified foundation for computational pathology across local, semantic, and whole-slide clinical contexts.

Several observations from our benchmark are noteworthy. First, ALICE did not simply reproduce the average behavior of its teacher models. Instead, it retained strong local visual discrimination from vision-only PFMs, acquired language-guided interpretation from vision-language PFMs, and remained effective for high-resolution WSI representation learning. This suggests that the apparent boundaries between different families of PFMs are not fixed, but can be softened through structured knowledge integration. Second, the benefits of ALICE were observed across distinct transfer regimes, including frozen feature evaluation, nonparametric retrieval, few-shot learning, multimodal inference, and task-specific fine-tuning. This indicates that agglomerative distillation can produce representations that are not only broadly reusable but also adaptable when downstream supervision is available. Third, the slide-level results support the value of integrating local morphology and whole-slide context for clinically complex endpoints, including fine-grained diagnosis, biomarker prediction, and survival analysis. Together, these findings suggest that future PFM development may benefit from being organized along the continuum of clinical reasoning, from local tissue morphology to semantic interpretation and whole-slide decision-making.

Despite these encouraging results, several limitations point to future directions. First, ALICE was evaluated primarily in retrospective settings. Prospective, multi-institutional studies will be needed to assess its robustness under real-world variation in staining protocols, scanner platforms, tissue processing, and patient populations. Second, the current pretraining data from TCGA and HISTAI \cite{nechaev2025histai} provide an effective basis for multi-stage agglomerative distillation, but their scale and source distribution can be further expanded. Future work could incorporate larger, more diverse institutional, international, and disease-specific cohorts to improve tissue coverage, domain robustness, and generalization to rare diagnostic entities \cite{vaidya2024demographic}. Third, although the current multi-stage architecture enables structured integration of vision-only, multimodal, and slide-level expertise, it may not yet be optimal for computational efficiency and streamlined deployment. A more unified and efficient model framework could preserve the benefits of staged knowledge aggregation while reducing structural fragmentation, improving inference speed, and facilitating scalable downstream adaptation. Finally, additional clinically informative modalities, such as immunohistochemistry \cite{li2026stainnet} and genomic profiles, could be integrated to extend ALICE toward a broader computational pathology system.

In conclusion, ALICE demonstrates that complementary capabilities from existing pathology foundation models can be integrated into a unified and broadly transferable backbone. By consolidating morphology-oriented, language-aligned, and slide-level expertise, ALICE provides a flexible foundation for computational pathology tasks spanning local tissue analysis, multimodal interpretation, and whole-slide clinical inference. We envision ALICE and its future iterations serving as a general-purpose platform for developing more efficient, scalable, and clinically adaptable pathology AI systems.

\newpage


\section{Methods}
\subsection{Pretraining datasets and image preprocessing}
ALICE pretraining consisted of three stages, including vision-only distillation, multimodal distillation, and slide-level distillation. The first two stages were performed using low-resolution histopathology image patches, and the final stage used high-resolution regions derived from whole-slide images (WSIs).

For low-resolution image pretraining, we used TCGA-12K \cite{karasikov2025training}, a large-scale patch dataset containing 24,985,184 image patches of $224 \times 224$ pixels extracted from approximately 12K WSIs from The Cancer Genome Atlas (TCGA), covering 32 cancer types. The patches were randomly sampled across multiple magnification levels, and tissue-containing patches were retained using HSV-threshold-based tissue filtering.

For high-resolution image pretraining, we used TCGA-UT-8K \cite{ding2025multimodal} and additional high-resolution image regions cropped from WSIs in HISTAI \cite{nechaev2025histai}. TCGA-UT-8K contains 25,495 tumor-centered ROI images of $8192 \times 8192$ pixels extracted from 9,662 H\&E-stained FFPE diagnostic WSIs in TCGA, covering 32 pan-cancer subtyping classes. Each ROI was center-cropped from pathologist-annotated tumor contours to include both dense tumor areas and surrounding tissue context. For the HISTAI-derived high-resolution data, tissue-containing regions were tiled at $20\times$ magnification into non-overlapping image regions of $8192 \times 8192$ pixels. After tissue filtering and quality control, we obtained 130,109 high-resolution regions from 35,363 slides. These regions were distributed across 5 HISTAI subsets, including breast, colorectal, mixed, skin, and thorax cohorts. The resulting TCGA-UT-8K and HISTAI high-resolution feature sets were used for slide-level distillation.

\subsection{Teacher foundation models}
We selected eight publicly available expert pathology foundation models as teacher models for ALICE pretraining. These models were organized into three groups corresponding to the three stages of agglomerative distillation: vision-only PFMs, vision-language PFMs, and slide-level PFMs. This grouping was designed to cover complementary representation capabilities, including dense morphological encoding, language-aligned diagnostic semantics, and whole-slide clinical context.

For the vision-only distillation stage, we used UNI-2 \cite{chen2024towards}, Virchow-2 \cite{zimmermann2024virchow2}, and H-Opt-1 \cite{scalbert2026abstract} as teacher models. This group was selected to provide strong morphology-oriented visual supervision from large-scale self-supervised histopathology pretraining. UNI-2 is a vision-only PFM based on a ViT-H/14 architecture with 681 million parameters, pretrained with DINOv2 \cite{oquab2023dinov2} on more than 200 million H\&E- and IHC-stained histology tiles sampled from over 350,000 slides from Mass General Brigham. Virchow-2 is a vision-only PFM based on a ViT-H/14 architecture with 632 million parameters, pretrained with a pathology-adapted DINOv2 strategy on 3.1 million WSIs from diverse institutions. H-Opt-1 is a vision-only PFM based on a ViT-G/14 architecture with 1.1 billion parameters, pretrained with self-supervised learning on billions of histology images sampled from more than one million H\&E-stained slides across over 50 organs.

For the multimodal distillation stage, we used CONCH \cite{lu2024visual}, MUSK \cite{xiang2025vision}, and KEEP \cite{zhou2026knowledge} as teacher models. This group was selected to introduce language-aligned diagnostic semantics into ALICE, enabling histological morphology to be associated with natural-language descriptions, diagnostic concepts, and pathology-specific knowledge. CONCH is a vision-language PFM based on a ViT-B/16 image encoder paired with a text encoder, jointly pretrained on 1.17 million histopathology image-caption pairs using contrastive captioning. MUSK is a vision-language PFM based on a unified multimodal transformer, pretrained with unified masked modeling on 50 million pathology images and one billion pathology-related text tokens from unpaired sources. KEEP is a knowledge-enhanced vision-language PFM based on a ViT-L/16 image encoder, pretrained with disease-knowledge-guided contrastive learning using a disease knowledge graph containing 11,454 disease entities to organize pathology image-text pairs into semantically structured groups.

For the slide-level distillation stage, we used TITAN \cite{ding2025multimodal} and CARE \cite{zhang2026care} as teacher models. This group was selected to provide high-resolution WSI-level supervision and to extend ALICE from local image representation to whole-slide contextual modeling. TITAN is a slide-level PFM that operates on CONCH 1.5 patch features and aggregates them into slide-level representations through a transformer backbone with ALiBi-based self-attention \cite{press2021train}. It was pretrained on 335,645 WSIs through visual self-supervised learning and vision-language alignment using pathology reports and synthetic ROI captions. CARE is a slide-level PFM that also operates on CONCH 1.5 patch features but adopts adaptive region modeling to partition WSIs into morphologically coherent regions. It was pretrained on 34,277 WSIs through visual self-supervised learning and cross-modal alignment with RNA and protein expression profiles.

\subsection{ALICE architecture}
ALICE is organized into three transformer-based modules that are sequentially activated during pretraining: a vision-only transformer, a multimodal transformer, and a slide-level transformer (\textbf{Extended Data Figure \ref{fig:extended_data_figure_1}}). The vision-only transformer encodes low-resolution histopathology patches and provides the shared image representation backbone. The multimodal transformer is attached after the vision-only transformer to adapt visual tokens for vision-language alignment. The slide-level transformer aggregates high-resolution patch features and their spatial coordinates into WSI-level representations. This modular design allows ALICE to progressively incorporate morphology-oriented, language-aligned, and slide-level knowledge while keeping each stage structurally distinct.

The vision-only transformer is implemented as a ViT-H/14 encoder. Input images are first converted into patch tokens through a patch embedding layer, followed by conditional positional embedding \cite{chu2021conditional} and learnable prefix tokens \cite{darcet2024vision}. The prefix tokens comprise three summary tokens for image-level representation learning and additional register tokens to stabilize token-level encoding. The vision-only transformer processes the resulting token sequence and outputs both summary-level representations and spatial patch-token representations, enabling feature distillation at both global and local levels.

The multimodal transformer is implemented as a lightweight self-attention adaptor placed on top of the pretrained vision-only transformer. During multimodal distillation, the vision-only transformer is frozen, and only this multimodal transformer is optimized. The module contains two transformer layers and operates on the visual tokens produced by the frozen image encoder. Its role is to transform the morphology-oriented visual representations learned in the first stage into a representation space suitable for alignment with vision-language teacher models. The slide-level transformer is not used during this stage.

The slide-level transformer is trained in the final distillation stage using offline patch features extracted from the pretrained image encoder. Specifically, each high-resolution ROI was first divided into several image patches, which were then encoded by the vision-only transformer pretrained in the first stage to obtain fixed patch-level features. In this stage, the vision-only transformer is effectively frozen, the multimodal transformer is unused, and only the slide-level transformer is updated. For each high-resolution region or WSI, the slide-level transformer takes the extracted patch features and their spatial coordinates as input. Patch features are projected from 3,840 dimensions into a 1,024-dimensional token space, and normalized patch coordinates are encoded through a coordinate projection module to provide spatial information. The slide-level transformer uses two learnable summary tokens as global aggregation tokens. These tokens are prepended to the patch-token sequence and are updated together with the patch tokens by two ALiBi \cite{press2021train} transformer blocks. Each block uses multi-head self-attention with spatial bias computed from pairwise distances between patch coordinates, allowing the model to account for the relative organization of tissue regions within the slide. The final slide-level representation is obtained by flattening the two summary tokens into a 2,048-dimensional embedding. Through this architecture, ALICE combines local patch morphology, multimodal semantic adaptation, and spatially aware whole-slide representation learning within a staged transformer framework.

\subsection{Multi-stage agglomerative distillation pretraining}
Existing PFMs encode complementary forms of expertise, but these capabilities are distributed across separate architectures, objectives, and spatial scales. Inspired by RADIO \cite{heinrich2025radiov2}, we pretrained ALICE through a three-stage agglomerative distillation process that sequentially transfers knowledge from vision-only, vision-language, and slide-level teachers into a unified student representation.

In the first stage, ALICE learned morphology-oriented visual representations from three frozen vision-only PFMs: H-Opt-1, Virchow-2, and UNI-2. The student model contained three learnable summary tokens, each assigned to a single teacher, allowing the student to absorb teacher-specific global representations without forcing all teachers into a shared target space. During training, for an image $i$, each teacher-specific summary token was projected to the corresponding teacher feature dimension and aligned with the frozen teacher CLS embedding using a balanced angular distillation loss:
\begin{equation}
    \mathcal{L}_{\mathrm{cls}}^t = \frac{\Theta(P_t(z_i^t), c_i^t)^2}{\mathrm{Disp}(\Theta_t)}, \quad \Theta(\mathbf{x}, \mathbf{y})=\arccos(\cos(\mathbf{x}, \mathbf{y})), \quad \mathrm{Disp}(\Theta_t)=\mathbb{E}\left[ \Theta(c_i^t, \mu_t)^2 \right],
\end{equation}
where $z_i^t$ is the student summary token assigned to teacher $t$, $c_i^t$ is the teacher CLS embedding, $P_t$ is a teacher-specific projection head, and $\mu_t$ is the expected unit direction of teacher CLS embeddings over the training set. This normalization balances teachers with different feature-space dispersions and prevents teachers with larger angular variance from dominating the global distillation objective. To further transfer local morphology, we used shift-equivariant patch distillation, in which the student and each teacher independently received grid-aligned crops sampled from the same augmented image, and patch tokens were matched only at overlapping spatial positions. The patch-level loss is defined as:
\begin{equation}
\mathcal{L}_{\mathrm{patch}}^t =
\frac{1}{|\Omega_t|}
\sum_{(i,u)\in\Omega_t}
\left\| Q_t(p_{i,u}) - q_{i,u}^t \right\|_2^2 ,
\end{equation}
where $p_{i,u}$ denotes the student patch tokens and $q_{i,u}^t$ denotes PHI-S-normalized teacher patch tokens at spatial position $u$, $Q_t$ is a teacher-specific patch projection head, and $\Omega_t$ denotes the set of overlapping spatial positions between the student and teacher crops. This loss transfers local morphological information while preserving spatial correspondence under independently sampled crops. Finally, to stabilize patch-level learning, an exponential moving-average (EMA) copy of the student was maintained as an additional self-distillation target.

In the second stage, ALICE incorporated language-aligned knowledge from three vision-language PFMs: CONCH, MUSK, and KEEP. To preserve the morphology-oriented representations learned in the first stage, the vision-only transformer was initialized from the first-stage checkpoint and kept frozen. The multimodal transformer was then trained on top of the frozen backbone tokens.  For each summary token, ALICE combined it with the average patch representation and aligned it with the image embedding using $\mathcal{L}_{\mathrm{cls}}^t$. For patch token supervision, each crop was generated at the native input resolution, then spatially resampled onto the student grid before shift-equivariant alignment.

In the third stage, ALICE was extended from local tissue representation learning to slide-level representation learning by distilling high-resolution image knowledge from TITAN and CARE. For each $8192 \times 8192$ high-resolution region, we divided the image into a $16 \times 16$ grid of non-overlapping crops, generating 256 tiles per region. Tile-level features were extracted using the first-stage ALICE vision-only transformer and used as fixed inputs to the slide-level transformer. The corresponding teacher slide embeddings were obtained from TITAN and CARE using their required tile-level CONCH 1.5 feature inputs. Each ALICE tile feature was projected into a 1,024-dimensional hidden space through a projection and augmented with coordinate embeddings derived from normalized tile coordinates. Two learnable summary tokens were prepended to the tile sequence and jointly processed with all tile tokens by the slide-level transformer. The first summary token was aligned with the TITAN slide embedding, and the second summary token was aligned with the CARE slide embedding using $\mathcal{L}{\mathrm{cls}}^t$. Since TITAN additionally produces tile-level output tokens together with its slide embedding, we applied $\mathcal{L}{\mathrm{patch}}^t$ between the student tile tokens and the corresponding TITAN tile tokens at matched spatial positions, providing fine-grained contextual supervision within the slide representation. CARE was used only for slide-level supervision. Through this stage, ALICE learned to aggregate fixed local tissue features into spatially aware slide-level representations.

\subsection{Pretraining settings}
For all three pretraining stages, we used distributed training on eight 80GB NVIDIA A100 GPUs. For vision-only and multimodal stages, each input image first underwent a shared augmentation pipeline consisting of random resized cropping, horizontal flipping, and color jittering, followed by teacher-specific cropping and normalization for summary-level and patch-level distillation. For the slide-level stage, since no raw image pixels were used, we applied embedding-level feature-brightness perturbation as feature-space augmentation. We used gradient accumulation to keep the total effective batch size fixed at 1,024 across stages. To further enhance the robustness of our ALICE, we adopted DAMP \cite{trinh2024improving}, which applies multiplicative noise to the model weight across three stages. Training was monitored every 1,000 optimization steps, and early stopping was used if the monitored loss did not improve for 5 consecutive intervals. Complete stage-specific hyperparameters are reported in \textbf{Extended Data Table~\ref{tab:stage1_pretrain_settings}}, \textbf{\ref{tab:stage2_pretrain_settings}}, and \textbf{\ref{tab:stage3_pretrain_settings}}.

\subsection{Downstream benchmark}
We evaluated ALICE on a broad downstream benchmark spanning ROI analysis, vision-language multimodal analysis and whole-slide image analysis. The benchmark included 96 downstream tasks from 48 data sources.

\noindent\textbf{AGGC}\cite{huo2024comprehensive} was constructed from the AGGC2022 prostate histopathology dataset for ROI-level automated Gleason grading. We conducted experiments on the biopsy subset, which included 37 training WSIs and 16 test WSIs. Using the provided label masks, we extracted non-overlapping $224\times224$ image patches and retained only patches for which more than 50\% of the patch area overlapped with one target annotation mask. The resulting dataset consisted of 42,157 patches classified into five categories: Gleason pattern 3 (10,486), Gleason pattern 4 (5,615), Gleason pattern 5 (106), normal glands (12,776), and stroma (13,174). We used the officially defined AGGC2022 train:test split, yielding 29,366 training patches and 12,791 test patches (69.66\%:30.34\%). We used AGGC for downstream ROI classification, ROI retrieval, ROI few-shot classification, and ROI zero-shot classification tasks.  

\noindent\textbf{BRACS}\cite{brancati2022bracs} was constructed from the BRACS H\&E breast histopathology dataset for ROI- and WSI-level breast cancer subtyping. For ROI-level experiments, we used 4,539 annotated ROI images under two label granularities. The 3-class setting consisted of benign tumors (1,837), atypical tumors (1,263), and malignant tumors (1,439), and the 7-class setting consisted of normal (484), pathological benign (836), usual ductal hyperplasia (517), flat epithelial atypia (756), atypical ductal hyperplasia (507), ductal carcinoma in situ (790), and invasive carcinoma (649). We used the officially defined ROI train:test split for both settings, yielding 3,969 training images and 570 test images (87.44\%:12.56\%). For WSI-level experiments, BRACS consisted of 546 WSIs. The coarse WSI task used benign tumors (264), atypical tumors (89), and malignant tumors (193), and the fine-grained WSI task used normal (43), pathological benign (147), usual ductal hyperplasia (74), flat epithelial atypia (41), atypical ductal hyperplasia (48), ductal carcinoma in situ (61), and invasive carcinoma (132). We used the officially defined WSI train:val:test split for both WSI tasks, yielding 394 training, 65 validation, and 87 test WSIs (72.16\%:11.90\%:15.93\%). We used BRACS for downstream ROI classification, ROI retrieval, ROI few-shot classification, ROI zero-shot classification, WSI classification, and WSI retrieval tasks.  

\noindent\textbf{BreakHis}\cite{spanhol2015dataset} is a dataset of H\&E-stained breast tumor histopathological images for ROI-level tasks. We used the 2-class setting, which included 7,909 images from 82 cases across four magnification levels (40$\times$, 100$\times$, 200$\times$, and 400$\times$). The dataset consisted of two categories: benign tumors (2,480) and malignant tumors (5,429). The benign group included adenosis, fibroadenoma, phyllodes tumor, and tubular adenoma, whereas the malignant group included ductal carcinoma, lobular carcinoma, mucinous carcinoma, and papillary carcinoma. We split the dataset at the case level with stratification by histological subtype, yielding 5,344 training images and 2,565 test images (67.57\%:32.43\%). We used BreakHis for downstream ROI classification, ROI retrieval, ROI few-shot classification, and ROI zero-shot classification tasks.  

\noindent\textbf{CCRCC}\cite{brummer2023computational} is constructed from a clear-cell renal cell carcinoma tissue classification dataset for ROI-level renal tissue recognition. The dataset consisted of 52,713 ROI images classified into six categories: blood (996), cancer (13,057), empty or background regions (16,026), normal tissue (8,652), other tissue (8,522), and stroma (5,460). We split the dataset at the slide level with case stratification, yielding 36,900 training images and 15,813 test images (70.00\%:30.00\%). We used CCRCC for downstream ROI classification, ROI retrieval, ROI few-shot classification, and ROI zero-shot classification tasks.  

\noindent\textbf{Chaoyang}\cite{zhu2021hard} is a colorectal biopsy histopathology dataset for ROI-level gastrointestinal tissue classification. The dataset consisted of 6,160 ROI images classified into four categories: normal (1,816), serrated (1,163), adenocarcinomas (2,244), and adenomas (937). We used the officially defined train:test split, yielding 4,021 training images and 2,139 test images (65.28\%:34.72\%). We used Chaoyang for downstream ROI classification, ROI retrieval, ROI few-shot classification, and ROI zero-shot classification tasks.  

\noindent\textbf{CRC-100K}\cite{kather_jakob_nikolas_2018_1214456} is constructed from the NCT-CRC-HE-100K and CRC-VAL-HE-7K H\&E colorectal histopathology patch datasets for tissue classification. The dataset consisted of 87,174 image patches classified into nine tissue categories: adipose tissue (ADI, 9,663), background (BACK, 9,299), debris (DEB, 9,548), lymphocytes (LYM, 9,879), mucus (MUC, 8,151), smooth muscle (MUS, 11,420), normal colon mucosa (NORM, 7,751), cancer-associated stroma (STR, 8,777), and colorectal adenocarcinoma epithelium (TUM, 12,686). We used NCT-CRC-HE-100K as the training set and CRC-VAL-HE-7K as the held-out test set, yielding 79,994 training patches and 7,180 test patches (91.76\%:8.24\%). We used CRC-100K for downstream ROI classification, ROI retrieval, ROI few-shot classification, ROI zero-shot classification, and PathMMU PathCLS VQA tasks.  

\noindent\textbf{CRC-MSI}\cite{kather2019deep} is constructed from H\&E colorectal cancer histopathology patches with microsatellite instability labels for biomarker-associated classification. The dataset consisted of 51,918 image patches from 428 slides and 423 patients, classified into two categories: microsatellite instability-high (MSIH, 15,002) and non-microsatellite instability-high (nonMSIH, 36,916). We used the officially defined train:test split, yielding 19,557 training patches and 32,361 test patches (37.67\%:62.33\%).We used CRC-MSI for downstream ROI classification, ROI retrieval, ROI few-shot classification, and ROI zero-shot classification tasks.  

\noindent\textbf{ESCA-Tolkach}\cite{tolkach2023artificial} is constructed from an H\&E esophageal cancer histopathology tissue classification dataset for esophageal tissue recognition. The dataset consisted of 334,533 image patches classified into eleven categories using the original directory labels: adventitia (68,118), lamina propria (1,901), muscularis mucosae (2,542), muscularis propria (73,532), regressive tumor (56,490), gastric mucosa (40,928), esophageal mucosa (17,276), submucosa (19,150), submucosal glands (1,406), tumor (52,642), and ulcer (548). We used a cohort-level train:test split, with the UKK and WNS cohorts for training and the CHA cohort for testing, yielding 156,346 training patches and 178,187 test patches (46.74\%:53.26\%). We used ESCA-Tolkach for downstream ROI classification, ROI retrieval, ROI few-shot classification, and ROI zero-shot classification tasks.   

\noindent\textbf{GCHTID}\cite{lou2025large} is a gastric cancer histopathology tissue image dataset for gastric tissue classification. The dataset consisted of 26,744 image patches classified into eight balanced tissue categories: adipose tissue (3,343), debris (3,343), lymphocytes (,343), mucus (3,343), muscle (3,343), normal tissue (3,343), stroma (3,343), and tumor (3,343). We used the benchmark train:test split, yielding 17,408 training patches and 9,336 test patches (65.09\%:34.91\%). We used GCHTID for downstream ROI classification, ROI retrieval, ROI few-shot classification, and ROI zero-shot classification tasks.  

\noindent\textbf{OTA}\cite{leavey2019osteosarcoma} is the H\&E histopathology image dataset for ROI-level osteosarcoma tissue classification. The dataset consisted of 1,091 image patches classified into three categories: non-tumor tissue (536), non-viable tumor (263), and viable tumor (292). We split the dataset into train:test sets, yielding 763 training patches and 328 test patches (69.94\%:30.06\%). We used OTA for downstream ROI classification, ROI retrieval, ROI few-shot classification, and ROI zero-shot classification tasks.  

\noindent\textbf{Unitopatho}\cite{barbano2021unitopatho} is a colorectal polyp histopathology dataset for ROI-level polyp classification and adenoma dysplasia grading. The dataset consisted of 8,669 H\&E image patches extracted from colorectal polyp WSIs. The dataset contained six categories: normal tissue (950), hyperplastic polyp (545), tubular adenoma with high-grade dysplasia (454), tubular adenoma with low-grade dysplasia (3,618), tubulo-villous adenoma with high-grade dysplasia (916), and tubulo-villous adenoma with low-grade dysplasia (2,186). We used the officially defined train:test split, yielding 6,065 training patches and 2,604 test patches (69.96\%:30.04\%). We used Unitopatho for downstream ROI classification, ROI retrieval, ROI few-shot classification, and ROI zero-shot classification tasks.  

\noindent\textbf{CoNSeP}\cite{graham2019hover} is a colorectal adenocarcinoma H\&E nuclear segmentation dataset. The dataset contains 41 ROI images, and a total of 24,332 kernel instances are labeled. The annotations covered seven nuclear categories: other (932), inflammatory (5,579), healthy epithelial (1,620), dysplastic or malignant epithelial (7,131), fibroblast (7,554), muscle (1,396), and endothelial (120). We used the officially defined train:test split and further divided the training set into train and validation sets, yielding 23 training images, 4 validation images, and 14 test images (56.10\%:9.76\%:34.15\%). We used CoNSeP for the downstream ROI instance segmentation task.  

\noindent\textbf{CRAG}\cite{graham2019mild} is a colorectal adenocarcinoma H\&E gland segmentation dataset. The dataset consisted of 213 ROI images with binary gland masks, from which gland instances were derived using connected components. The dataset contained 3,047 annotated instances. We used the officially defined train:val:test split, yielding 158 training images, 15 validation images, and 40 test images (74.18\%:7.04\%:18.78\%). We used CRAG for the downstream ROI instance segmentation task. 

\noindent\textbf{GlaS}\cite{sirinukunwattana2017gland} is a colorectal H\&E gland segmentation dataset containing benign and malignant histology images with instance-level annotations. The dataset consisted of 165 ROI images and 1,530 annotated gland instances. We used the officially defined train, testA, and testB cohort as train:val:test, yielding 85 training images, 60 validation images, and 20 test images (51.52\%:36.36\%:12.12\%). We used GlaS for the downstream ROI instance segmentation task. 

\noindent\textbf{CoCaHis}\cite{sitnik2021dataset} is an ROI-level H\&E segmentation dataset for colorectal cancer histopathology. The dataset consisted of 82 ROI images, each with a size of+- $1388\times1037$ pixels. The pixel-level annotations covered two categories: background and tumor region, with one mask for each category in every image. We used the officially defined train:val:test split, yielding 65 training images, 6 validation images, and 11 test images (79.27\%:7.32\%:13.41\%). We used CoCaHis for the downstream ROI semantic segmentation task. 

\noindent\textbf{COSAS24} is an ROI-level H\&E segmentation dataset for colorectal adenocarcinoma histopathology. The dataset consisted of 360 ROI images with pixel-level segmentation masks. The annotations covered 350 background regions and 301 adenocarcinoma regions. We used the officially defined train:val:test split, yielding 289 training images, 33 validation images, and 38 test images (80.28\%:9.17\%:10.56\%). We used COSAS24 for the downstream ROI semantic segmentation task. 

\noindent\textbf{EBHI}\cite{hu2023ebhi} is an endoscopic biopsy H\&E histopathology dataset for ROI-level segmentation tasks. The dataset consisted of 2,174 image patches, each with a size of $224\times224$ pixels. The pixel-level annotations covered seven categories: background (2,139 regions), adenocarcinoma (775), high-grade intraepithelial neoplasia (177), low-grade intraepithelial neoplasia (615), normal tissue (76), polyp (473), and serrated adenoma (58). We used the officially defined train:val:test split, yielding 1,719 training images, 223 validation images, and 232 test images (79.07\%:10.26\%:10.67\%). We used EBHI for the downstream ROI semantic segmentation task. 

\noindent\textbf{Janowczyk}\cite{janowczyk2016deep} is an ROI-level histopathology segmentation dataset consisting of 141 ROI images, each with a size of $2000\times2000$ pixels. The pixel-level annotations covered two categories: background and foreground, with one mask for each category in every image. We used the benchmark train:val:test split, yielding 106 training images, 20 validation images, and 15 test images (75.18\%:14.18\%:10.64\%). We used Janowczyk for the downstream ROI semantic segmentation task. 

\noindent\textbf{BookSet}\cite{gamper2020multiple} is an English pathology image-text dataset derived from the ARCH histopathology captioning resource, which collects pathology textbook and article figures with dense diagnostic and morphological captions. The original matched set contained 4,267 image-caption pairs. We filtered non-H\&E-stained images and retained 3,335 image-caption pairs. We used BookSet for the downstream ROI image-to-text and text-to-image retrieval task. 

\noindent\textbf{ChineseBook} is a self-constructed pathology image-text dataset collected from 8 Chinese clinical diagnostic atlas books. To support the vision-language benchmark, we extracted histopathology figures and their accompanying Chinese captions using PaddleOCR and translated the captions into English using Qwen2.5-14B, creating 1,437 image-English text pairs for evaluation. We used ChineseBook for the downstream ROI cross-modal retrieval task. 

\noindent\textbf{EnglishBook} is a self-constructed pathology image-text dataset collected from 31 English pathology textbooks. We extracted histopathology figures and their accompanying English captions using PaddleOCR, creating 2,724 image-English text pairs for evaluation. We used ChineseBook for the downstream ROI cross-modal retrieval task. 

\noindent\textbf{Atlas} is a PathMMU\cite{sun2024pathmmu} VQA source derived from pathology atlases, textbooks, and guidelines. We used the official PathMMU val, test-tiny, and test splits, and evaluated 77, 189, and 745 pairs. We used Atlas for the downstream ROI VQA task. 

\noindent\textbf{EduContent} is a PathMMU\cite{sun2024pathmmu} VQA source derived from educational pathology teaching content, including histopathology video material. We used the official PathMMu val, test-tiny, and test benchmarks, obtaining 143, 239, and 1626 evaluation pairs, respectively. We used EduContent for the downstream ROI VQA task. 

\noindent\textbf{LC25000}\cite{borkowski2019lung} is a PathMMU\cite{sun2024pathmmu} PathCLS VQA source reformulated from the lung and colon histopathology image classification dataset. The questions cover 5 diagnostic categories: benign colonic tissue, colon adenocarcinoma tissue, benign lung tissue, lung adenocarcinoma tissue, and lung squamous cell carcinoma tissue. We used the official PathMMU val, test-tiny, and test benchmarks, obtaining 10, 20, and 170 evaluation pairs, respectively. We used LC25000 for the downstream ROI VQA task. 

\noindent\textbf{Osteo}\cite{arunachalam2019viable} is a PathMMU\cite{sun2024pathmmu} PathCLS VQA source derived from osteosarcoma histopathology images. The questions cover 3 tissue categories: non-tumor tissue, necrotic tumor, and viable tumor. We used the official PathMMU val, test-tiny, and test benchmarks, obtaining 6, 12, and 102 evaluation pairs, respectively. We used Osteo for the downstream ROI VQA task. 

\noindent\textbf{SICAPv2}\cite{silva2020going} is a PathMMU\cite{sun2024pathmmu} PathCLS VQA source derived from prostate H\&E histopathology images for Gleason-pattern recognition. The questions cover 4 categories: non-cancerous tissue, Gleason grade 3, Gleason grade 4, and Gleason grade 5. We used the official PathMMU val, test-tiny, and test benchmarks, obtaining 8, 16, and 136 evaluation pairs, respectively. We used SICAPv2 for the downstream ROI VQA task. 

\noindent\textbf{SkinCancer}\cite{kriegsmann2022deep} is a PathMMU\cite{sun2024pathmmu} PathCLS VQA source derived from skin histopathology images for tissue and tumor subtype recognition. The questions use 16 candidate categories, including non-tumor skin structures and common skin tumor subtypes. We used the official PathMMU val, test-tiny, and test benchmarks, obtaining 18, 35, and 306 evaluation pairs after label filtering, respectively. We used SkinCancer for the downstream ROI VQA task. 

\noindent\textbf{WSSSLUAD}\cite{han2022wsss4luad} is a PathMMU\cite{sun2024pathmmu} PathCLS VQA source derived from lung adenocarcinoma histopathology images. The questions cover 3 categories: tumor epithelial tissue, tumor-associated stroma tissue, and normal tissue. We used the official PathMMU val, test-tiny, and test benchmarks, obtaining 6, 12, and 102 evaluation pairs, respectively. We used WSSSLUAD for the downstream ROI VQA task. 

\noindent\textbf{PubMed} is a PathMMU\cite{sun2024pathmmu} VQA source constructed from pathology image-text pairs in open-access PubMed Central scientific documents. We used the official PathMMU val, test-tiny, and test benchmarks, obtaining 224, 274, and 2,712 evaluation pairs, respectively. We used PubMed for the downstream ROI VQA task. 

\noindent\textbf{SocialPath} is a PathMMU\cite{sun2024pathmmu} VQA source constructed from pathology image-text pairs shared by pathology experts on social media. We used the official PathMMU val, test-tiny, and test benchmarks, obtaining 135, 210, and 1,466 evaluation pairs after image and label filtering, respectively. We used SocialPath for the downstream ROI VQA task. 

\noindent\textbf{BCNB}\cite{xu2021predicting} is a breast cancer WSI cohort consisting of 1,058 biopsy slides. We used BCNB for WSI biomarker prediction, WSI-text multimodal analysis and WSI fine-tuning tasks. For biomarker prediction, we used case-level label-stratified splitting to conduct experiments for ER status prediction (831 positive and 227 negative, train:val:test = 677:169:212), HER2 status prediction (277 positive and 781 negative, train:val:test = 677:169:212), and PR status prediction (790 positive and 268 negative, train:val:test = 677:169:212). For WSI-text multimodal analysis, we conducted experiments for the 2-class axillary lymph node metastasis (655 N0 and 403 N+; train:val:test = 677:169:212) and the 3-class axillary lymph node metastasis (655 N0, 210 N+(1-2), and 193 N+(>2); train:val:test = 677:169:212). We used 10 clinical variables as textual information, including age, tumor size, tumor type, ER status, PR status, HER2 status, HER2 expression score, histological grade, Ki-67 index, and molecular subtype. 

\noindent\textbf{CAMELYON+}\cite{ling2025comprehensive} is a breast cancer sentinel lymph node metastasis WSI cohort consisting of 1,349 slides. We used CAMELYON+ for WSI classification, WSI retrieval, WSI zero-shot, WSI few-shot classification, and WSI fine-tuning tasks. For classification, we used case-level label-stratified splitting to conduct experiments for the 2-class lymph node metastasis coarse-grained classification task (870 negative and 479 positive; train:val:test = 869:220:260) and the 4-class metastatic fine-grained classification task (870 negative, 54 isolated tumor cells, 174 micro metastases, and 251 macro metastases; train:val:test = 865:220:264). For WSI zero-shot classification, we used the 2-class setting. 

\noindent\textbf{CMB} is a pan-cancer primary-site WSI cohort consisting of 947 slides. We used CMB for slide-level PFM few-shot classification and WSI retrieval tasks. For classification, we used case-level label-stratified splitting to conduct experiments for the 9-class pan-cancer classification task (28 AML, 119 BRCA, 186 CRC, 32 GEC, 294 LCA, 119 MEL, 11 MML, 69 OV, and 89 PCA; train:val:test = 602:149:196). 

\noindent\textbf{CPTAC} is a pan-cancer WSI cohort covering multiple tumor types. We used CPTAC for WSI classification, WSI retrieval, WSI zero-shot classification, WSI few-shot classification, and WSI survival prediction tasks. For classification, we conducted experiments for a 9-class pan-cancer primary-site task (2,059 WSIs: BRCA 112, CCRCC 245, COAD 107, GBM 243, HNSC 259, LUAD 326, LSCC 304, OV 221, and PDA 242; train:val:test = 1313:340:406) and a 2-class non-small cell lung cancer subtyping task (326 LUAD and 304 LSCC; train:val:test = 402:108:120). For zero-shot, we used the 2-class setting. For survival prediction, we used 4 CPTAC cancer types: CCRCC (218 WSIs, 48 events), HNSC (243 WSIs, 73 events), LUAD (313 WSIs, 54 events), and PDA (227 WSIs, 167 events).  

\noindent\textbf{DHMC-RCC}\cite{zhu2021development} is a renal cell carcinoma WSI cohort consisting of 563 slides. We used DHMC-RCC for WSI classification and WSI retrieval tasks. For classification, we used case-level label-stratified splitting to conduct experiments for the 5-class renal cell carcinoma subtyping task (29 benign, 23 chromophobe renal cell carcinoma, 344 clear-cell renal cell carcinoma, 66 oncocytoma, and 101 papillary renal cell carcinoma; train:val:test = 383:23:157). 

\noindent\textbf{EBRAINS}\cite{roetzer2022digital} is a CNS tumor WSI cohort consisting of 2,310, including diagnostic, tumor family, WHO grading, and molecular annotations. We used EBRAINS for WSI classification, WSI retrieval, WSI zero-shot classification, WSI few-shot classification, and WSI biomarker prediction tasks. For classification, we case-level label-stratified splitting to conduct experiments for the 30-class CNS tumor diagnosis task (83 adamantinomatous craniopharyngioma, 47 anaplastic astrocytoma IDH-mutant, 47 anaplastic astrocytoma IDH-wildtype, 50 anaplastic ependymoma, 46 anaplastic meningioma, 91 anaplastic oligodendroglioma IDH-mutant and 1p/19q codeleted, 31 angiomatous meningioma, 82 atypical meningioma, 70 diffuse astrocytoma IDH-mutant, 59 diffuse large B-cell lymphoma of the CNS, 46 ependymoma, 57 fibrous meningioma, 88 ganglioglioma, 34 glioblastoma IDH-mutant, 469 glioblastoma IDH-wildtype, 59 gliosarcoma, 88 haemangioblastoma, 30 haemangioma, 34 haemangiopericytoma, 32 Langerhans cell histiocytosis, 38 lipoma, 32 medulloblastoma non-WNT/non-SHH, 104 meningothelial meningioma, 47 metastatic tumours, 85 oligodendroglioma IDH-mutant and 1p/19q codeleted, 172 pilocytic astrocytoma, 99 pituitary adenoma, 81 schwannoma, 41 secretory meningioma, and 68 transitional meningioma; train:val:test = 1386:353:571), the 9-class CNS tumor-family classification task (1,074 diffuse glioma, 91 hematolymphoid/histiocytic tumors, 429 meningioma, 38 mesenchymal/lipomatous tumors, 47 metastatic tumors, 81 nerve sheath tumors, 216 other neuroepithelial tumors, 182 sellar-region tumors, and 152 vascular/mesenchymal tumors; train:val:test = 1395:344:571), and the 4-class WHO grade prediction task (815 grade I, 295 grade II, 299 grade III, and 594 grade IV; train:val:test = 1206:304:493). For biomarker prediction, we used glioma IDH status prediction (327 IDH-mutant and 516 IDH-wildtype; train:val:test = 509:124:210). For WSI retrieval, we used the diagnosis, tumor family, and WHO grade settings. For WSI zero-shot classification, we used the 9-class setting, and for WSI few-shot classification, we used the 30-class CNS diagnosis setting. 

\noindent\textbf{Hancock}\cite{dorrich2025multimodal} is a head-and-neck squamous cell carcinoma WSI cohort with clinicopathological annotations, consisting of 693 slides. We used Hancock for WSI-text multimodal analysis and WSI survival prediction tasks. For multimodal analysis, we used case-level label-stratified splitting to conduct experiments for keratinizing squamous cell carcinoma grading (185 G2 and 197 G3; train:val:test = 244:60:78), lymphovascular invasion prediction (555 negative and 121 positive; train:val:test = 432:110:134), and vascular invasion prediction (632 negative and 44 positive; train:val:test = 433:109:134). For keratinizing squamous cell carcinoma grading, we used age, sex, smoking status, primary tumor site, and HPV/p16 status as textual clinical variables. For the other two task settings, we additionally used histological type and grade as two supplementary variables. For survival prediction, we used 203 WSIs with post-treatment survival records, including 52 events. 

\noindent\textbf{HunCRC}\cite{pataki2022huncrc} is a colorectal neoplasia WSI cohort consisting of 199 WSIs. We used HunCRC for WSI classification and WSI retrieval tasks. For classification, we used case-level label-stratification to conduct experiments for 4-class colorectal neoplasia screening (38 colorectal cancer, 66 adenoma, 14 negative, and 81 non-neoplastic lesions; train:val:test = 125:32:42). 

\noindent\textbf{KidRare}\cite{zhou2026knowledge} is a pediatric rare tumor WSI cohort consisting of 1,283 slides. We used KidRare for WSI classification, WSI retrieval, WSI zero-shot classification, WSI few-shot classification, and WSI fine-tuning tasks. For classification, we used case-level label-stratified splitting to conduct experiments for the 4-class pediatric rare tumor classification task (442 hepatoblastoma, 238 medulloblastoma, 467 nephroblastoma, and 136 neuroblastoma; train:val:test = 821:206:256) and the 13-class pediatric rare tumor fine-grained subtyping task (197 classic medulloblastoma, 30 desmoplastic nodular medulloblastoma, 33 differentiating neuroblastoma, 10 epithelial macrotrabecular pattern of hepatoblastoma, 177 epithelial mixed fetal and embryonal hepatoblastoma, 33 ganglioneuroblastoma intermixed, 11 large cell/anaplastic medulloblastoma, 76 mixed epithelial and mesenchymal hepatoblastoma, 51 normal, 70 poorly differentiated neuroblastoma, 179 pure fetal hepatoblastoma with low mitotic activity, 59 tumor, and 357 unknown; train:val:test = 821:206:256). For WSI zero-shot classification, we used the coarse-grained settings. 

\noindent\textbf{PANDA}\cite{bulten2022artificial} is a prostate cancer WSI cohort consisting of 9,555 WSIs. We used PANDA for WSI classification and WSI retrieval. For classification, we used case-level label-stratified splitting to conduct experiments for the 2-class prostate cancer screening (2,603 non-tumor and 6,952 tumor; train:val:test = 6115:1530:1910) and the 6-class Gleason grading (ISUP 0 to 5 with 2,603, 2,399, 1,209, 1,118, 1,124, and 1,102, respectively; train:val:test = 6121:1530:1904). 

\noindent\textbf{PTRC-HGSOC}\cite{chowdhury2023proteogenomic} is a high-grade serous ovarian carcinoma WSI cohort consisting of 348 slides with treatment response and tumor-type annotations. We used PTRC-HGSOC for predicting treatment response (202 sensitive and 146 refractory; train:val:test = 222:56:70) and tumor origin classification (174 primary and 174 metastatic; train:val:test = 226:54:68) in WSI-text multimodal analysis tasks. For tumor origin classification, we used age, race, ethnicity, histological grade, neoadjuvant treatment, tumor stage, tumor substage, and other cancer diagnosis as textual information. For treatment response, we additionally used tumor type, tumor location, and tumor location group as three supplementary variables. 

\noindent\textbf{SYSBM}\cite{zhu2023accurate} is an internal bone metastasis primary-site WSI cohort consisting of 835 slides. We used SYSBM for WSI classification, WSI retrieval, WSI zero-shot classification, and WSI few-shot classification. For classification, we used case-level label-stratified splitting to conduct experiments for 10-class bone metastasis primary site prediction (106 breast, 35 gastric, 66 kidney, 89 liver, 234 lung, 33 neuroendocrine, 17 no cancer, 54 prostate, 102 squamous carcinoma, and 99 thyroid; train:val:test = 529:135:171). 

\noindent\textbf{SYSFL+}\cite{li2025diagnostic} is an internal frozen lung tissue WSI cohort, consisting of 1,466 slides. We used SYSFL+ for WSI classification, WSI retrieval, and WSI few-shot classification. For classification, we used case-level label-stratified splitting to conduct experiments for 7-class lung cancer subtyping and invasion grading (101 atypical adenomatous hyperplasia, 307 adenocarcinoma in situ, 250 minimally invasive adenocarcinoma, 296 invasive lung adenocarcinoma, 205 lung inflammation, 116 lung squamous cell carcinoma, and 191 no carcinoma; train:val:test = 936:232:298) and 4-class lung lesion triage (658 favorable prognosis lesions, 296 invasive lung adenocarcinomas, 396 non-neoplastic lesions, and 116 squamous cell carcinomas; train:val:test = 936:232:298).

\noindent\textbf{TissueNet}\cite{fick2021partial} is a cervical epithelial lesion WSI cohort consisting of 983 slides. We used TissueNet for WSI classification and WSI retrieval. For classification, we used case-level label-stratified splitting to conduct experiments for 4-class cervical epithelial lesion screening (261 normal or subnormal tissue, 278 low-grade squamous intraepithelial lesion, 225 high-grade squamous intraepithelial lesion, and 219 invasive squamous carcinoma; train:val:test = 633:159:191). 

\noindent\textbf{UBC-OCEAN}\cite{asadi2024machine} is an ovarian carcinoma WSI cohort consisting of 538 slides. We used UBC-OCEAN for WSI classification, WSI zero-shot classification, and WSI retrieval. For classification, we used case-level label-stratified splitting to conduct experiments for ovarian carcinoma subtyping (222 high-grade serous ovarian carcinoma, 47 low-grade serous ovarian carcinoma, 99 ovarian clear cell carcinoma, 124 ovarian endometrioid carcinoma, 46 ovarian mucinous carcinoma; train:val:test = 343:85:110). 

\noindent\textbf{Visiomel} is a melanoma WSI cohort consisting of 1,203 slides. We used Visiomel for predicting relapse with (1,006 negative and 197 positive; train:val:test = 766:192:245) and without (424 negative and 122 positive; train:val:test = 350:87:109) prior melanoma in WSI-text multimodal analysis tasks. We used clinical variables as textual information, including age at initial diagnosis, sex, melanoma body site, histological type, Breslow thickness category, and ulceration.  

\noindent\textbf{MUT-HET-RCC} is a renal cell carcinoma WSI cohort consisting of 1,291 slides. We used MUT-HET-RCC for WSI biomarker prediction tasks. We used case-level label-stratified splitting to conduct experiments for BAP1 mutation (162 mutant and 1,129 wildtype; train:val:test = 825:207:259), PBRM1 mutation (669 mutant and 622 wildtype; train:val:test = 825:207:259), and SETD2 mutation prediction (348 mutant and 943 wildtype; train:val:test = 825:207:259). 

\noindent\textbf{SURGEN}\cite{myles2025surgen} is a colorectal cancer WSI cohort consisting of 1,020 slides. We use SR386 subset (140 slides) to conduct experiments of survival prediction since this subset contains survival records. 

\subsection{Evaluation settings}
We evaluated ALICE in three downstream scenarios: vision-only tasks, vision--language tasks, and slide-level tasks. All encoders were kept frozen unless explicitly stated, and the same train, validation, and test splits were used for all compared models within each task. All the downstream tasks were conducted on single 48GB NVIDIA L20 GPU.

For ROI vision-only tasks, we compared ALICE with UNI-2, Virchow-2, H-Opt-1, and GPFM. For linear probing, ROI images were encoded with frozen PFMs and L2-normalized before training a single-layer classifier with AdamW (learning rate $10^{-4}$, weight decay $10^{-4}$, batch size 256, 100 epochs). The training split was divided into five stratified folds. Four folds were used for classifier fitting and one fold for validation, with early stopping after 10 epochs without improvement. We selected the checkpoint based on validation balanced accuracy and reported test balanced accuracy as the mean and standard deviation across the five folds. For KNN classification, the training split was used as the reference set and the test split as queries. We evaluated cosine and Euclidean distances with $K\in\{5,10,20\}$ and used balanced accuracy as the primary metric. For ROI image-to-image retrieval, the training split was used as the gallery and the test split as queries. Retrieval used cosine similarity and was evaluated at $K\in\{1,3,5,10\}$ using Recall@K, and majority-vote accuracy (MVAcc@5).

For ROI few-shot classification, we used all-way ProtoNet \cite{snell2017prototypical} episodes with the training split as support and the test split as query. We evaluated $n$-shot ($n\in\{1,2,4,8,16,32,64,128,256\}$), sampled 20 query examples per class, repeated each valid setting for 500 episodes, and reported balanced accuracy as mean $\pm$ s.d. Settings were skipped when any class lacked enough support or query samples. For ROI semantic and instance segmentation, we trained a PMT decoder on top of frozen PFMs for 50 epochs using AdamW (learning rate $10^{-4}$, weight decay 0.05, batch size 8), 224-pixel crops with a stride of 112, and early stopping with patience 10. Each segmentation experiment was repeated with five random seeds (42-46). Semantic segmentation checkpoints were selected by validation mIoU and evaluated by test DICE score. Instance segmentation checkpoints were selected by validation AP@[50:95] and evaluated by test AP@50.

For ROI vision-language tasks, we compared ALICE with MUSK, KEEP, and CONCH. For zero-shot ROI classification, each class was represented by five label descriptions and ten pathology prompt templates, yielding 50 prompts per class. We evaluated two schemes: a prompt-level scheme, in which each prompt slot independently produced predictions, and a prompt-ensemble scheme, in which class probabilities were averaged across the 50 prompts. For ALICE, probabilities were additionally averaged across its KEEP, CONCH, and MUSK-aligned heads. We reported balanced accuracy for the ensemble score and mean $\pm$ s.d. across prompt-level classifiers. For ROI image-text retrieval, normalized image and text embeddings were evaluated bidirectionally on paired image-caption datasets using Recall@K with $K\in\{1,5,10,20,25,50\}$. For ROI VQA, multiple-choice PathMMU \cite{sun2024pathmmu} questions were evaluated by converting each answer option into pathology prompts and selecting the option with the highest image-text similarity. Accuracy was used as the metric. The generic ROI zero-shot templates and dataset-specific class prompts are listed in \textbf{Extended Data Table \ref{tab:roi_zero_shot_prompt_templates_general}-\ref{tab:prompt_unitopatho}}.

For WSI vision-language tasks, we evaluated MI-Zero \cite{lu2023visual}, PathPT \cite{he2026boosting}, and MICA-style multimodal analysis \cite{li2026ai}. In MI-Zero zero-shot WSI classification, each class was represented by three label texts and four WSI prompt templates, yielding 12 prompts per class. Slide-level scores were computed by Top-$K$ MIL pooling of patch-text cosine similarities with $K\in\{1,5,10,50\}$, followed by softmax averaging across prompts and, for ALICE, across teacher-aligned heads. We reported balanced accuracy on the test split. For PathPT few-shot WSI classification, we sampled $n\in\{1,5,10,15,20\}$ labeled slides per class, optimized learnable 32 context tokens and the lightweight adaptor for 20 epochs with Adam (learning rate $10^{-4}$, batch size 1), and selected checkpoints by validation balanced accuracy. Each setting was repeated over five seeds, and test balanced accuracy was reported as mean $\pm$ s.d. For MICA, clinical variables were converted into text segments and encoded by the corresponding text encoder, then input with WSI patch features through co-attention in a 256-dimensional hidden space. MICA models were trained for 20 epochs with Adam (learning rate $10^{-4}$, dropout 0.25, batch size 1, gradient accumulation over 8 steps) and class-weighted cross-entropy loss. Results were averaged over 5 seeds, using validation balanced accuracy for checkpoint selection and test balanced accuracy for reporting. The MI-Zero WSI prompt templates and class prompts are provided in \textbf{Extended Data Table \ref{tab:mizero_prompt_templates}-\ref{tab:mizero_ubc_ocean}}. Task-specific clinical text segment templates used for MICA are listed in \textbf{Extended Data Table \ref{tab:clinical_text_templates_bcnb_aln_2class}-\ref{tab:clinical_text_templates_visiomel_relapse_noprev_melanoma_2class}}.

For slide-level WSI tasks, we compared ALICE with TITAN and CARE. Patch features were extracted with frozen patch encoders and aggregated by pretrained slide encoders. TITAN and CARE used the CONCH 1.5 patch encoder, and ALICE used its own vision-only module. For slide-level KNN diagnosis, we used the training split as the reference set, the test split as queries. We reported the balanced accuracy (mean $\pm$ s.d) of $K\in\{5,10,20\}$ on both cosine and Euclidean distances. For slide-level linear probing, a single-layer classifier was trained on frozen slide embeddings with AdamW (learning rate $3\times10^{-5}$, weight decay $10^{-4}$, batch size 16, 500 epochs). The training split was stratified into five folds, checkpoints were selected on the validation fold using the prespecified validation metric, and test performance was averaged across folds. In the fine-tuning-aligned protocol, the official train and validation splits were used, five seeds were run, and test balanced accuracy was reported as mean $\pm$ s.d. Biomarker prediction used the same linear-probe protocol and was evaluated by AUC. Slide-to-slide retrieval used cosine similarity, with the training split as the gallery and the test split as queries, and was evaluated with MVAcc@3. WSI few-shot classification used the same linear-probe protocol on $n$-shot subsets with $n\in\{1,2,4,8,16,32,64\}$. For slide-level fine-tuning, the slide encoder and classifier were optimized jointly for 50 epochs with Adam (learning rate $3\times10^{-5}$, weight decay $10^{-4}$, batch size 1), early stopping with a patience of 5, and validation balanced accuracy for checkpoint selection. Experiments were repeated over five seeds. For survival prediction, slide embeddings were evaluated in a cross-validation setting using Cox proportional hazards models, and performance was summarized by the C-index.

\section{Data availability}
The pretraining dataset TCGA-12K used for the vision-only and multimodal stages can be accessed at \url{https://huggingface.co/datasets/medarc/TCGA-12K-parquet}. The pretraining dataset TCGA-UT-8K used for the slide-level stage can be accessed at \url{https://huggingface.co/datasets/MahmoodLab/TCGA-UniformTumor-8K}. The pretraining dataset HISTAI used for the slide-level stage can be accessed at \url{https://huggingface.co/collections/histai}.

For downstream benchmarks, all publicly available datasets can be accessed through their original data portals or publications, including AGGC2022 (\url{https://aggc22.grand-challenge.org/}), BRACS (\url{https://www.bracs.icar.cnr.it/}), BreakHis (\url{https://web.inf.ufpr.br/vri/databases/breast-cancer-histopathological-database-breakhis/}), CCRCC (\url{https://zenodo.org/records/7898308}), Chaoyang (\url{https://bupt-ai-cz.github.io/HSA-NRL/}), CRC-100K (\url{https://zenodo.org/records/1214456}), CRC-MSI (\url{https://zenodo.org/records/2530835}), ESCA-Tolkach (\url{https://zenodo.org/records/7548828}), GCHTID (\url{https://www.kaggle.com/datasets/orvile/gastric-cancer-histopathology-tissue-image-dataset/data}), OTA (\url{https://www.cancerimagingarchive.net/collection/osteosarcoma-tumor-assessment/}), Unitopatho (\url{https://ieee-dataport.org/open-access/unitopatho}), CoNSeP (\url{https://www.kaggle.com/datasets/rftexas/tiled-consep-224x224px}), CRAG (\url{https://warwick.ac.uk/fac/cross_fac/tia/data/mildnet/}), GlaS (\url{https://www.kaggle.com/datasets/sani84/glasmiccai2015-gland-segmentation}), CoCaHis (\url{https://cocahis.irb.hr/}), COSAS24 (\url{https://cosas.grand-challenge.org/}), EBHI (\url{https://www.kaggle.com/datasets/alibabaei78/ebhi-seg}), Janowczyk (\url{https://andrewjanowczyk.com/use-case-1-nuclei-segmentation/}), BookSet (\url{https://warwick.ac.uk/fac/cross_fac/tia/data/arch/}), PathMMU (\url{https://huggingface.co/datasets/jamessyx/PathMMU}), BCNB (\url{https://bupt-ai-cz.github.io/BCNB/}), CAMELYON+ (\url{https://www.scidb.cn/en/detail?dataSetId=cc1f911b75ca4610bd02ac33a51898a9}), CMB (\url{https://www.cancerimagingarchive.net/research/cmb/}), CPTAC (\url{https://www.cancerimagingarchive.net/collection}), DHMC-RCC (\url{https://bmirds.github.io/KidneyCancer/}), EBRAINS (\url{https://doi.org/10.25493/WQ48-ZGX}), HunCRC (\url{https://www.cancerimagingarchive.net/collection/hungarian-colorectal-screening/}), KidRare (\url{https://huggingface.co/datasets/Firehdx233/KidRare}), PANDA (\url{https://panda.grand-challenge.org/data/}), PTRC-HGSOC (\url{https://www.cancerimagingarchive.net/collection/ptrc-hgsoc/}), TissueNet (\url{https://www.drivendata.org/competitions/67/competition-cervical-biopsy/}), UBC-OCEAN (\url{https://www.kaggle.com/competitions/UBC-OCEAN/}), Visiomel (\url{https://www.drivendata.org/competitions/148/visiomel-melanoma/}), MUT-HET-RCC (\url{https://doi.org/10.25452/figshare.plus.c.5983795}) and SURGEN (\url{https://www.ebi.ac.uk/biostudies/bioimages/studies/S-BIAD1285}).

\section{Code availability}
The code and model weight for ALICE can be accessed at \url{https://github.com/WonderLandxD/ALICE}.

\section{Author contributions}
J.L., T.G., H.S., A.H., C.H., and Y.H. conceived and designed the study. J.L., T.G., and H.S. developed the ALICE framework. J.L. and T.G. implemented the model training, inference, and evaluation pipelines. J.L., H.S., X.L., and M.F. curated and processed the pretraining and downstream evaluation datasets. J.L., T.G., X.L., and M.F. performed the benchmark experiments and statistical analyses. H.S. and A.H. provided pathology expertise and contributed to clinical interpretation. J.L. and X.L. prepared the figures and tables. J.L., T.G., and H.S. wrote the initial manuscript draft. A.H., C.H., and Y.H. supervised the study and revised the manuscript. Y.H. acquired funding. All authors reviewed and approved the final manuscript.

\section{Competing interests}
The authors declare no competing interests.

\section{Acknowledgements}
This work was supported by the National Natural Science Foundation of China (NSFC) (82430062), the Beijing Municipal Health Commission Research Ward Excellence Clinical Research Program (BRWEP2024WO32240114), the Shenzhen Engineering Research Centre (XMHT20230115004), and the National High Level Hospital Clinical Research Funding (BJ-2024-155).

\newpage

\noindent{\bfseries References}\setlength{\parskip}{12pt}%
\bibliography{ref_main}

@article{song2023artificial,
  title={Artificial intelligence for digital and computational pathology},
  author={Song, Andrew H and Jaume, Guillaume and Williamson, Drew FK and Lu, Ming Y and Vaidya, Anurag and Miller, Tiffany R and Mahmood, Faisal},
  journal={Nature Reviews Bioengineering},
  volume={1},
  number={12},
  pages={930--949},
  year={2023},
  publisher={Nature Publishing Group UK London}
}

@article{zheng2026lazyslide,
  title={LazySlide: accessible and interoperable whole-slide image analysis},
  author={Zheng, Yimin and Abila, Ernesto and Chrenkov{\'a}, Eva and Buljan, Iva and Winkler, Juliane and Rendeiro, Andr{\'e} F},
  journal={Nature Methods},
  pages={1--4},
  year={2026},
  publisher={Nature Publishing Group US New York}
}

@article{lu2021data,
  title={Data-efficient and weakly supervised computational pathology on whole-slide images},
  author={Lu, Ming Y and Williamson, Drew FK and Chen, Tiffany Y and Chen, Richard J and Barbieri, Matteo and Mahmood, Faisal},
  journal={Nature Biomedical Engineering},
  volume={5},
  number={6},
  pages={555--570},
  year={2021},
  publisher={Nature Publishing Group UK London}
}

@article{jiang2024transformer,
  title={A transformer-based weakly supervised computational pathology method for clinical-grade diagnosis and molecular marker discovery of gliomas},
  author={Jiang, Rui and Yin, Xiaoxu and Yang, Pengshuai and Cheng, Lingchao and Hu, Juan and Yang, Jiao and Wang, Ying and Fu, Xiaodan and Shang, Li and Li, Liling and others},
  journal={Nature Machine Intelligence},
  volume={6},
  number={8},
  pages={876--891},
  year={2024},
  publisher={Nature Publishing Group UK London}
}

@article{campanella2019clinical,
  title={Clinical-grade computational pathology using weakly supervised deep learning on whole slide images},
  author={Campanella, Gabriele and Hanna, Matthew G and Geneslaw, Luke and Miraflor, Allen and Werneck Krauss Silva, Vitor and Busam, Klaus J and Brogi, Edi and Reuter, Victor E and Klimstra, David S and Fuchs, Thomas J},
  journal={Nature Medicine},
  volume={25},
  number={8},
  pages={1301--1309},
  year={2019},
  publisher={Nature Publishing Group US New York}
}

@article{tsai2023histopathology,
  title={Histopathology images predict multi-omics aberrations and prognoses in colorectal cancer patients},
  author={Tsai, Pei-Chen and Lee, Tsung-Hua and Kuo, Kun-Chi and Su, Fang-Yi and Lee, Tsung-Lu Michael and Marostica, Eliana and Ugai, Tomotaka and Zhao, Melissa and Lau, Mai Chan and V{\"a}yrynen, Juha P and others},
  journal={Nature Communications},
  volume={14},
  number={1},
  pages={2102},
  year={2023},
  publisher={Nature Publishing Group UK London}
}

@article{zhu2023accurate,
  title={An accurate prediction of the origin for bone metastatic cancer using deep learning on digital pathological images},
  author={Zhu, Lianghui and Shi, Huijuan and Wei, Huiting and Wang, Chengjiang and Shi, Shanshan and Zhang, Fenfen and Yan, Renao and Liu, Yiqing and He, Tingting and Wang, Liyuan and others},
  journal={EBioMedicine},
  volume={87},
  year={2023},
  publisher={Elsevier}
}

@article{chen2024towards,
  title={Towards a general-purpose foundation model for computational pathology},
  author={Chen, Richard J and Ding, Tong and Lu, Ming Y and Williamson, Drew FK and Jaume, Guillaume and Song, Andrew H and Chen, Bowen and Zhang, Andrew and Shao, Daniel and Shaban, Muhammad and others},
  journal={Nature Medicine},
  volume={30},
  number={3},
  pages={850--862},
  year={2024},
  publisher={Nature Publishing Group US New York}
}

@article{lu2024visual,
  title={A visual-language foundation model for computational pathology},
  author={Lu, Ming Y and Chen, Bowen and Williamson, Drew FK and Chen, Richard J and Liang, Ivy and Ding, Tong and Jaume, Guillaume and Odintsov, Igor and Le, Long Phi and Gerber, Georg and others},
  journal={Nature Medicine},
  volume={30},
  number={3},
  pages={863--874},
  year={2024},
  publisher={Nature Publishing Group US New York}
}

@article{xu2024whole,
  title={A whole-slide foundation model for digital pathology from real-world data},
  author={Xu, Hanwen and Usuyama, Naoto and Bagga, Jaspreet and Zhang, Sheng and Rao, Rajesh and Naumann, Tristan and Wong, Cliff and Gero, Zelalem and Gonz{\'a}lez, Javier and Gu, Yu and others},
  journal={Nature},
  volume={630},
  number={8015},
  pages={181--188},
  year={2024},
  publisher={Nature Publishing Group UK London}
}

@article{yan2025pathorchestra,
  title={Pathorchestra: A comprehensive foundation model for computational pathology with over 100 diverse clinical-grade tasks},
  author={Yan, Fang and Wu, Jianfeng and Li, Jiawen and Wang, Wei and Chen, Yirong and Wei, Linda and Lu, Jiaxuan and Chen, Wen and Gao, Zizhao and Li, Jianan and others},
  journal={npj Digital Medicine},
  volume={8},
  number={1},
  pages={695},
  year={2025},
  publisher={Nature Publishing Group UK London}
}

@article{vorontsov2024foundation,
  title={A foundation model for clinical-grade computational pathology and rare cancers detection},
  author={Vorontsov, Eugene and Bozkurt, Alican and Casson, Adam and Shaikovski, George and Zelechowski, Michal and Severson, Kristen and Zimmermann, Eric and Hall, James and Tenenholtz, Neil and Fusi, Nicolo and others},
  journal={Nature Medicine},
  volume={30},
  number={10},
  pages={2924--2935},
  year={2024},
  publisher={Nature Publishing Group US New York}
}

@article{xiang2025vision,
  title={A vision--language foundation model for precision oncology},
  author={Xiang, Jinxi and Wang, Xiyue and Zhang, Xiaoming and Xi, Yinghua and Eweje, Feyisope and Chen, Yijiang and Li, Yuchen and Bergstrom, Colin and Gopaulchan, Matthew and Kim, Ted and others},
  journal={Nature},
  volume={638},
  number={8051},
  pages={769--778},
  year={2025},
  publisher={Nature Publishing Group UK London}
}

@article{ding2025multimodal,
  title={A multimodal whole-slide foundation model for pathology},
  author={Ding, Tong and Wagner, Sophia J and Song, Andrew H and Chen, Richard J and Lu, Ming Y and Zhang, Andrew and Vaidya, Anurag J and Jaume, Guillaume and Shaban, Muhammad and Kim, Ahrong and others},
  journal={Nature Medicine},
  pages={1--13},
  year={2025},
  publisher={Nature Publishing Group US New York}
}

@article{neidlinger2025benchmarking,
  title={Benchmarking foundation models as feature extractors for weakly supervised computational pathology},
  author={Neidlinger, Peter and El Nahhas, Omar SM and Muti, Hannah Sophie and Lenz, Tim and Hoffmeister, Michael and Brenner, Hermann and van Treeck, Marko and Langer, Rupert and Dislich, Bastian and Behrens, Hans Michael and others},
  journal={Nature Biomedical Engineering},
  pages={1--11},
  year={2025},
  publisher={Nature Publishing Group UK London}
}

@article{campanella2025clinical,
  title={A clinical benchmark of public self-supervised pathology foundation models},
  author={Campanella, Gabriele and Chen, Shengjia and Singh, Manbir and Verma, Ruchika and Muehlstedt, Silke and Zeng, Jennifer and Stock, Aryeh and Croken, Matt and Veremis, Brandon and Elmas, Abdulkadir and others},
  journal={Nature Communications},
  volume={16},
  number={1},
  pages={3640},
  year={2025},
  publisher={Nature Publishing Group UK London}
}

@inproceedings{heinrich2025radiov2,
  title={Radiov2.5: Improved baselines for agglomerative vision foundation models},
  author={Heinrich, Greg and Ranzinger, Mike and Yin, Hongxu and Lu, Yao and Kautz, Jan and Tao, Andrew and Catanzaro, Bryan and Molchanov, Pavlo},
  booktitle={Proceedings of the Computer Vision and Pattern Recognition Conference},
  pages={22487--22497},
  year={2025}
}

@article{zhu2026efficient,
  title={Efficient Universal Perception Encoder},
  author={Zhu, Chenchen and Suri, Saksham and Jose, Cijo and Oquab, Maxime and Szafraniec, Marc and Wen, Wei and Xiong, Yunyang and Labatut, Patrick and Bojanowski, Piotr and Krishnamoorthi, Raghuraman and others},
  journal={arXiv preprint arXiv:2603.22387},
  year={2026}
}

@article{ma2026generalizable,
  title={A generalizable pathology foundation model using a unified knowledge distillation pretraining framework},
  author={Ma, Jiabo and Guo, Zhengrui and Zhou, Fengtao and Wang, Yihui and Xu, Yingxue and Li, Jinbang and Yan, Fang and Cai, Yu and Zhu, Zhengjie and Jin, Cheng and others},
  journal={Nature Biomedical Engineering},
  volume={10},
  number={3},
  pages={545--564},
  year={2026},
  publisher={Nature Publishing Group UK London}
}

@article{zimmermann2024virchow2,
  title={Virchow2: Scaling self-supervised mixed magnification models in pathology},
  author={Zimmermann, Eric and Vorontsov, Eugene and Viret, Julian and Casson, Adam and Zelechowski, Michal and Shaikovski, George and Tenenholtz, Neil and Hall, James and Klimstra, David and Yousfi, Razik and others},
  journal={arXiv preprint arXiv:2408.00738},
  year={2024}
}

@article{scalbert2026abstract,
  title={Abstract LB174: H-optimus-1: A foundation model for computational histopathology},
  author={Scalbert, Marin and Saillard, Charlie and Peeters, Thomas and Gonzalez, Liam and Valter, Dasha and Llinares-L{\'o}pez, Felipe and Mariet, Zelda E and Jenatton, Rodolphe},
  journal={Cancer Research},
  volume={86},
  number={8\_Supplement},
  pages={LB174--LB174},
  year={2026},
  publisher={American Association for Cancer Research}
}

@article{zhou2026knowledge,
  title={Knowledge-enhanced pretraining for vision-language pathology foundation model on cancer diagnosis},
  author={Zhou, Xiao and Sun, Luoyi and He, Dexuan and Guan, Wenbin and Wang, Ge and Wang, Ruifen and Wang, Lifeng and Yuan, Xiaojun and Sun, Xin and Zhang, Ya and others},
  journal={Cancer Cell},
  volume={44},
  number={4},
  pages={777--791},
  year={2026},
  publisher={Elsevier}
}

@inproceedings{zhang2026care,
  title={CARE: A Molecular-Guided Foundation Model with Adaptive Region Modeling for Whole Slide Image Analysis},
  author={Zhang, Di and Gong, Zhangpeng and Pang, Xiaobo and Liu, Jiashuai and Lu, Junbo and Cui, Hao and Ge, Jiusong and Zeng, Zhi and Yi, Kai and Li, Yinghua and others},
  booktitle={Proceedings of the IEEE/CVF Conference on Computer Vision and Pattern Recognition},
  pages={21078--21088},
  year={2026}
}

@inproceedings{ilse2018attention,
  title={Attention-based deep multiple instance learning},
  author={Ilse, Maximilian and Tomczak, Jakub and Welling, Max},
  booktitle={International conference on machine learning},
  pages={2127--2136},
  year={2018},
  organization={PMLR}
}

@article{acosta2022multimodal,
  title={Multimodal biomedical AI},
  author={Acosta, Juli{\'a}n N and Falcone, Guido J and Rajpurkar, Pranav and Topol, Eric J},
  journal={Nature medicine},
  volume={28},
  number={9},
  pages={1773--1784},
  year={2022},
  publisher={Nature Publishing Group US New York}
}

@article{he2026boosting,
  title={Boosting pathology foundation models via few-shot prompt-tuning for rare cancer subtyping},
  author={He, Dexuan and Zhou, Xiao and Guan, Wenbin and Zhang, Liyuan and Zhang, Xiaoman and Xu, Sinuo and Wang, Ge and Wang, Lifeng and Yuan, Xiaojun and Ma, Jing and others},
  journal={Nature Communications},
  year={2026},
  publisher={Nature Publishing Group UK London}
}

@article{li2026ai,
  title={AI-enabled virtual spatial proteomics from histopathology for interpretable biomarker discovery in lung cancer},
  author={Li, Zhe and Li, Yuchen and Xiang, Jinxi and Wang, Xiyue and Yang, Sen and Zhang, Xiaoming and Eweje, Feyisope and Chen, Yijiang and Luo, Xiangde and Li, Yuanyuan and others},
  journal={Nature Medicine},
  pages={1--14},
  year={2026},
  publisher={Nature Publishing Group US New York}
}

@article{dorrich2025multimodal,
  title={A multimodal dataset for precision oncology in head and neck cancer},
  author={D{\"o}rrich, Marion and Balk, Matthias and Heusinger, Tatjana and Beyer, Sandra and Mirbagheri, Hamed and Fischer, David J and Kanso, Hassan and Matek, Christian and Hartmann, Arndt and Iro, Heinrich and others},
  journal={Nature Communications},
  volume={16},
  number={1},
  pages={7163},
  year={2025},
  publisher={Nature Publishing Group UK London}
}

@article{chowdhury2023proteogenomic,
  title={Proteogenomic analysis of chemo-refractory high-grade serous ovarian cancer},
  author={Chowdhury, Shrabanti and Kennedy, Jacob J and Ivey, Richard G and Murillo, Oscar D and Hosseini, Noshad and Song, Xiaoyu and Petralia, Francesca and Calinawan, Anna and Savage, Sara R and Berry, Anna B and others},
  journal={Cell},
  volume={186},
  number={16},
  pages={3476--3498},
  year={2023},
  publisher={Elsevier}
}

@article{wang2024pathology,
  title={A pathology foundation model for cancer diagnosis and prognosis prediction},
  author={Wang, Xiyue and Zhao, Junhan and Marostica, Eliana and Yuan, Wei and Jin, Jietian and Zhang, Jiayu and Li, Ruijiang and Tang, Hongping and Wang, Kanran and Li, Yu and others},
  journal={Nature},
  volume={634},
  number={8035},
  pages={970--978},
  year={2024},
  publisher={Nature Publishing Group UK London}
}

@article{xu2025multimodal,
  title={A multimodal knowledge-enhanced whole-slide pathology foundation model},
  author={Xu, Yingxue and Wang, Yihui and Zhou, Fengtao and Ma, Jiabo and Jin, Cheng and Yang, Shu and Li, Jinbang and Zhang, Zhengyu and Zhao, Chenglong and Zhou, Huajun and others},
  journal={Nature Communications},
  year={2025},
  publisher={Nature Publishing Group UK London}
}

@article{nechaev2025histai,
  title={Histai: an open-source, large-scale whole slide image dataset for computational pathology},
  author={Nechaev, Dmitry and Pchelnikov, Alexey and Ivanova, Ekaterina},
  journal={arXiv preprint arXiv:2505.12120},
  year={2025}
}

@article{vaidya2024demographic,
  title={Demographic bias in misdiagnosis by computational pathology models},
  author={Vaidya, Anurag and Chen, Richard J and Williamson, Drew FK and Song, Andrew H and Jaume, Guillaume and Yang, Yuzhe and Hartvigsen, Thomas and Dyer, Emma C and Lu, Ming Y and Lipkova, Jana and others},
  journal={Nature Medicine},
  volume={30},
  number={4},
  pages={1174--1190},
  year={2024},
  publisher={Nature Publishing Group US New York}
}

@inproceedings{li2026stainnet,
  title={StainNet: Scaling Self-Supervised Foundation Models on Immunohistochemistry and Special Stains for Computational Pathology},
  author={Li, Jiawen and Hu, Jiali and Ling, Xitong and Lv, Yongqiang and Chen, Yuxuan and Wang, Yizhi and Guan, Tian and Liu, Yifei and He, Yonghong},
  booktitle={Medical Imaging with Deep Learning},
  pages={544--569},
  year={2026},
  organization={PMLR}
}

@inproceedings{karasikov2025training,
  title={Training state-of-the-art pathology foundation models with orders of magnitude less data},
  author={Karasikov, Mikhail and van Doorn, Joost and K{\"a}nzig, Nicolas and Erdal Cesur, Melis and Horlings, Hugo Mark and Berke, Robert and Tang, Fei and Ot{\'a}lora, Sebastian},
  booktitle={International Conference on Medical Image Computing and Computer-Assisted Intervention},
  pages={573--583},
  year={2025},
  organization={Springer}
}

@article{oquab2023dinov2,
  title={Dinov2: Learning robust visual features without supervision},
  author={Oquab, Maxime and Darcet, Timoth{\'e}e and Moutakanni, Th{\'e}o and Vo, Huy and Szafraniec, Marc and Khalidov, Vasil and Fernandez, Pierre and Haziza, Daniel and Massa, Francisco and El-Nouby, Alaaeldin and others},
  journal={arXiv preprint arXiv:2304.07193},
  year={2023}
}

@article{chu2021conditional,
  title={Conditional positional encodings for vision transformers},
  author={Chu, Xiangxiang and Tian, Zhi and Zhang, Bo and Wang, Xinlong and Shen, Chunhua},
  journal={arXiv preprint arXiv:2102.10882},
  year={2021}
}

@inproceedings{darcet2024vision,
  title={Vision transformers need registers},
  author={Darcet, Timoth{\'e}e and Oquab, Maxime and Mairal, Julien and Bojanowski, Piotr},
  booktitle={International conference on learning representations},
  volume={2024},
  pages={2632--2652},
  year={2024}
}

@article{press2021train,
  title={Train short, test long: Attention with linear biases enables input length extrapolation},
  author={Press, Ofir and Smith, Noah A and Lewis, Mike},
  journal={arXiv preprint arXiv:2108.12409},
  year={2021}
}

@article{huo2024comprehensive,
  title={A comprehensive AI model development framework for consistent Gleason grading},
  author={Huo, Xinmi and Ong, Kok Haur and Lau, Kah Weng and Gole, Laurent and Young, David M and Tan, Char Loo and Zhu, Xiaohui and Zhang, Chongchong and Zhang, Yonghui and Li, Longjie and others},
  journal={Communications Medicine},
  volume={4},
  number={1},
  pages={84},
  year={2024},
  publisher={Nature Publishing Group UK London}
}

@article{brancati2022bracs,
  title={Bracs: A dataset for breast carcinoma subtyping in h\&e histology images},
  author={Brancati, Nadia and Anniciello, Anna Maria and Pati, Pushpak and Riccio, Daniel and Scognamiglio, Giosu{\`e} and Jaume, Guillaume and De Pietro, Giuseppe and Di Bonito, Maurizio and Foncubierta, Antonio and Botti, Gerardo and others},
  journal={Database},
  volume={2022},
  pages={baac093},
  year={2022},
  publisher={Oxford University Press UK}
}

@article{spanhol2015dataset,
  title={A dataset for breast cancer histopathological image classification},
  author={Spanhol, Fabio A and Oliveira, Luiz S and Petitjean, Caroline and Heutte, Laurent},
  journal={IEEE Transactions on Biomedical Engineering},
  volume={63},
  number={7},
  pages={1455--1462},
  year={2015},
  publisher={IEEE}
}

@article{brummer2023computational,
  title={Computational textural mapping harmonises sampling variation and reveals multidimensional histopathological fingerprints},
  author={Brummer, Otso and P{\"o}l{\"o}nen, Petri and Mustjoki, Satu and Br{\"u}ck, Oscar},
  journal={British Journal of Cancer},
  volume={129},
  number={4},
  pages={683--695},
  year={2023},
  publisher={Nature Publishing Group UK London}
}

@article{zhu2021hard,
  title={Hard sample aware noise robust learning for histopathology image classification},
  author={Zhu, Chuang and Chen, Wenkai and Peng, Ting and Wang, Ying and Jin, Mulan},
  journal={IEEE Transactions on Medical Imaging},
  volume={41},
  number={4},
  pages={881--894},
  year={2021},
  publisher={IEEE}
}

@dataset{kather_jakob_nikolas_2018_1214456,
  author       = {Kather, Jakob Nikolas and Halama, Niels and Marx, Alexander},
  title        = {{100,000 histological images of human colorectal cancer and healthy tissue}},
  month        = apr,
  year         = 2018,
  publisher    = {Zenodo},
  version      = {v0.1},
  doi          = {10.5281/zenodo.1214456},
  url          = {https://doi.org/10.5281/zenodo.1214456}
}

@article{kather2019deep,
  title={Deep learning can predict microsatellite instability directly from histology in gastrointestinal cancer},
  author={Kather, Jakob Nikolas and Pearson, Alexander T and Halama, Niels and J{\"a}ger, Dirk and Krause, Jeremias and Loosen, Sven H and Marx, Alexander and Boor, Peter and Tacke, Frank and Neumann, Ulf Peter and others},
  journal={Nature medicine},
  volume={25},
  number={7},
  pages={1054--1056},
  year={2019},
  publisher={Nature Publishing Group US New York}
}

@article{tolkach2023artificial,
  title={Artificial intelligence for tumour tissue detection and histological regression grading in oesophageal adenocarcinomas: a retrospective algorithm development and validation study},
  author={Tolkach, Yuri and Wolgast, Lisa Marie and Damanakis, Alexander and Pryalukhin, Alexey and Schallenberg, Simon and Hulla, Wolfgang and Eich, Marie-Lisa and Schroeder, Wolfgang and Mukhopadhyay, Anirban and Fuchs, Moritz and others},
  journal={The Lancet Digital Health},
  volume={5},
  number={5},
  pages={e265--e275},
  year={2023},
  publisher={Elsevier}
}

@article{lou2025large,
  title={A large histological images dataset of gastric cancer with tumour microenvironment annotation for AI},
  author={Lou, Shenghan and Ji, Jianxin and Li, Huiying and Zhang, Xuan and Jiang, Yang and Hua, Menglei and Chen, Kexin and Ge, Kaiyuan and Zhang, Qi and Wang, Liuying and others},
  journal={Scientific Data},
  volume={12},
  number={1},
  pages={138},
  year={2025},
  publisher={Nature Publishing Group UK London}
}

@article{leavey2019osteosarcoma,
  title={Osteosarcoma data from ut southwestern/ut dallas for viable and necrotic tumor assessment [data set]},
  author={Leavey, Patrick and Sengupta, Anita and Rakheja, Dinesh and Daescu, Ovidiu and Arunachalam, Harish B and Mishra, Rashika},
  journal={Cancer Imaging Arch},
  volume={14},
  year={2019}
}

@inproceedings{barbano2021unitopatho,
  title={Unitopatho, a labeled histopathological dataset for colorectal polyps classification and adenoma dysplasia grading},
  author={Barbano, Carlo Alberto and Perlo, Daniele and Tartaglione, Enzo and Fiandrotti, Attilio and Bertero, Luca and Cassoni, Paola and Grangetto, Marco},
  booktitle={2021 IEEE International Conference on Image Processing (ICIP)},
  pages={76--80},
  year={2021},
  organization={IEEE}
}

@article{graham2019hover,
  title={Hover-net: Simultaneous segmentation and classification of nuclei in multi-tissue histology images},
  author={Graham, Simon and Vu, Quoc Dang and Raza, Shan E Ahmed and Azam, Ayesha and Tsang, Yee Wah and Kwak, Jin Tae and Rajpoot, Nasir},
  journal={Medical Image Analysis},
  volume={58},
  pages={101563},
  year={2019},
  publisher={Elsevier}
}

@article{graham2019mild,
  title={MILD-Net: Minimal information loss dilated network for gland instance segmentation in colon histology images},
  author={Graham, Simon and Chen, Hao and Gamper, Jevgenij and Dou, Qi and Heng, Pheng-Ann and Snead, David and Tsang, Yee Wah and Rajpoot, Nasir},
  journal={Medical Image Analysis},
  volume={52},
  pages={199--211},
  year={2019},
  publisher={Elsevier}
}

@article{sirinukunwattana2017gland,
  title={Gland segmentation in colon histology images: The glas challenge contest},
  author={Sirinukunwattana, Korsuk and Pluim, Josien PW and Chen, Hao and Qi, Xiaojuan and Heng, Pheng-Ann and Guo, Yun Bo and Wang, Li Yang and Matuszewski, Bogdan J and Bruni, Elia and Sanchez, Urko and others},
  journal={Medical Image Analysis},
  volume={35},
  pages={489--502},
  year={2017},
  publisher={Elsevier}
}

@article{sitnik2021dataset,
  title={A dataset and a methodology for intraoperative computer-aided diagnosis of a metastatic colon cancer in a liver},
  author={Sitnik, Dario and Aralica, Gorana and Had{\v{z}}ija, Mirko and Had{\v{z}}ija, Marijana Popovi{\'c} and Pa{\v{c}}i{\'c}, Arijana and Peri{\v{s}}a, Marija Milkovi{\'c} and Manojlovi{\'c}, Luka and Krstanac, Karolina and Plaveti{\'c}, Andrija and Kopriva, Ivica},
  journal={Biomedical Signal Processing and Control},
  volume={66},
  pages={102402},
  year={2021},
  publisher={Elsevier}
}

@article{hu2023ebhi,
  title={EBHI: A new enteroscope biopsy histopathological H\&E image dataset for image classification evaluation},
  author={Hu, Weiming and Li, Chen and Rahaman, Md Mamunur and Chen, Haoyuan and Liu, Wanli and Yao, Yudong and Sun, Hongzan and Grzegorzek, Marcin and Li, Xiaoyan},
  journal={Physica Medica},
  volume={107},
  pages={102534},
  year={2023},
  publisher={Elsevier}
}

@article{janowczyk2016deep,
  title={Deep learning for digital pathology image analysis: A comprehensive tutorial with selected use cases},
  author={Janowczyk, Andrew and Madabhushi, Anant},
  journal={Journal of Pathology Informatics},
  volume={7},
  number={1},
  pages={29},
  year={2016},
  publisher={Elsevier}
}

@inproceedings{gamper2020multiple,
  title={Multiple Instance Captioning: Learning Representations from 
Histopathology Textbooks and Articles},
  author={Gamper, Jevgenij and Rajpoot, Nasir},
  booktitle={Proceedings of the IEEE Conference on Computer Vision and Pattern Recognition},
  year={2021}
}

@inproceedings{sun2024pathmmu,
  title={Pathmmu: A massive multimodal expert-level benchmark for understanding and reasoning in pathology},
  author={Sun, Yuxuan and Wu, Hao and Zhu, Chenglu and Zheng, Sunyi and Chen, Qizi and Zhang, Kai and Zhang, Yunlong and Wan, Dan and Lan, Xiaoxiao and Zheng, Mengyue and others},
  booktitle={European Conference on Computer Vision},
  pages={56--73},
  year={2024},
  organization={Springer}
}

@article{borkowski2019lung,
  title={Lung and colon cancer histopathological image dataset (lc25000)},
  author={Borkowski, Andrew A and Bui, Marilyn M and Thomas, L Brannon and Wilson, Catherine P and DeLand, Lauren A and Mastorides, Stephen M},
  journal={arXiv preprint arXiv:1912.12142},
  year={2019}
}

@article{arunachalam2019viable,
  title={Viable and necrotic tumor assessment from whole slide images of osteosarcoma using machine-learning and deep-learning models},
  author={Arunachalam, Harish Babu and Mishra, Rashika and Daescu, Ovidiu and Cederberg, Kevin and Rakheja, Dinesh and Sengupta, Anita and Leonard, David and Hallac, Rami and Leavey, Patrick},
  journal={PloS One},
  volume={14},
  number={4},
  pages={e0210706},
  year={2019},
  publisher={Public Library of Science San Francisco, CA USA}
}

@article{silva2020going,
  title={Going deeper through the Gleason scoring scale: An automatic end-to-end system for histology prostate grading and cribriform pattern detection},
  author={Silva-Rodr{\'\i}guez, Julio and Colomer, Adri{\'a}n and Sales, Mar{\'\i}a A and Molina, Rafael and Naranjo, Valery},
  journal={Computer Methods and Programs in Biomedicine},
  volume={195},
  pages={105637},
  year={2020},
  publisher={Elsevier}
}

@article{kriegsmann2022deep,
  title={Deep learning for the detection of anatomical tissue structures and neoplasms of the skin on scanned histopathological tissue sections},
  author={Kriegsmann, Katharina and Lobers, Frithjof and Zgorzelski, Christiane and Kriegsmann, Joerg and Janssen, Charlotte and Meli{\ss}, Rolf R{\"u}dinger and Muley, Thomas and Sack, Ulrich and Steinbuss, Georg and Kriegsmann, Mark},
  journal={Frontiers in Oncology},
  volume={12},
  pages={1022967},
  year={2022}
}

@article{han2022wsss4luad,
  title={Wsss4luad: Grand challenge on weakly-supervised tissue semantic segmentation for lung adenocarcinoma},
  author={Han, Chu and Pan, Xipeng and Yan, Lixu and Lin, Huan and Li, Bingbing and Yao, Su and Lv, Shanshan and Shi, Zhenwei and Mai, Jinhai and Lin, Jiatai and others},
  journal={arXiv preprint arXiv:2204.06455},
  year={2022}
}

@article{xu2021predicting,
  title={Predicting axillary lymph node metastasis in early breast cancer using deep learning on primary tumor biopsy slides},
  author={Xu, Feng and Zhu, Chuang and Tang, Wenqi and Wang, Ying and Zhang, Yu and Li, Jie and Jiang, Hongchuan and Shi, Zhongyue and Liu, Jun and Jin, Mulan},
  journal={Frontiers in Oncology},
  volume={11},
  pages={759007},
  year={2021},
  publisher={Frontiers}
}

@article{ling2025comprehensive,
  title={Comprehensive benchmark dataset for pathological lymph node metastasis in breast cancer sections},
  author={Ling, Xitong and Lei, Yuanyuan and Li, Jiawen and Cheng, Junru and Huang, Wenting and Guan, Tian and Guan, Jian and He, Yonghong},
  journal={Scientific Data},
  volume={12},
  number={1},
  pages={1381},
  year={2025},
  publisher={Nature Publishing Group UK London}
}

@article{zhu2021development,
  title={Development and evaluation of a deep neural network for histologic classification of renal cell carcinoma on biopsy and surgical resection slides},
  author={Zhu, Mengdan and Ren, Bing and Richards, Ryland and Suriawinata, Matthew and Tomita, Naofumi and Hassanpour, Saeed},
  journal={Scientific Reports},
  volume={11},
  number={1},
  pages={7080},
  year={2021},
  publisher={Nature Publishing Group UK London}
}

@article{roetzer2022digital,
  title={The digital brain tumour atlas, an open histopathology resource},
  author={Roetzer-Pejrimovsky, Thomas and Moser, Anna-Christina and Atli, Baran and Vogel, Clemens Christian and Mercea, Petra A and Prihoda, Romana and Gelpi, Ellen and Haberler, Christine and H{\"o}ftberger, Romana and Hainfellner, Johannes A and others},
  journal={Scientific Data},
  volume={9},
  number={1},
  pages={55},
  year={2022},
  publisher={Nature Publishing Group UK London}
}

@article{pataki2022huncrc,
  title={HunCRC: annotated pathological slides to enhance deep learning applications in colorectal cancer screening},
  author={Pataki, B{\'a}lint {\'A}rmin and Olar, Alex and Ribli, Dezs{\H{o}} and Pesti, Adri{\'a}n and Kontsek, Endre and Gy{\"o}ngy{\"o}si, Benedek and Bilecz, {\'A}gnes and Kov{\'a}cs, Tekla and Kov{\'a}cs, Krist{\'o}f Attila and Kramer, Zs{\'o}fia and others},
  journal={Scientific Data},
  volume={9},
  number={1},
  pages={370},
  year={2022},
  publisher={Nature Publishing Group UK London}
}

@article{bulten2022artificial,
  title={Artificial intelligence for diagnosis and Gleason grading of prostate cancer: the PANDA challenge},
  author={Bulten, Wouter and Kartasalo, Kimmo and Chen, Po-Hsuan Cameron and Str{\"o}m, Peter and Pinckaers, Hans and Nagpal, Kunal and Cai, Yuannan and Steiner, David F and Van Boven, Hester and Vink, Robert and others},
  journal={Nature medicine},
  volume={28},
  number={1},
  pages={154--163},
  year={2022},
  publisher={Nature Publishing Group US New York}
}

@article{li2025diagnostic,
  title={Diagnostic text-guided representation learning in hierarchical classification for pathological whole slide image},
  author={Li, Jiawen and Sun, Qiehe and Yan, Renao and Wang, Yizhi and Fu, Yuqiu and Wei, Yani and Guan, Tian and Shi, Huijuan and He, Yonghong and Han, Anjia},
  journal={Medical Image Analysis},
  pages={103894},
  year={2025},
  publisher={Elsevier}
}

@inproceedings{fick2021partial,
  title={A partial label-based machine learning approach for cervical whole-slide image classification: the winning tissuenet solution},
  author={Fick, Rutger HJ and Tayart, Brice and Bertrand, Capucine and Lang, Solene Chan and Rey, Tina and Ciompi, Francesco and Tilmant, Cyprien and Farr{\'e}, Isabelle and Hadj, Saima Ben},
  booktitle={2021 43rd Annual International Conference of the IEEE Engineering in Medicine \& Biology Society (EMBC)},
  pages={2127--2131},
  year={2021},
  organization={IEEE}
}

@article{asadi2024machine,
  title={Machine learning-driven histotype diagnosis of ovarian carcinoma: insights from the OCEAN AI challenge},
  author={Asadi-Aghbolaghi, Maryam and Farahani, Hossein and Zhang, Allen and Akbari, Ardalan and Kim, Sirim and Chow, Ashley and Dane, Sohier and OCEAN Challenge Consortium and OTTA Consortium and Huntsman, David G and others},
  journal={medRxiv},
  pages={2024--04},
  year={2024},
  publisher={Cold Spring Harbor Laboratory Press}
}

@article{myles2025surgen,
  title={SurGen: 1020 H\&E-stained whole-slide images with survival and genetic markers},
  author={Myles, Craig and Um, In Hwa and Marshall, Craig and Harris-Birtill, David and Harrison, David J},
  journal={GigaScience},
  volume={14},
  pages={giaf086},
  year={2025},
  publisher={Oxford University Press}
}

@article{snell2017prototypical,
  title={Prototypical networks for few-shot learning},
  author={Snell, Jake and Swersky, Kevin and Zemel, Richard},
  journal={Advances in Neural Information Processing Systems},
  volume={30},
  year={2017}
}

@inproceedings{lu2023visual,
  title={Visual language pretrained multiple instance zero-shot transfer for histopathology images},
  author={Lu, Ming Y and Chen, Bowen and Zhang, Andrew and Williamson, Drew FK and Chen, Richard J and Ding, Tong and Le, Long Phi and Chuang, Yung-Sung and Mahmood, Faisal},
  booktitle={Proceedings of the IEEE/CVF conference on computer vision and pattern recognition},
  pages={19764--19775},
  year={2023}
}

@article{trinh2024improving,
  title={Improving robustness to corruptions with multiplicative weight perturbations},
  author={Trinh, Trung and Heinonen, Markus and Acerbi, Luigi and Kaski, Samuel},
  journal={Advances in Neural Information Processing Systems},
  volume={37},
  pages={35492--35516},
  year={2024}
}

@article{tizhoosh2026rethinking,
  title={Rethinking foundation models in pathology},
  author={Tizhoosh, Hamid R},
  journal={Nature Biomedical Engineering},
  pages={1--6},
  year={2026},
  publisher={Nature Publishing Group}
}

\clearpage

\renewcommand{\figurename}{Extended Data Figure}
\setcounter{figure}{0}

\renewcommand{\tablename}{Extended Data Table}
\setcounter{table}{0}

\begin{figure*}[htbp]
    \centering
    \includegraphics[width=0.8\linewidth]{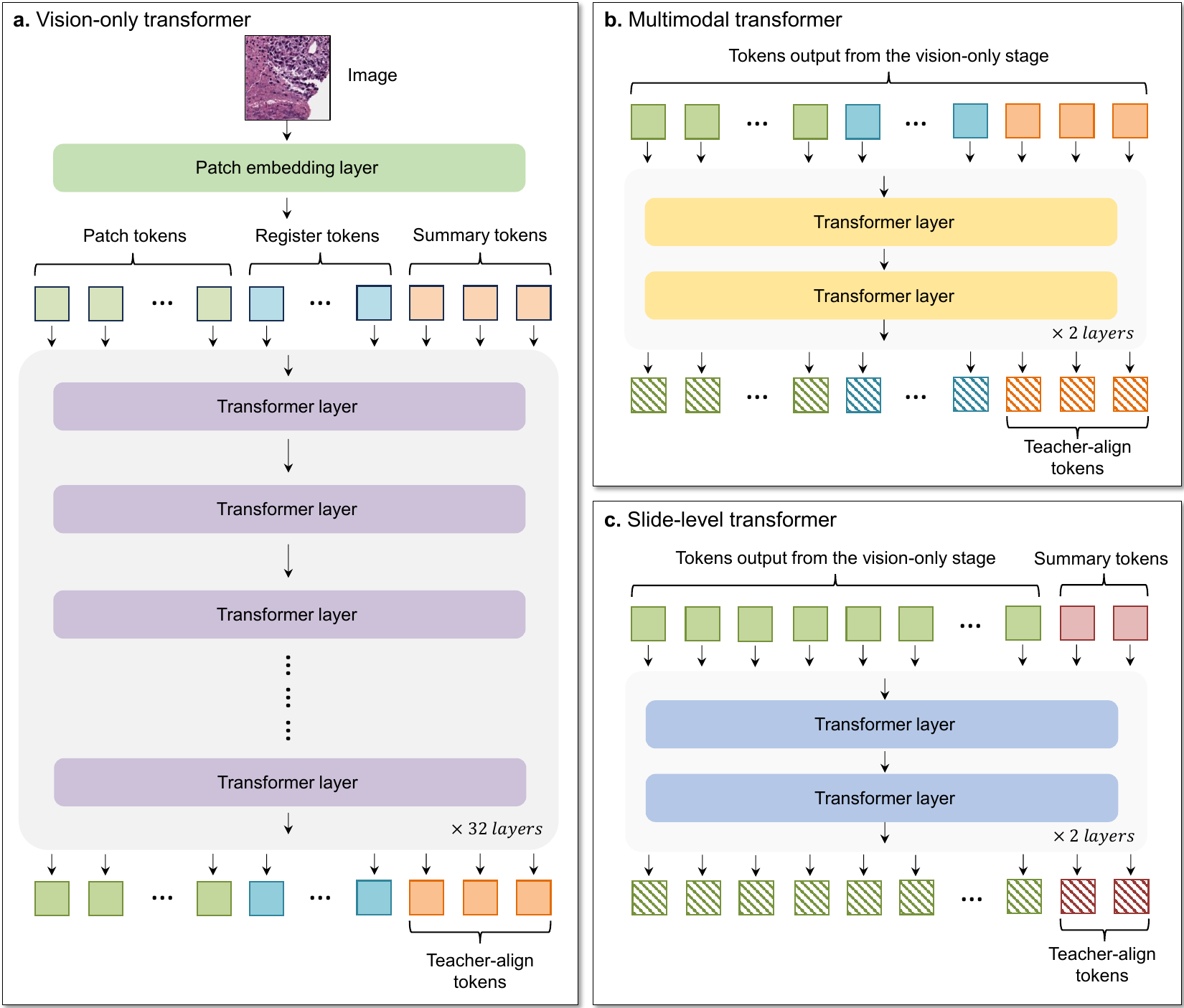}
    \caption{\textbf{Transformer architectures used in ALICE.} \textbf{a.} The vision-only transformer converts pathology images into patch tokens and combines them with register and summary tokens before processing them through $N$ transformer layers. Output summary tokens serve as teacher-alignment tokens. \textbf{b.} The multimodal transformer takes tokens from the vision-only stage and applies a two-layer transformer to generate teacher-alignment tokens for multimodal representation learning. \textbf{c.} The slide-level transformer combines vision-only tokens with slide-level summary tokens and applies a two-layer transformer to produce teacher-alignment tokens for slide-level representation learning.}
    \label{fig:extended_data_figure_1}
\end{figure*}

\clearpage

\begin{figure*}[htbp]
    \centering
    \includegraphics[width=0.99\linewidth]{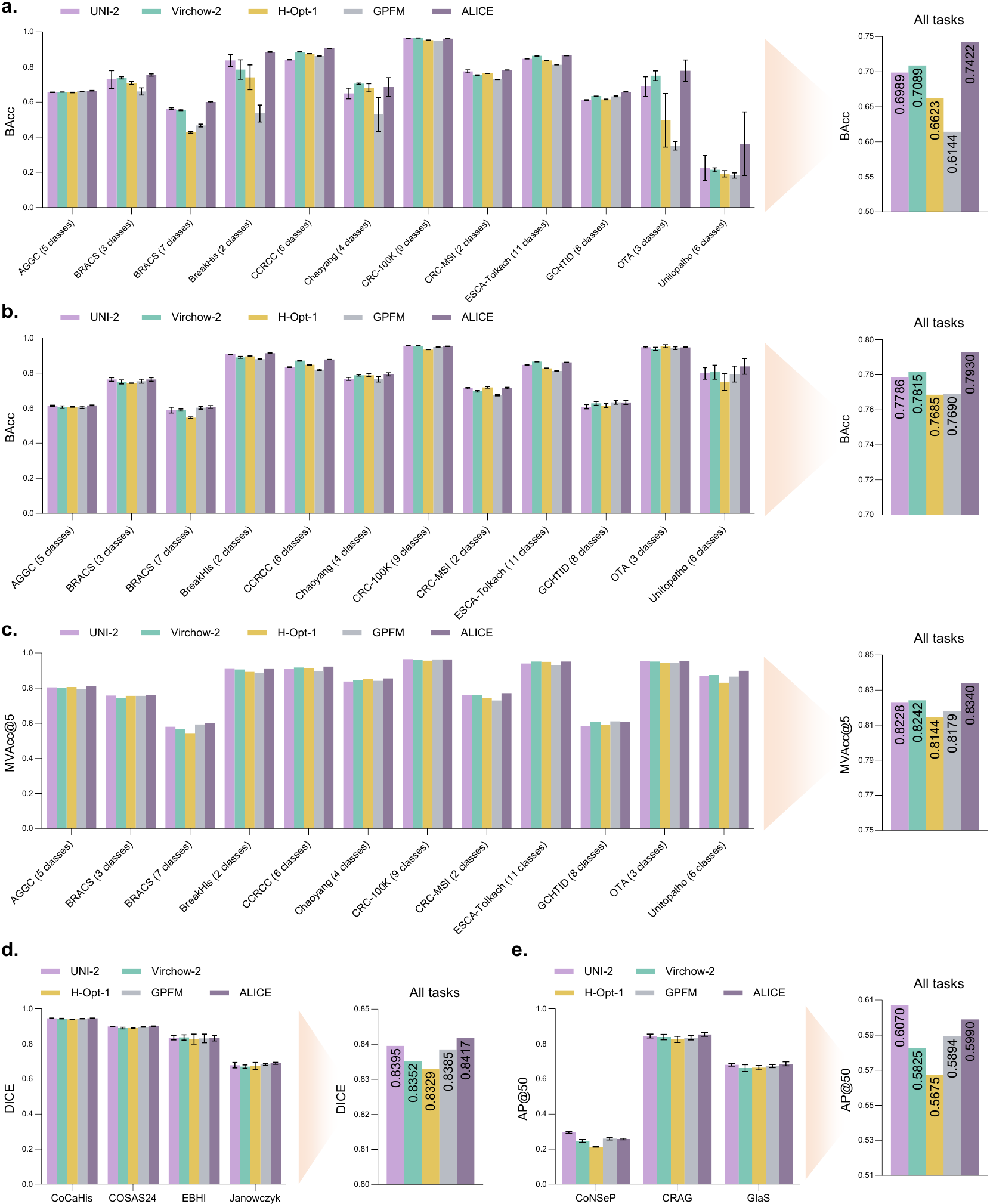}
    \caption{\textbf{Detailed ROI vision-only transfer results.} \textbf{a-e.} ROI-level evaluation of ALICE against UNI-2, Virchow-2, H-Opt-1 and GPFM using frozen image features. \textbf{a.} Linear-probe classification. \textbf{b.} KNN classification. \textbf{c.} Image-to-image retrieval. \textbf{d.} Semantic segmentation. \textbf{e.} Instance segmentation.}
    \label{fig:extended_data_figure_2}
\end{figure*}

\clearpage

\begin{figure*}[htbp]
    \centering
    \includegraphics[width=0.99\linewidth]{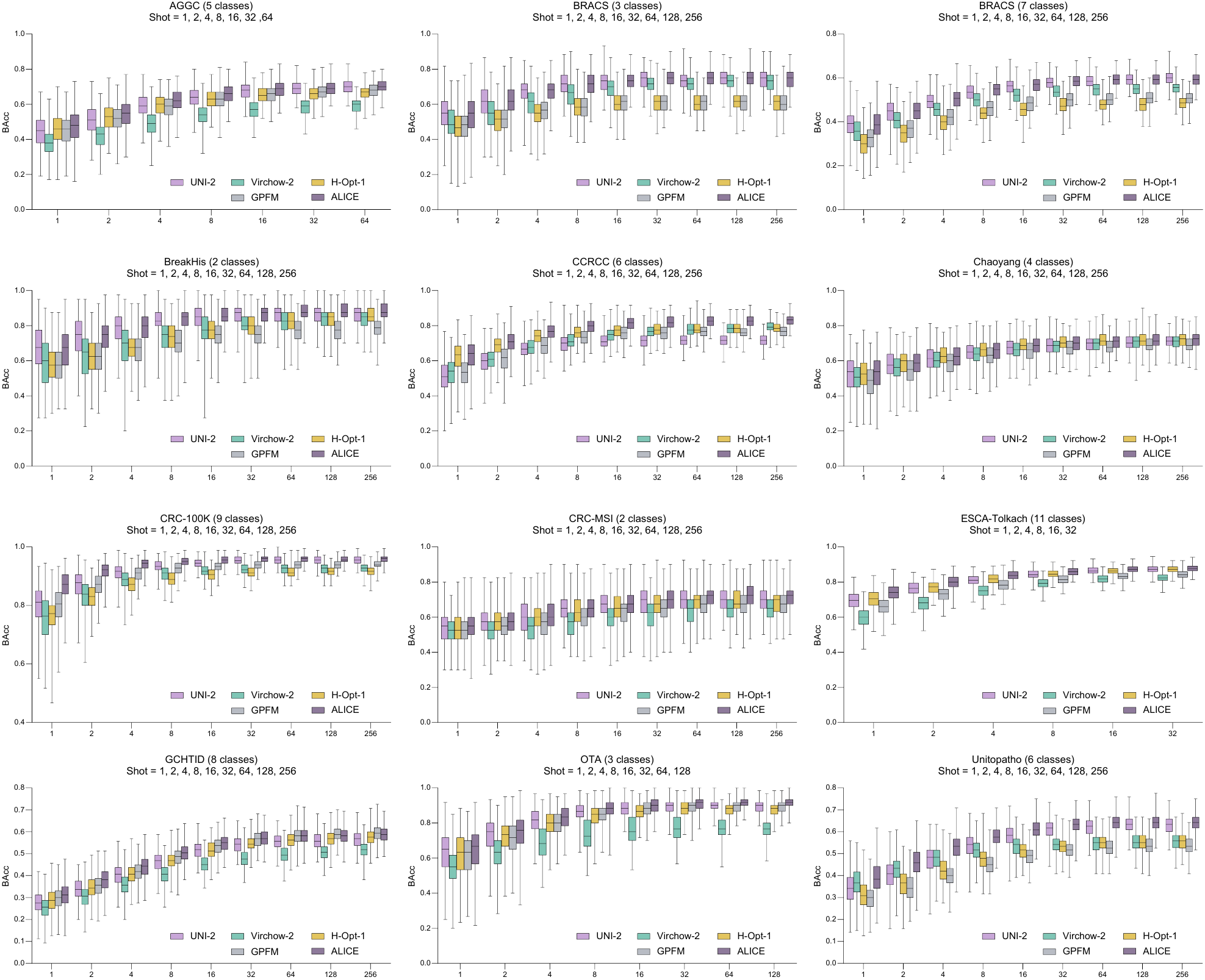}
    \caption{\textbf{Detailed ROI few-shot classification performance.} Few-shot ROI classification of ALICE and four vision-only PFMs across 12 datasets using all-way ProtoNet episodes.}
    \label{fig:extended_data_figure_3}
\end{figure*}

\clearpage

\begin{figure*}[htbp]
    \centering
    \includegraphics[width=0.99\linewidth]{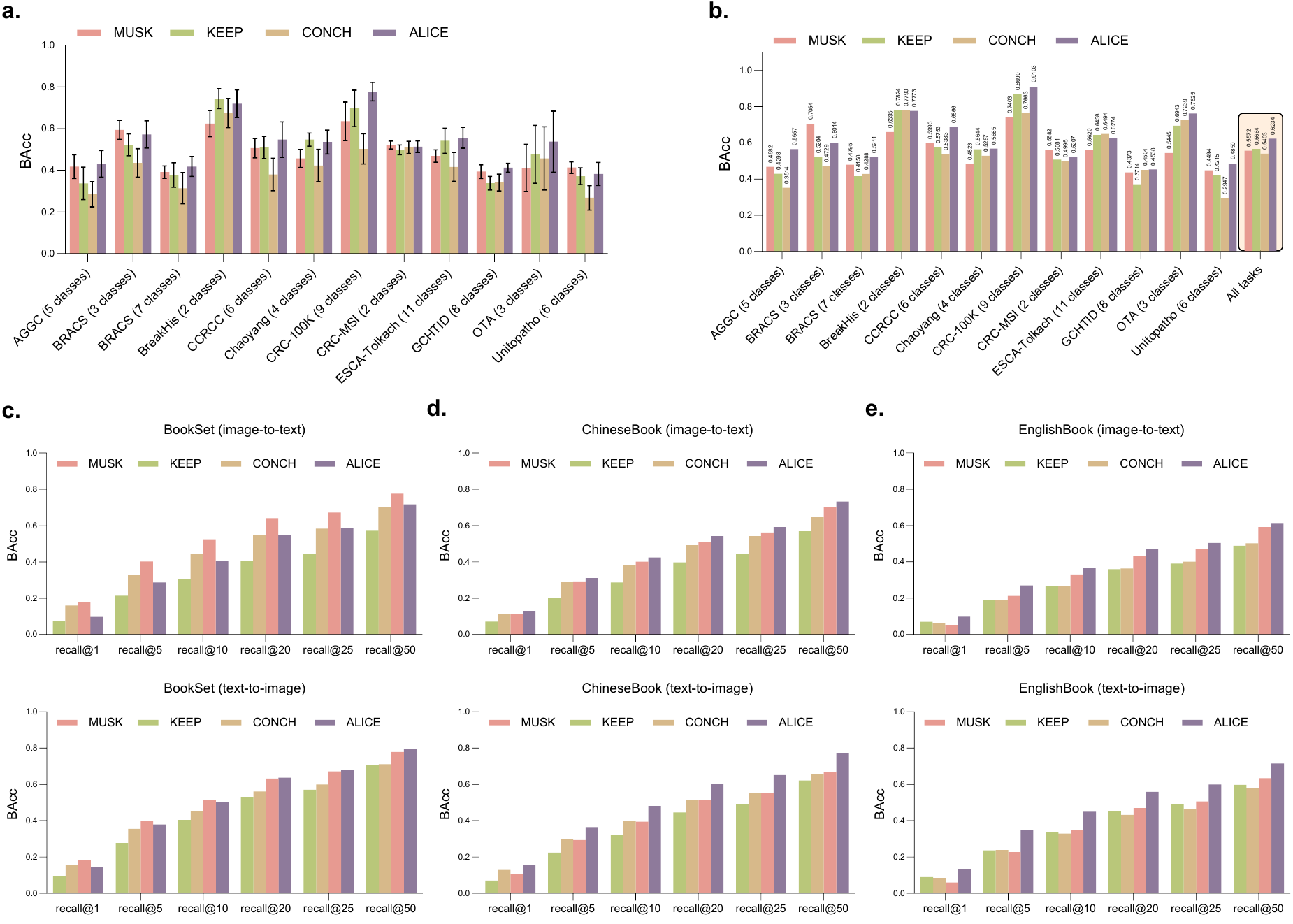}
    \caption{\textbf{Detailed ROI vision-language transfer results.} \textbf{a.} Prompt-level zero-shot ROI classification of ALICE, MUSK, KEEP and CONCH across 12 ROI datasets. \textbf{b.} Prompt-ensemble zero-shot ROI classification across the same 12 datasets. \textbf{c-e.} Image-to-text and text-to-image retrieval on BookSet, ChineseBook and EnglishBook, respectively.}
    \label{fig:extended_data_figure_4}
\end{figure*}

\clearpage

\begin{figure*}[htbp]
    \centering
    \includegraphics[width=0.99\linewidth]{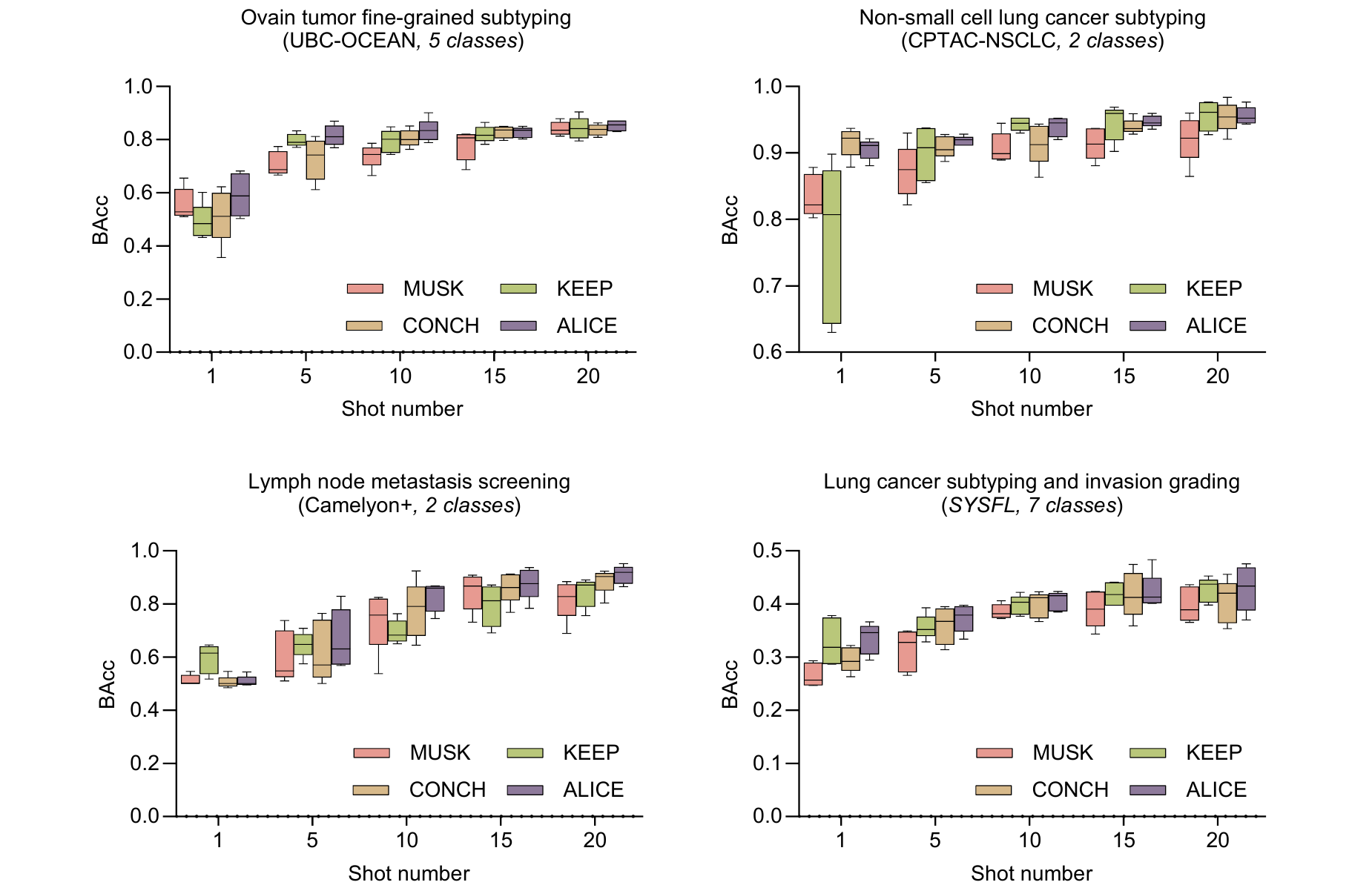}
    \caption{\textbf{Detailed WSI few-shot vision-language classification results.} PathPT-based few-shot WSI classification of ALICE, MUSK, KEEP and CONCH across four clinical tasks using 1, 5, 10, 15, and 20 labelled slides per class.}
    \label{fig:extended_data_figure_5}
\end{figure*}

\clearpage

\begin{figure*}[htbp]
    \centering
    \includegraphics[width=0.99\linewidth]{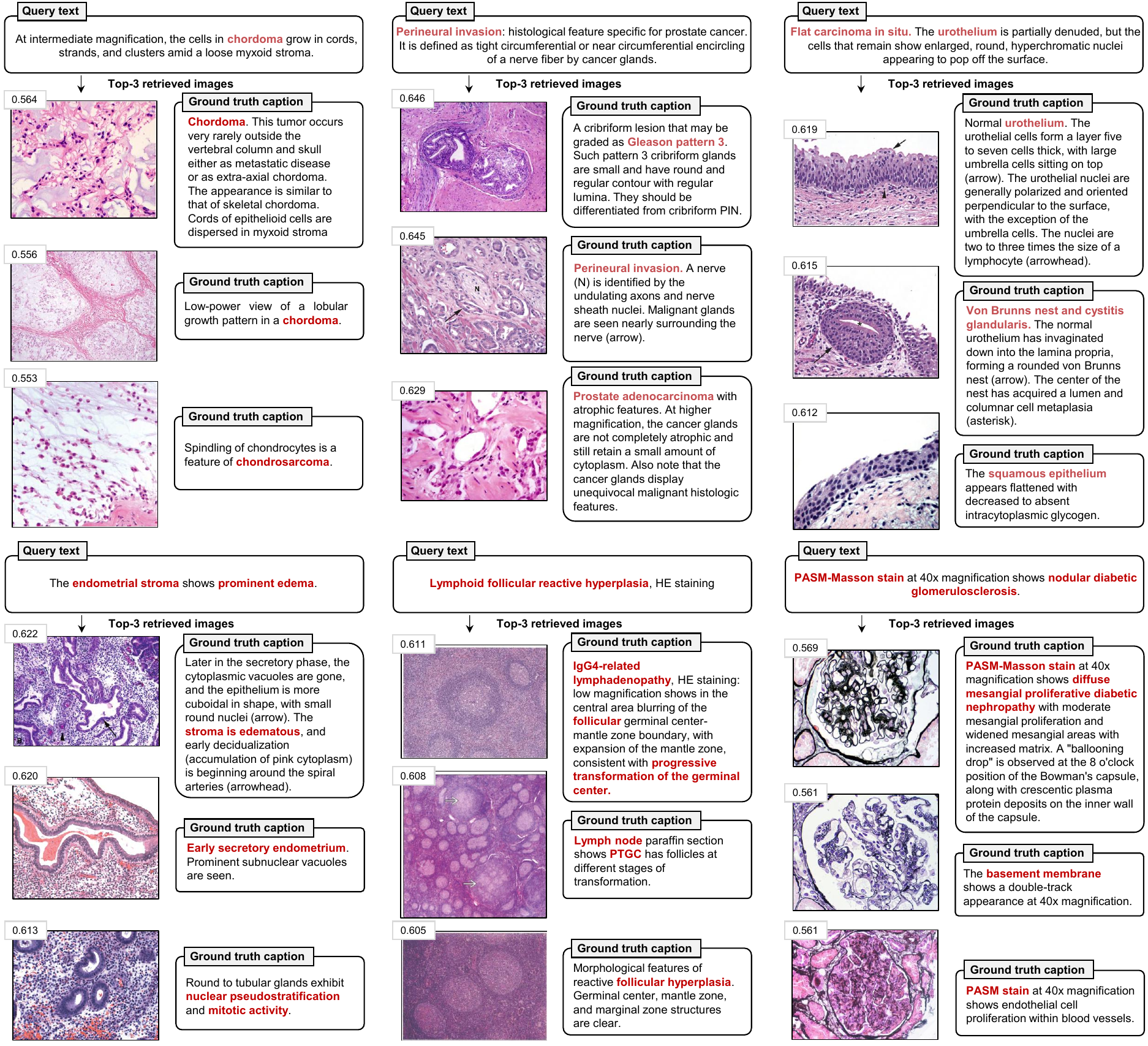}
    \caption{\textbf{Representative ROI text-to-image cross-modal retrieval examples.} Text-to-image retrieval examples using ALICE on paired pathology image-caption data. Each query text is used to retrieve the top three ROI images, with cosine similarity scores and the corresponding ground-truth captions shown. Red text highlights key pathological terms shared between the query and retrieved results.}
    \label{fig:extended_data_figure_6}
\end{figure*}

\clearpage

\begin{figure*}[htbp]
    \centering
    \includegraphics[width=0.99\linewidth]{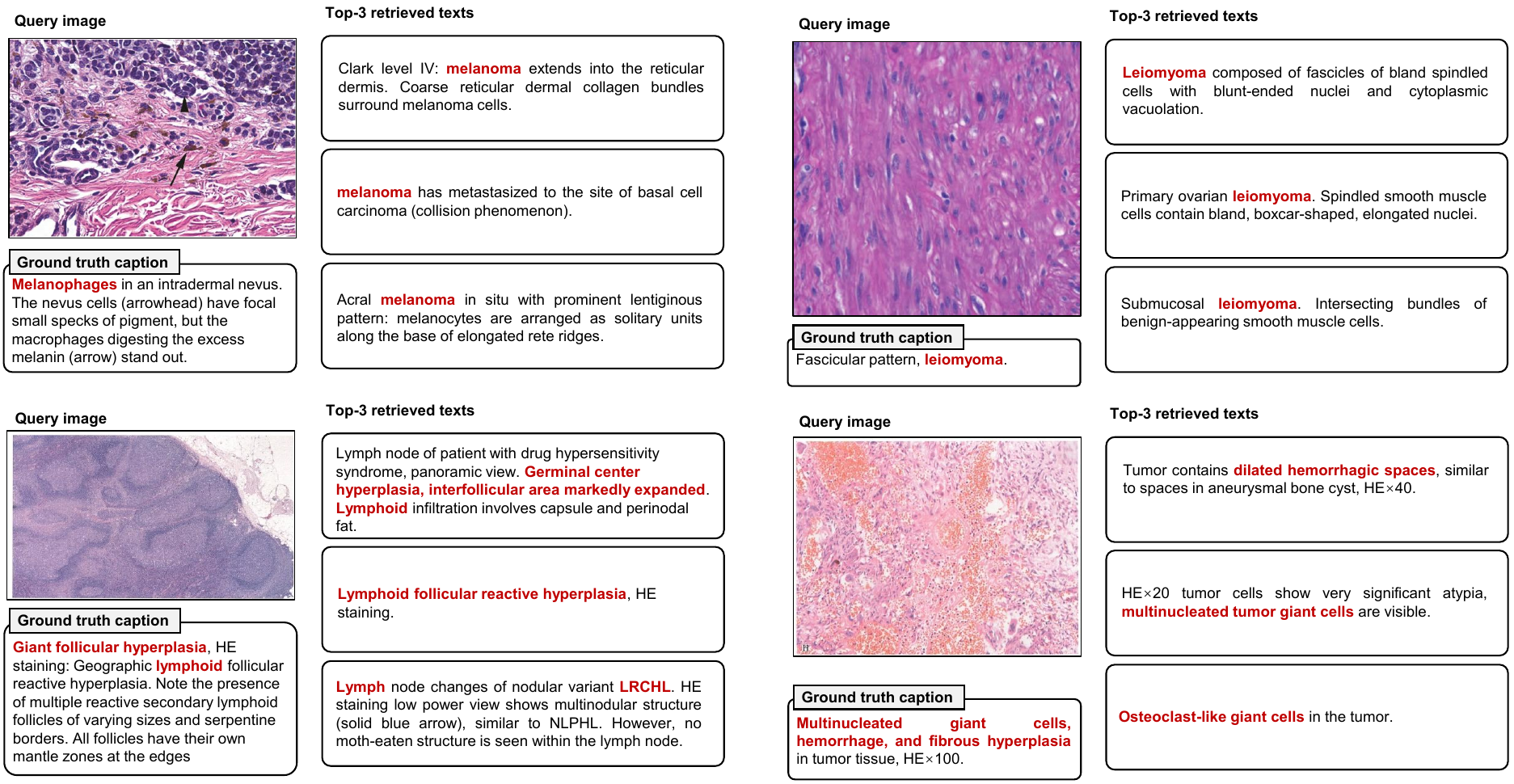}
    \caption{\textbf{Representative ROI image-to-text cross-modal retrieval examples.} Image-to-text retrieval examples using ALICE on paired pathology image-caption data. Each query image is used to retrieve the top three text descriptions, with the ground-truth caption shown for reference. Red text highlights key pathological terms shared between the query and retrieved results.}
    \label{fig:extended_data_figure_7}
\end{figure*}

\clearpage

\begin{figure*}[htbp]
    \centering
    \includegraphics[width=0.99\linewidth]{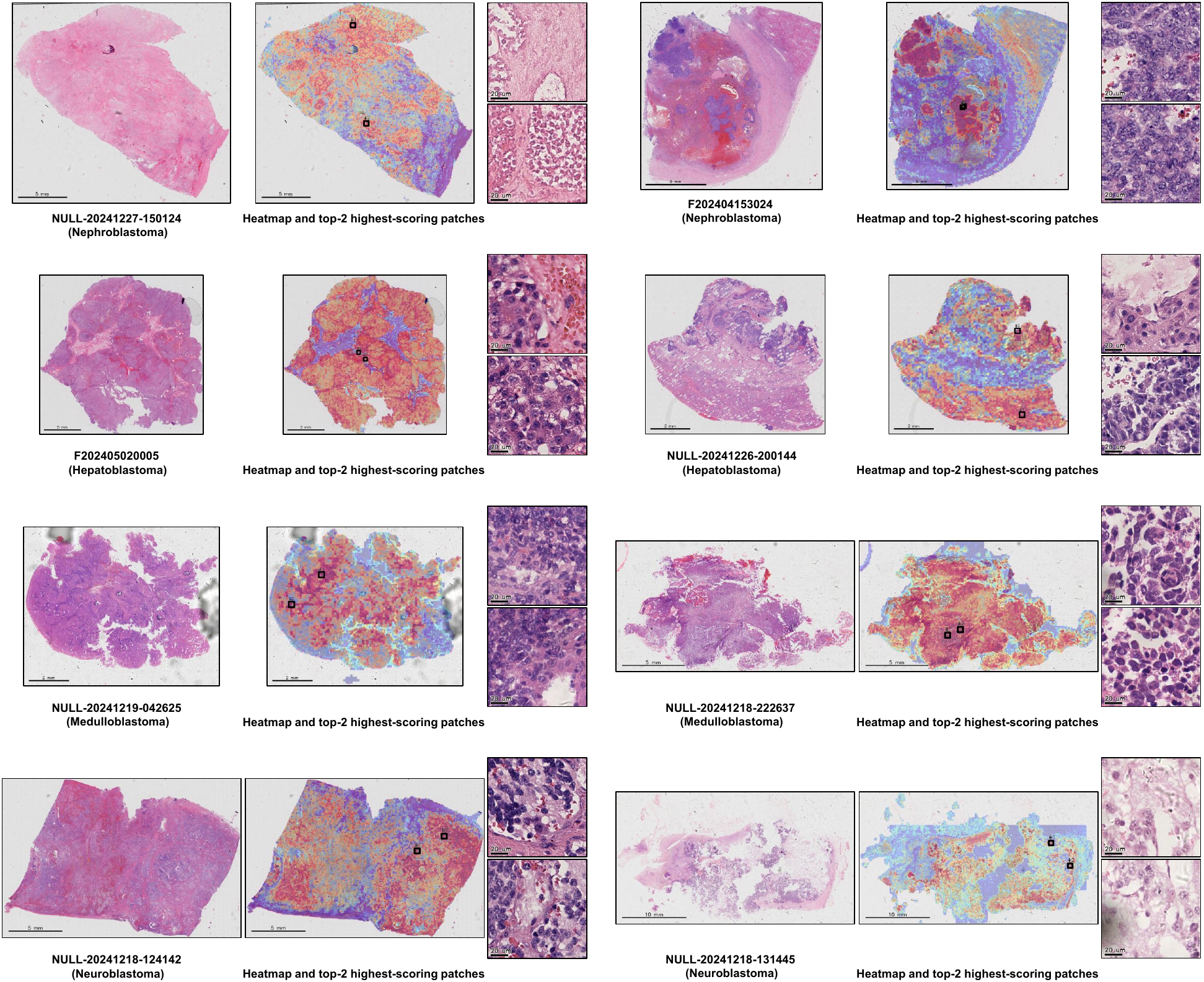}
    \caption{\textbf{Representative WSI heatmaps and high-scoring regions for pediatric rare tumor classification.} Representative ALICE predictions on KidRare whole-slide images covering nephroblastoma, hepatoblastoma, medulloblastoma and neuroblastoma. For each case, the original H\&E-stained WSI is shown together with the corresponding prediction heatmap and the top two highest-scoring patches.}
    \label{fig:extended_data_figure_8}
\end{figure*}

\clearpage

\begin{figure*}[htbp]
    \centering
    \includegraphics[width=0.99\linewidth]{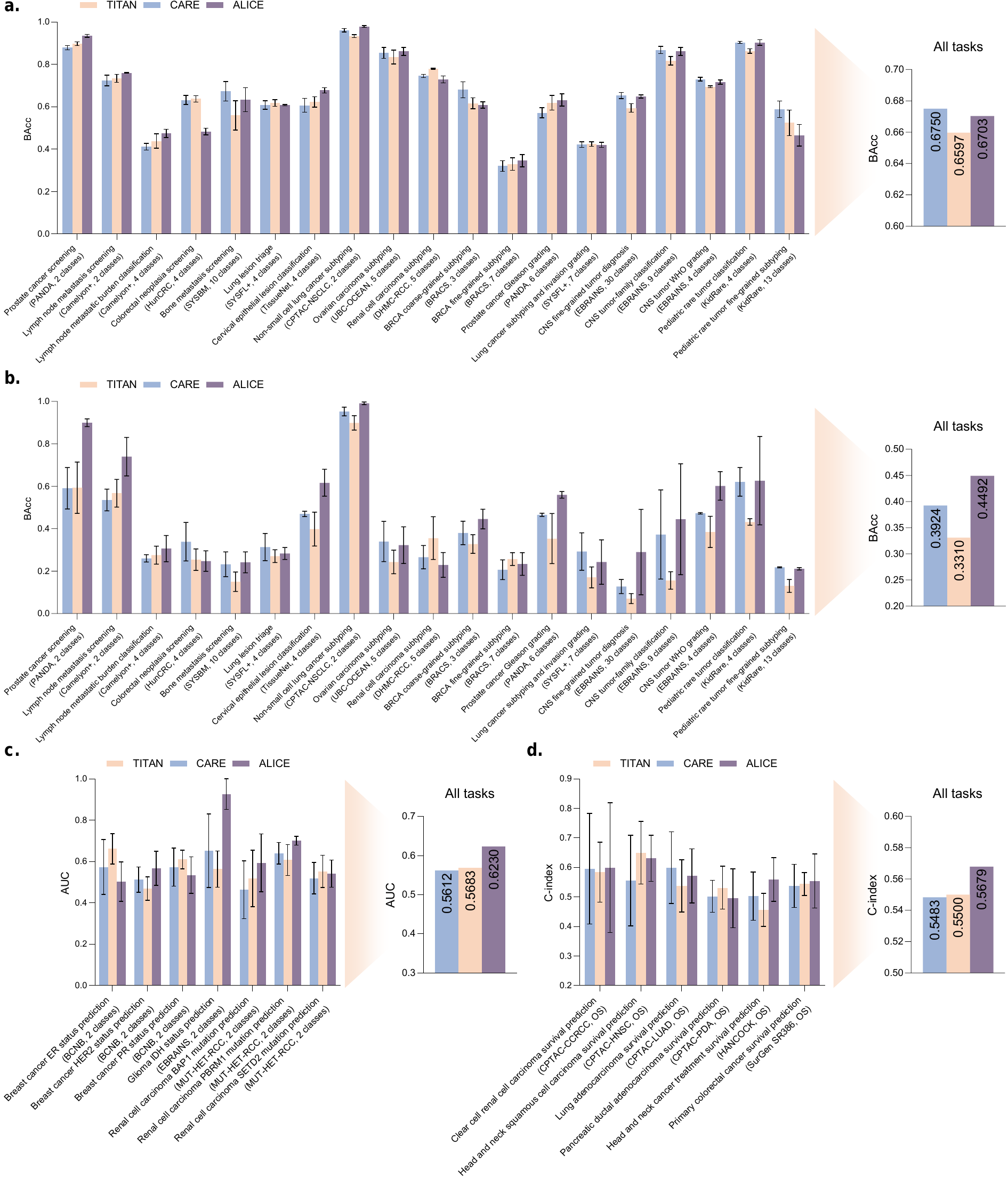}
    \caption{\textbf{Detailed slide-level WSI transfer results.} \textbf{a.} Slide-level diagnostic classification using KNN on frozen WSI embeddings from ALICE, TITAN and CARE across 19 tasks. \textbf{b.} Slide-level diagnostic classification using linear probing on frozen WSI embeddings across the same 19 tasks. \textbf{c.} Slide-level biomarker prediction using linear probing across seven tasks. \textbf{d,} Slide-level survival prediction across six tasks using Cox proportional hazards models.}
    \label{fig:extended_data_figure_9}
\end{figure*}

\clearpage

\begin{figure*}[htbp]
    \centering
    \includegraphics[width=0.85\linewidth]{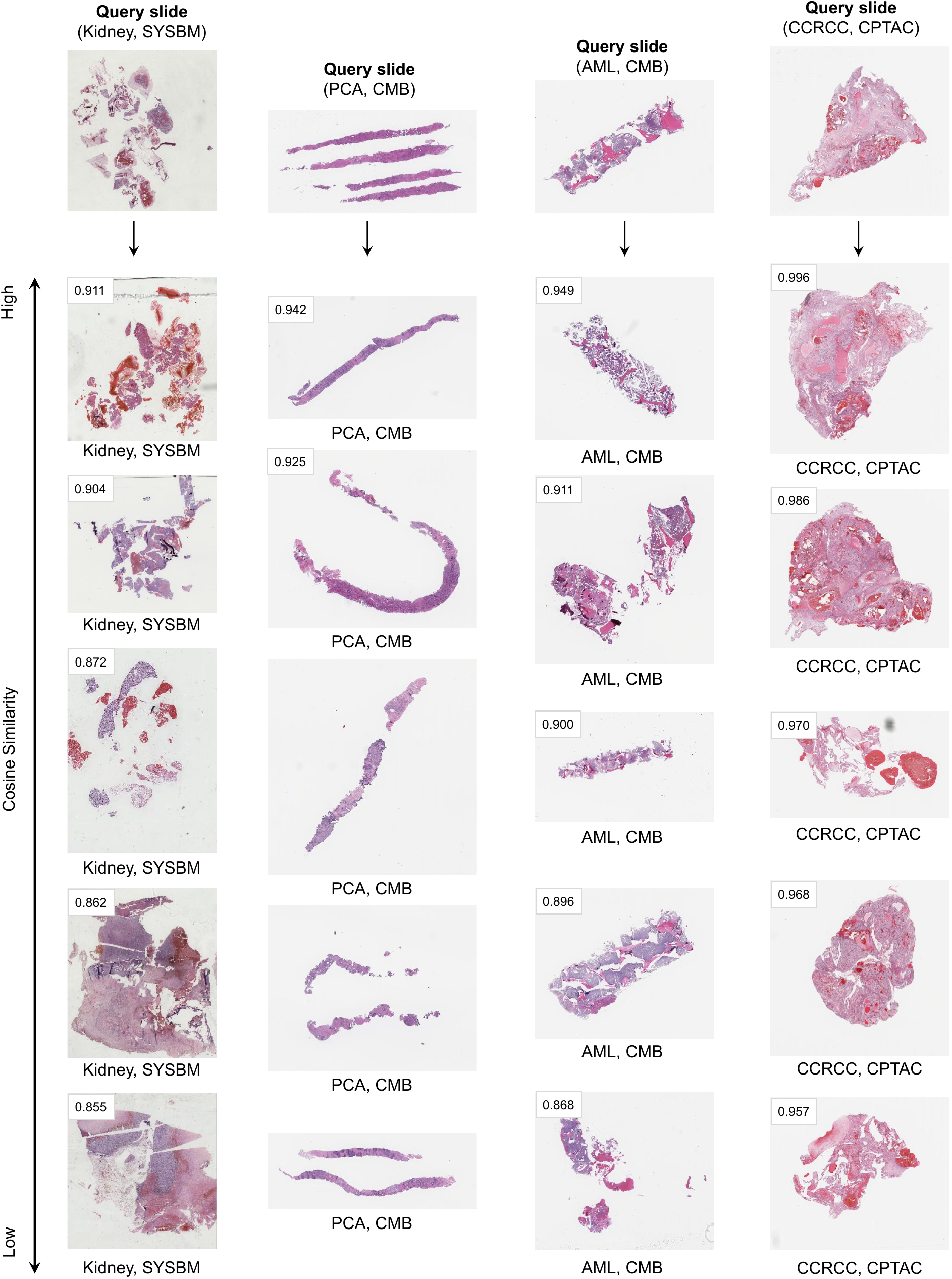}
    \caption{\textbf{Representative slide-to-slide retrieval examples.} Slide-to-slide retrieval using ALICE slide-level embeddings. Four query WSIs from SYSBM, CMB and CPTAC are shown with their top five retrieved slides ranked by cosine similarity. Similarity scores, diagnostic labels and data sources are shown for each retrieved slide.}
    \label{fig:extended_data_figure_10}
\end{figure*}

\clearpage

\begin{table*}[htbp]
\centering
\resizebox{0.45\columnwidth}{!}{
\renewcommand{\arraystretch}{1}{
\begin{tabular}{ll}
\toprule
Hyperparameter & Value \\
\midrule
Transformer layers & 32\\
Transformer heads & 16 \\
Transformer embedding dimension & 1280 \\
Transformer patch size & 14 \\
Summary tokens & 3 \\
Register tokens & 7 \\
\midrule
Patch crop size & 98 $\times$ 98 \\
Patch crop number & 8 \\
Patch crop grid & 7 $\times$ 7\\
Random resized crop scale & (0.5, 1.0) \\
Random resized crop aspect ratio & (0.75, 1.33) \\
Horizontal flip probability & 0.5 \\
Vertical flip probability & 0.5 \\
Colour jitter & 0.4 \\
EMA decay (cosine schedule) &  (0.996, 1.0) \\
DAMP & 0.05 \\
\midrule
Training precision & fp16 \\
AdamW $\beta$ & (0.9, 0.999) \\
Weight decay & 0.05 \\
Linear warmup steps & $10^4$\\
Training steps & $6\times10^5$ \\
Peak learning rate (cosine decay) & $3\times10^{-4}$ \\
Minimum learning rate (cosine decay) & $1\times10^{-5}$ \\
Gradient clipping & 1.0 \\
\bottomrule
\end{tabular}}}
\caption{\textbf{ALICE vision-only pretraining hyperparameters.} 8 $\times$ NVIDIA A100 80GB GPUs were used for training. The total effective batch size is 1024 with 64 per GPU and 2 gradient accumulations.}
\label{tab:stage1_pretrain_settings}
\end{table*}

\clearpage

\begin{table*}[htbp]
\centering
\resizebox{0.45\columnwidth}{!}{
\renewcommand{\arraystretch}{1}{
\begin{tabular}{ll}
\toprule
Hyperparameter & Value \\
\midrule
Transformer layers & 2 \\
Transformer heads & 16 \\
Transformer embedding dimension & 1280 \\
Patch crop size & 98 $\times$ 98 \\
Patch crop number & 4 \\
Patch crop grid & 7 $\times$ 7 \\
Random resized crop scale & (0.5, 1.0) \\
Random resized crop aspect ratio & (0.75, 1.33) \\
Horizontal flip probability & 0.5 \\
Vertical flip probability & 0.5 \\
Colour jitter & 0.4 \\
KEEP input size & 224 $\times$ 224 \\
CONCH input size & 448 $\times$ 448 \\
MUSK input size & 384 $\times$ 384 \\
EMA decay (cosine schedule) &  (0.996, 1.0) \\
DAMP & 0.02 \\
\midrule
Training precision & bf16 \\
AdamW \(\beta\) & (0.9, 0.999) \\
Weight decay & 0.02 \\
Linear warmup steps & $10^4$\\
Training steps & $6\times10^5$ \\
Peak learning rate & \(3\times10^{-4}\) \\
Minimum learning rate & \(1\times10^{-5}\) \\
Gradient clipping & 1.0 \\
\bottomrule
\end{tabular}}}
\caption{\textbf{ALICE multimodal pretraining hyperparameters.} 8 $\times$ NVIDIA A100 80GB GPUs were used for training. The total effective batch size is 1024 with 32 per GPU and 4 gradient accumulations.}
\label{tab:stage2_pretrain_settings}
\end{table*}

\clearpage

\begin{table*}[htbp]
\centering
\resizebox{0.45\columnwidth}{!}{
\renewcommand{\arraystretch}{1}{
\begin{tabular}{ll}
\toprule
Hyperparameter & Value \\
\midrule
Transformer layers & 2 \\
Transformer heads & 8 \\
Transformer input dimension & 3840 \\
Transformer hidden dimension & 1024 \\
Attention heads & 8 \\
Dropout & 0.25 \\
Summary tokens & 2 \\
\midrule
Feature brightness strength & 0.05 \\
TITAN patch loss weight & 0.5 \\
EMA decay (cosine schedule) &  (0.996, 1.0) \\
DAMP & 0.02 \\
\midrule
Training precision & bf16 \\
AdamW \(\beta\) & (0.9, 0.999) \\
Weight decay & 0.05 \\
Linear warmup steps & $5\times10^3$\\
Training steps & $2\times10^5$ \\
Peak learning rate & \(3\times10^{-4}\) \\
Minimum learning rate & \(1\times10^{-5}\) \\
Gradient clipping & 1.0 \\
\bottomrule
\end{tabular}}}
\caption{\textbf{ALICE slide-level pretraining hyperparameters.} 8 $\times$ NVIDIA A100 80GB GPUs were used for training. The total effective batch size is 1024 with 128 per GPU.}
\label{tab:stage3_pretrain_settings}
\end{table*}


\begin{table*}[htbp]
\centering
\resizebox{0.85\textwidth}{!}{
\renewcommand{\arraystretch}{1.}{
\begin{tabular}{lp{0.75\textwidth}}
\toprule
ID & Prompt template \\
\midrule
1 & an H\&E stained pathology image patch of \{label\_text\}. \\
2 & a histopathology image showing \{label\_text\}. \\
3 & a microscopic pathology image of \{label\_text\}. \\
4 & a tissue patch containing \{label\_text\}. \\
5 & a pathology slide region with \{label\_text\}. \\
6 & a pathology ROI consistent with \{label\_text\}. \\
7 & a diagnostic histology image of \{label\_text\}. \\
8 & a histologic tissue patch of \{label\_text\}. \\
9 & a pathology image whose main finding is \{label\_text\}. \\
10 & a microscopic image of \{label\_text\}. \\
\bottomrule
\end{tabular}}}
\caption{\textbf{Generic prompt templates for ROI zero-shot classification.}}
\label{tab:roi_zero_shot_prompt_templates_general}
\end{table*}

\newcommand{\classnames}[1]{\begin{tabular}[t]{@{}l@{}}#1\end{tabular}}

\begin{table*}[htbp]
\centering
\resizebox{0.85\textwidth}{!}{
\renewcommand{\arraystretch}{1.}{
\begin{tabular}{p{0.25\textwidth}|p{0.62\textwidth}}
\toprule
Class & Class names \\
\midrule
Gleason pattern 3 & \classnames{prostate tissue with Gleason pattern 3\\
prostatic adenocarcinoma of Gleason pattern 3\\
Gleason grade 3 prostate cancer\\
well formed glands of Gleason pattern 3\\
prostate tumor tissue showing Gleason pattern 3} \\
\midrule
Gleason pattern 4 & \classnames{prostate tissue with Gleason pattern 4\\
prostatic adenocarcinoma of Gleason pattern 4\\
Gleason grade 4 prostate cancer\\
fused or cribriform glands of Gleason pattern 4\\
prostate tumor tissue showing Gleason pattern 4} \\
\midrule
Gleason pattern 5 & \classnames{prostate tissue with Gleason pattern 5\\
prostatic adenocarcinoma of Gleason pattern 5\\
Gleason grade 5 prostate cancer\\
poorly differentiated Gleason pattern 5 tumor\\
prostate tumor tissue showing Gleason pattern 5} \\
\midrule
Normal prostate tissue & \classnames{normal prostate tissue\\
benign prostate tissue\\
non neoplastic prostate glands\\
histologically normal prostate\\
normal prostatic glandular tissue} \\
\midrule
Prostatic stroma & \classnames{prostatic stroma\\
stromal tissue of the prostate\\
fibromuscular prostate stroma\\
non epithelial stromal prostate tissue\\
prostate stromal region} \\
\bottomrule
\end{tabular}}}
\caption{\textbf{Class prompts for AGGC ROI zero-shot classification.}}
\label{tab:prompt_aggc}
\end{table*}

\begin{table*}[htbp]
\centering
\resizebox{0.85\textwidth}{!}{
\renewcommand{\arraystretch}{1.}{
\begin{tabular}{p{0.25\textwidth}|p{0.62\textwidth}}
\toprule
Class & Class names \\
\midrule
Benign breast tumors & \classnames{a benign breast lesion\\
benign breast tumor tissue\\
non malignant breast lesion\\
benign breast pathology\\
pathologically benign breast tissue} \\
\midrule
Atypical breast tumors & \classnames{an atypical breast lesion\\
atypical breast tumor tissue\\
borderline atypical breast pathology\\
breast tissue with atypia\\
an atypical proliferative breast lesion} \\
\midrule
Malignant breast tumors & \classnames{a malignant breast tumor\\
malignant breast carcinoma tissue\\
invasive or in situ malignant breast lesion\\
breast cancer tissue\\
pathologically malignant breast tissue} \\
\bottomrule
\end{tabular}}}
\caption{\textbf{Class prompts for BRACS (3 classes) ROI zero-shot classification.}}
\label{tab:prompt_bracs_3classes}
\end{table*}

\clearpage

\begin{table*}[htbp]
\centering
\resizebox{0.9\textwidth}{!}{
\renewcommand{\arraystretch}{1.}{
\begin{tabular}{p{0.28\textwidth}|p{0.60\textwidth}}
\toprule
Class & Class names \\
\midrule
Normal breast tissue & \classnames{normal breast tissue\\
non neoplastic breast tissue\\
histologically normal breast parenchyma\\
normal mammary tissue\\
benign normal breast epithelium} \\
\midrule
Pathologically benign breast lesion & \classnames{a pathologically benign breast lesion\\
benign proliferative breast lesion\\
pathological benign breast tissue\\
benign breast abnormality\\
non malignant pathological breast lesion} \\
\midrule
Usual ductal hyperplasia & \classnames{usual ductal hyperplasia of the breast\\
breast tissue with usual ductal hyperplasia\\
usual ductal hyperplasia lesion\\
UDH breast lesion\\
benign ductal hyperplasia in breast tissue} \\
\midrule
Flat epithelial atypia & \classnames{flat epithelial atypia of the breast\\
breast tissue with flat epithelial atypia\\
FEA breast lesion\\
flat epithelial atypia lesion\\
atypical flat epithelial breast lesion} \\
\midrule
Atypical ductal hyperplasia & \classnames{atypical ductal hyperplasia of the breast\\
breast tissue with atypical ductal hyperplasia\\
ADH breast lesion\\
atypical proliferative ductal breast lesion\\
atypical ductal hyperplasia lesion} \\
\midrule
Ductal carcinoma in situ & \classnames{ductal carcinoma in situ of the breast\\
breast tissue with ductal carcinoma in situ\\
DCIS breast lesion\\
in situ ductal breast carcinoma\\
non invasive ductal breast carcinoma} \\
\midrule
Invasive breast carcinoma & \classnames{invasive breast carcinoma\\
breast tissue with invasive carcinoma\\
infiltrating breast carcinoma\\
IC breast lesion\\
malignant invasive breast tumor} \\
\bottomrule
\end{tabular}}}
\caption{\textbf{Class prompts for BRACS (7 classes) ROI zero-shot classification.}}
\label{tab:prompt_bracs_7classes}
\end{table*}

\begin{table*}[htbp]
\centering
\resizebox{0.8\textwidth}{!}{
\renewcommand{\arraystretch}{1.}{
\begin{tabular}{p{0.25\textwidth}|p{0.60\textwidth}}
\toprule
Class & Class names \\
\midrule
Benign breast tumor & \classnames{a benign breast tumor\\
benign breast lesion\\
non malignant breast pathology\\
benign mammary tumor tissue\\
histologically benign breast tissue} \\
\midrule
Malignant breast tumor & \classnames{a malignant breast tumor\\
breast carcinoma tissue\\
malignant breast pathology\\
cancerous breast lesion\\
histologically malignant breast tissue} \\
\bottomrule
\end{tabular}}}
\caption{\textbf{Class prompts for BreakHis ROI zero-shot classification.}}
\label{tab:prompt_breakhis}
\end{table*}

\begin{table*}[htbp]
\centering
\resizebox{0.9\textwidth}{!}{
\renewcommand{\arraystretch}{1.}{
\begin{tabular}{p{0.28\textwidth}|p{0.60\textwidth}}
\toprule
Class & Class names \\
\midrule
Blood-rich renal tissue & \classnames{blood rich renal tissue\\
hemorrhagic tissue region\\
blood filled histology patch\\
red blood cell rich tissue\\
vascular blood region} \\
\midrule
Clear cell renal cell carcinoma & \classnames{clear cell renal cell carcinoma\\
ccRCC tumor tissue\\
renal carcinoma tissue\\
kidney cancer tissue\\
clear cell kidney tumor} \\
\midrule
Background or empty region & \classnames{background or empty region\\
empty histology patch\\
blank slide background\\
tissue free region\\
non tissue background patch} \\
\midrule
Normal kidney tissue & \classnames{normal kidney tissue\\
non neoplastic renal tissue\\
histologically normal renal parenchyma\\
benign kidney tissue\\
normal renal tissue} \\
\midrule
Other renal tissue component & \classnames{other renal tissue component\\
miscellaneous kidney tissue\\
kidney tissue of other type\\
other non tumor renal component\\
renal tissue not in the main categories} \\
\midrule
Renal stroma & \classnames{renal stroma\\
stromal kidney tissue\\
fibrous tumor associated stroma\\
stromal tissue region\\
connective stromal tissue} \\
\bottomrule
\end{tabular}}}
\caption{\textbf{Class prompts for CCRCC ROI zero-shot classification.}}
\label{tab:prompt_ccrcc}
\end{table*}

\begin{table*}[htbp]
\centering
\resizebox{0.85\textwidth}{!}{
\renewcommand{\arraystretch}{1.}{
\begin{tabular}{p{0.28\textwidth}|p{0.58\textwidth}}
\toprule
Class & Class names \\
\midrule
Normal colorectal tissue & \classnames{normal colorectal tissue\\
histologically normal colon mucosa\\
normal colorectal mucosa\\
non neoplastic colorectal tissue\\
benign normal colorectal epithelium} \\
\midrule
Serrated colorectal lesion & \classnames{a serrated colorectal lesion\\
serrated colorectal polyp\\
colorectal tissue with serrated architecture\\
serrated lesion of the colon\\
serrated colorectal pathology} \\
\midrule
Colorectal adenocarcinoma & \classnames{colorectal adenocarcinoma\\
colorectal cancer tissue\\
adenocarcinoma of the colon or rectum\\
malignant colorectal glandular tumor\\
tumor epithelium of colorectal adenocarcinoma} \\
\midrule
Colorectal adenoma & \classnames{colorectal adenoma\\
adenomatous colorectal polyp\\
colorectal tissue with adenoma\\
benign colorectal adenomatous lesion\\
adenoma of the colon or rectum} \\
\bottomrule
\end{tabular}}}
\caption{\textbf{Class prompts for Chaoyang ROI zero-shot classification.}}
\label{tab:prompt_chaoyang}
\end{table*}

\begin{table*}[htbp]
\centering
\resizebox{0.9\textwidth}{!}{
\renewcommand{\arraystretch}{1.}{
\begin{tabular}{p{0.28\textwidth}|p{0.60\textwidth}}
\toprule
Class & Class names \\
\midrule
Adipose tissue & \classnames{adipose tissue\\
fat tissue in colorectal histology\\
adipocytes\\
colorectal adipose tissue\\
fat rich tissue region} \\
\midrule
Background & \classnames{histology image background\\
background region\\
non tissue background patch\\
slide background\\
empty histologic background} \\
\midrule
Debris & \classnames{debris\\
necrotic debris\\
cellular debris in colorectal tissue\\
degenerated tissue debris\\
tumor associated debris} \\
\midrule
Lymphocytes & \classnames{lymphocytes\\
lymphocyte rich tissue\\
immune cell rich region\\
lymphoid infiltrate\\
dense lymphocytic infiltrate} \\
\midrule
Mucus & \classnames{mucus\\
mucin rich tissue\\
mucus producing region\\
mucinous material\\
colorectal mucus region} \\
\midrule
Smooth muscle & \classnames{smooth muscle\\
muscular tissue\\
smooth muscle region\\
muscle layer tissue\\
colorectal smooth muscle} \\
\bottomrule
\end{tabular}}}
\caption{\textbf{Class prompts for CRC-100K ROI zero-shot classification.} Continued.}
\label{tab:prompt_crc100k_v1}
\end{table*}
\begin{table*}[htbp]
\centering
\resizebox{0.9\textwidth}{!}{
\renewcommand{\arraystretch}{1.}{
\begin{tabular}{p{0.28\textwidth}|p{0.60\textwidth}}
\toprule
Class & Class names \\
\midrule
Normal colon mucosa & \classnames{normal colon mucosa\\
histologically normal colorectal mucosa\\
non neoplastic colon mucosa\\
normal colorectal epithelium\\
benign colon mucosal tissue} \\
\midrule
Cancer-associated stroma & \classnames{cancer associated stroma\\
stromal tissue around tumor\\
fibrovascular stroma\\
tumor stroma\\
stromal connective tissue} \\
\midrule
Colorectal adenocarcinoma epithelium & \classnames{colorectal adenocarcinoma epithelium\\
tumor epithelium\\
colorectal cancer epithelium\\
adenocarcinoma tissue\\
malignant colorectal glandular epithelium} \\
\bottomrule
\end{tabular}}}
\caption{\textbf{Class prompts for CRC-100K ROI zero-shot classification.} Continued.}
\label{tab:prompt_crc100k_v2}
\end{table*}

\begin{table*}[htbp]
\centering
\resizebox{0.9\textwidth}{!}{
\renewcommand{\arraystretch}{1.}{
\begin{tabular}{p{0.28\textwidth}|p{0.60\textwidth}}
\toprule
Class & Class names \\
\midrule
MSI-high colorectal cancer & \classnames{colorectal cancer with microsatellite instability high status\\
MSI high colorectal adenocarcinoma\\
microsatellite instability high colorectal tumor\\
MMR deficient colorectal cancer\\
MSI H colorectal carcinoma} \\
\midrule
Non-MSI-high colorectal cancer & \classnames{colorectal cancer without microsatellite instability high status\\
microsatellite stable or low instability colorectal cancer\\
non MSI high colorectal adenocarcinoma\\
MSS or MSI low colorectal tumor\\
non MSIH colorectal carcinoma} \\
\bottomrule
\end{tabular}}}
\caption{\textbf{Class prompts for CRC-MSI ROI zero-shot classification.}}
\label{tab:prompt_crc_msi}
\end{table*}

\begin{table*}[htbp]
\centering
\resizebox{0.95\textwidth}{!}{
\renewcommand{\arraystretch}{1.}{
\begin{tabular}{p{0.30\textwidth}|p{0.60\textwidth}}
\toprule
Class & Class names \\
\midrule
Esophageal adventitia & \classnames{esophageal adventitia\\
adventitial connective tissue\\
outer adventitial layer of the esophagus\\
adventitia around the esophagus\\
esophageal outer fibrous tissue} \\
\midrule
Lamina propria & \classnames{lamina propria\\
lamina propria of the esophagus\\
mucosal lamina propria\\
loose connective tissue of the lamina propria\\
esophageal lamina propria tissue} \\
\midrule
Muscularis mucosae & \classnames{muscularis mucosae\\
esophageal muscularis mucosae\\
thin smooth muscle of the mucosa\\
muscularis mucosa layer\\
mucosal muscle layer} \\
\midrule
Muscularis propria & \classnames{muscularis propria\\
esophageal muscularis propria\\
thick muscle wall of the esophagus\\
muscularis propria layer\\
smooth muscle of the esophageal wall} \\
\midrule
Regressive tumor tissue & \classnames{regressive tumor tissue\\
tumor regression bed\\
regressed esophageal adenocarcinoma tissue\\
post treatment regressive tumor area\\
regression associated tumor tissue} \\
\midrule
Gastric mucosa & \classnames{gastric mucosa\\
stomach mucosa\\
gastric glandular mucosa\\
mucosa of the stomach\\
gastric epithelial mucosa} \\
\bottomrule
\end{tabular}}}
\caption{\textbf{Class prompts for ESCA-Tolkach ROI zero-shot classification.} Continued.}
\label{tab:prompt_esca_tolkach_v1}
\end{table*}
\begin{table*}[htbp]
\centering
\resizebox{0.95\textwidth}{!}{
\renewcommand{\arraystretch}{1.}{
\begin{tabular}{p{0.30\textwidth}|p{0.60\textwidth}}
\toprule
Class & Class names \\
\midrule
Squamous esophageal mucosa & \classnames{squamous esophageal mucosa\\
normal squamous epithelium of the esophagus\\
esophageal squamous mucosa\\
squamous epithelial esophageal tissue\\
non glandular esophageal mucosa} \\
\midrule
Submucosa & \classnames{submucosa\\
esophageal submucosa\\
submucosal connective tissue\\
submucosal layer\\
submucosal tissue of the esophagus} \\
\midrule
Submucosal glands & \classnames{submucosal glands\\
esophageal submucosal glands\\
glandular tissue in the submucosa\\
submucosal gland region\\
mucous glands of the submucosa} \\
\midrule
Esophageal adenocarcinoma & \classnames{esophageal adenocarcinoma\\
tumor tissue of esophageal adenocarcinoma\\
malignant esophageal gland forming tumor\\
esophageal tumor epithelium\\
adenocarcinoma of the esophagus} \\
\midrule
Ulcer & \classnames{ulcer\\
ulcerated tissue\\
ulcer bed\\
ulcer associated tissue damage\\
esophageal ulceration} \\
\bottomrule
\end{tabular}}}
\caption{\textbf{Class prompts for ESCA-Tolkach ROI zero-shot classification.} Continued.}
\label{tab:prompt_esca_tolkach_v2}
\end{table*}

\begin{table*}[htbp]
\centering
\resizebox{0.85\textwidth}{!}{
\renewcommand{\arraystretch}{1.}{
\begin{tabular}{p{0.28\textwidth}|p{0.60\textwidth}}
\toprule
Class & Class names \\
\midrule
Adipose tissue & \classnames{adipose tissue\\
fat tissue in gastric histology\\
adipocytes\\
gastric adipose tissue\\
fat rich tissue region} \\
\midrule
Debris & \classnames{debris\\
necrotic debris\\
cellular debris in gastric tissue\\
degenerated tissue debris\\
tumor associated debris} \\
\midrule
Lymphocytes & \classnames{lymphocytes\\
lymphocyte rich tissue\\
immune cell rich region\\
lymphoid infiltrate\\
dense lymphocytic infiltrate} \\
\midrule
Mucus & \classnames{mucus\\
mucin rich tissue\\
mucus producing region\\
mucinous material\\
gastric mucus region} \\
\midrule
Smooth muscle & \classnames{smooth muscle\\
muscular tissue\\
smooth muscle region\\
muscle layer tissue\\
gastric smooth muscle} \\
\midrule
Normal gastric mucosa & \classnames{normal gastric mucosa\\
histologically normal stomach mucosa\\
non neoplastic gastric mucosa\\
normal gastric epithelium\\
benign stomach mucosal tissue} \\
\midrule
Cancer-associated stroma & \classnames{cancer associated stroma\\
stromal tissue around gastric tumor\\
fibrovascular stroma\\
tumor stroma\\
stromal connective tissue} \\
\midrule
Gastric tumor epithelium & \classnames{gastric tumor epithelium\\
gastric cancer tissue\\
tumor epithelium in stomach histology\\
malignant gastric epithelial tumor\\
gastric adenocarcinoma epithelium} \\
\bottomrule
\end{tabular}}}
\caption{\textbf{Class prompts for GCHTID ROI zero-shot classification.}}
\label{tab:prompt_gchtid}
\end{table*}

\begin{table*}[htbp]
\centering
\resizebox{0.9\textwidth}{!}{
\renewcommand{\arraystretch}{1.}{
\begin{tabular}{p{0.28\textwidth}|p{0.60\textwidth}}
\toprule
Class & Class names \\
\midrule
Non-tumor tissue & \classnames{non tumor osteosarcoma bed tissue\\
background bone or soft tissue without tumor\\
non neoplastic tissue in osteosarcoma histology\\
osteosarcoma slide region without tumor\\
benign non tumor tissue region} \\
\midrule
Non-viable osteosarcoma & \classnames{non viable osteosarcoma tumor\\
necrotic osteosarcoma tissue\\
treated non viable tumor tissue\\
dead osteosarcoma tumor cells\\
non viable malignant osteoid tumor} \\
\midrule
Viable osteosarcoma & \classnames{viable osteosarcoma tumor\\
living osteosarcoma tissue\\
viable malignant osteoid tumor\\
active osteosarcoma tumor cells\\
residual viable osteosarcoma} \\
\bottomrule
\end{tabular}}}
\caption{\textbf{Class prompts for OTA ROI zero-shot classification.}}
\label{tab:prompt_ota}
\end{table*}

\begin{table*}[htbp]
\centering
\resizebox{0.95\textwidth}{!}{
\renewcommand{\arraystretch}{1.}{
\begin{tabular}{p{0.32\textwidth}|p{0.58\textwidth}}
\toprule
Class & Class names \\
\midrule
Hyperplastic polyp & \classnames{hyperplastic polyp\\
colorectal hyperplastic polyp\\
hyperplastic polyp tissue\\
benign hyperplastic colorectal lesion\\
hyperplastic mucosal polyp} \\
\midrule
Normal colorectal tissue & \classnames{normal colorectal tissue\\
histologically normal colon mucosa\\
normal colorectal mucosa\\
non neoplastic colorectal tissue\\
benign normal colorectal epithelium} \\
\midrule
Tubular adenoma with high-grade dysplasia & \classnames{tubular adenoma with high grade dysplasia\\
high grade tubular adenoma\\
colorectal tubular adenoma high grade\\
tubular adenoma showing severe dysplasia\\
high grade dysplastic tubular adenoma} \\
\midrule
Tubular adenoma with low-grade dysplasia & \classnames{tubular adenoma with low grade dysplasia\\
low grade tubular adenoma\\
colorectal tubular adenoma low grade\\
tubular adenoma showing mild dysplasia\\
low grade dysplastic tubular adenoma} \\
\midrule
Tubulovillous adenoma with high-grade dysplasia & \classnames{tubulovillous adenoma with high grade dysplasia\\
high grade tubulovillous adenoma\\
colorectal tubulovillous adenoma high grade\\
tubulovillous adenoma showing severe dysplasia\\
high grade dysplastic tubulovillous adenoma} \\
\midrule
Tubulovillous adenoma with low-grade dysplasia & \classnames{tubulovillous adenoma with low grade dysplasia\\
low grade tubulovillous adenoma\\
colorectal tubulovillous adenoma low grade\\
tubulovillous adenoma showing mild dysplasia\\
low grade dysplastic tubulovillous adenoma} \\
\bottomrule
\end{tabular}}}
\caption{\textbf{Class prompts for Unitopatho ROI zero-shot classification.}}
\label{tab:prompt_unitopatho}
\end{table*}


\begin{table}[htbp]
\centering
\resizebox{0.85\textwidth}{!}{
\renewcommand{\arraystretch}{1.}{
\begin{tabular}{p{0.18\textwidth}p{0.68\textwidth}}
\toprule
Template ID & Prompt template \\
\midrule
1 & \{label\_text\}. \\
2 & a histopathological image showing \{label\_text\}. \\
3 & a histopathological image of \{label\_text\}. \\
4 & an H\&E stained image of \{label\_text\}. \\
\bottomrule
\end{tabular}}}
\caption{\textbf{Generic prompt templates for MI-Zero WSI zero-shot classification.}}
\label{tab:mizero_prompt_templates}
\end{table}

\begin{table}[htbp]
\centering
\resizebox{0.85\textwidth}{!}{
\renewcommand{\arraystretch}{1.}{
\begin{tabular}{p{0.30\textwidth}|p{0.56\textwidth}}
\toprule
Class & Class names \\
\midrule
Negative lymph node & \classnames{negative for tumor in lymph node\\
benign lymph node tissue\\
lymph node without metastatic carcinoma} \\
\midrule
Tumor-positive lymph node & \classnames{lymph node with metastatic carcinoma\\
tumor-positive lymph node\\
lymph node involved by carcinoma} \\
\bottomrule
\end{tabular}}}
\caption{\textbf{Class prompts for MI-Zero lymph node metastasis screening on CAMELYON+.}}
\label{tab:mizero_camelyon_plus}
\end{table}

\begin{table}[htbp]
\centering
\resizebox{0.85\textwidth}{!}{
\renewcommand{\arraystretch}{1.}{
\begin{tabular}{p{0.34\textwidth}|p{0.52\textwidth}}
\toprule
Class & Class names \\
\midrule
Lung adenocarcinoma & \classnames{lung adenocarcinoma\\
adenocarcinoma of the lung\\
LUAD lung adenocarcinoma} \\
\midrule
Lung squamous cell carcinoma & \classnames{lung squamous cell carcinoma\\
squamous cell carcinoma of the lung\\
LSCC lung squamous cell carcinoma} \\
\bottomrule
\end{tabular}}}
\caption{\textbf{Class prompts for MI-Zero non-small cell lung cancer subtyping on CPTAC-NSCLC.}}
\label{tab:mizero_cptac_nsclc}
\end{table}

\begin{table}[htbp]
\centering
\resizebox{0.9\textwidth}{!}{
\renewcommand{\arraystretch}{1.}{
\begin{tabular}{p{0.32\textwidth}|p{0.56\textwidth}}
\toprule
Class & Class names \\
\midrule
Diffuse glioma & \classnames{diffuse glioma\\
diffuse glioma in CNS histopathology\\
diffuse glioma brain tumor} \\
\midrule
Hematolymphoid or histiocytic tumor & \classnames{hematolymphoid or histiocytic tumor\\
hematolymphoid or histiocytic tumor in CNS histopathology\\
hematolymphoid or histiocytic brain tumor} \\
\midrule
Meningioma & \classnames{meningioma\\
meningioma in CNS histopathology\\
meningioma brain tumor} \\
\midrule
Mesenchymal or lipomatous tumor & \classnames{mesenchymal or lipomatous tumor\\
mesenchymal or lipomatous tumor in CNS histopathology\\
mesenchymal or lipomatous brain tumor} \\
\midrule
Metastatic tumor & \classnames{metastatic tumor\\
metastatic tumor in CNS histopathology\\
metastatic brain tumor} \\
\midrule
Nerve sheath tumor & \classnames{nerve sheath tumor\\
nerve sheath tumor in CNS histopathology\\
nerve sheath brain tumor} \\
\midrule
Other neuroepithelial tumor & \classnames{other neuroepithelial tumor\\
other neuroepithelial tumor in CNS histopathology\\
other neuroepithelial brain tumor} \\
\midrule
Sellar-region tumor & \classnames{sellar-region tumor\\
sellar-region tumor in CNS histopathology\\
sellar-region brain tumor} \\
\midrule
Vascular or mesenchymal tumor & \classnames{vascular or mesenchymal tumor\\
vascular or mesenchymal tumor in CNS histopathology\\
vascular or mesenchymal brain tumor} \\
\bottomrule
\end{tabular}}}
\caption{\textbf{Class prompts for MI-Zero CNS tumor-family classification on EBRAINS.}}
\label{tab:mizero_ebrains_cns_family}
\end{table}

\clearpage

\begin{table}[htbp]
\centering
\resizebox{0.85\textwidth}{!}{
\renewcommand{\arraystretch}{1.}{
\begin{tabular}{p{0.30\textwidth}|p{0.56\textwidth}}
\toprule
Class & Class names \\
\midrule
Hepatoblastoma & \classnames{hepatoblastoma\\
pediatric liver tumor hepatoblastoma\\
embryonal liver tumor hepatoblastoma} \\
\midrule
Medulloblastoma & \classnames{medulloblastoma\\
cerebellar medulloblastoma\\
embryonal cerebellar tumor medulloblastoma} \\
\midrule
Nephroblastoma & \classnames{nephroblastoma\\
Wilms tumor of the kidney\\
pediatric kidney tumor nephroblastoma} \\
\midrule
Neuroblastoma & \classnames{neuroblastoma\\
adrenal neuroblastoma\\
peripheral neuroblastic tumor neuroblastoma} \\
\bottomrule
\end{tabular}}}
\caption{\textbf{Class prompts for MI-Zero pediatric rare tumor classification on KidRare.}}
\label{tab:mizero_kidrare}
\end{table}

\begin{table}[htbp]
\centering
\resizebox{0.9\textwidth}{!}{
\renewcommand{\arraystretch}{1.}{
\begin{tabular}{p{0.30\textwidth}|p{0.58\textwidth}}
\toprule
Class & Class names \\
\midrule
Breast carcinoma & \classnames{breast carcinoma\\
breast cancer\\
malignant breast tumor} \\
\midrule
Gastric carcinoma & \classnames{gastric carcinoma\\
gastric cancer\\
stomach cancer} \\
\midrule
Renal cell carcinoma & \classnames{renal cell carcinoma\\
kidney cancer\\
renal tumor} \\
\midrule
Hepatocellular carcinoma & \classnames{hepatocellular carcinoma\\
liver cancer\\
hepatic malignancy} \\
\midrule
Lung carcinoma & \classnames{lung carcinoma\\
lung cancer\\
pulmonary malignancy} \\
\midrule
Neuroendocrine tumor & \classnames{neuroendocrine tumor\\
neuroendocrine carcinoma\\
neuroendocrine malignancy} \\
\midrule
Normal bone marrow & \classnames{normal bone marrow\\
benign bone marrow tissue\\
non-neoplastic bone marrow} \\
\midrule
Prostate adenocarcinoma & \classnames{prostate adenocarcinoma\\
prostate cancer\\
prostatic carcinoma} \\
\midrule
Squamous cell carcinoma & \classnames{squamous cell carcinoma\\
squamous carcinoma\\
malignant squamous tumor} \\
\midrule
Thyroid carcinoma & \classnames{thyroid carcinoma\\
thyroid cancer\\
thyroid malignancy} \\
\bottomrule
\end{tabular}}}
\caption{\textbf{Class prompts for MI-Zero bone metastasis primary site prediction on SYSBM.}}
\label{tab:mizero_sysbm}
\end{table}

\begin{table}[htbp]
\centering
\resizebox{0.9\textwidth}{!}{
\renewcommand{\arraystretch}{1.}{
\begin{tabular}{p{0.34\textwidth}|p{0.54\textwidth}}
\toprule
Class & Class names \\
\midrule
Ovarian clear cell carcinoma & \classnames{ovarian clear cell carcinoma\\
clear cell carcinoma of the ovary\\
ovarian clear cell adenocarcinoma} \\
\midrule
Ovarian endometrioid carcinoma & \classnames{ovarian endometrioid carcinoma\\
endometrioid carcinoma of the ovary\\
ovarian endometrioid adenocarcinoma} \\
\midrule
High-grade serous ovarian carcinoma & \classnames{high-grade serous ovarian carcinoma\\
high-grade serous carcinoma of the ovary\\
ovarian high-grade serous carcinoma} \\
\midrule
Low-grade serous ovarian carcinoma & \classnames{low-grade serous ovarian carcinoma\\
low-grade serous carcinoma of the ovary\\
ovarian low-grade serous carcinoma} \\
\midrule
Ovarian mucinous carcinoma & \classnames{ovarian mucinous carcinoma\\
mucinous carcinoma of the ovary\\
ovarian mucinous adenocarcinoma} \\
\bottomrule
\end{tabular}}}
\caption{\textbf{Class prompts for MI-Zero ovarian carcinoma subtyping on UBC-OCEAN.}}
\label{tab:mizero_ubc_ocean}
\end{table}


\begin{table*}[htbp]
\centering
\resizebox{0.5\textwidth}{!}{
\renewcommand{\arraystretch}{1.2}{
\begin{tabular}{ll}
\toprule
  Clinical variable & Text segment template \\
\midrule
  Age & Patient age is \texttt{[x]} years. \\
  Tumour size & Tumour size is \texttt{[x]} cm. \\
  Tumour type & Tumour type is \texttt{[x]}. \\
  ER status & ER status is \texttt{[x]}. \\
  PR status & PR status is \texttt{[x]}. \\
  HER2 status & HER2 status is \texttt{[x]}. \\
  HER2 expression & HER2 expression score is \texttt{[x]}. \\
  Histological grade & Histological grade is \texttt{[x]}. \\
  Ki-67 index & Ki-67 index is \texttt{[x]}. \\
  Molecular subtype & Molecular subtype is \texttt{[x]}. \\
\bottomrule
\end{tabular}}}
\caption{Clinical text segment templates for the breast cancer axillary lymph node metastasis (BCNB, 2 classes and 3 classes). The placeholder \texttt{[x]} represents the specific value of each clinical variable.}
\label{tab:clinical_text_templates_bcnb_aln_2class}
\end{table*}

\begin{table*}[htbp]
\centering
\resizebox{0.7\textwidth}{!}{
\renewcommand{\arraystretch}{1.2}{
\begin{tabular}{ll}
\toprule
  Clinical variable & Text segment template \\
\midrule
  Age at initial diagnosis & Patient age at initial diagnosis is \texttt{[x]} years. \\
  Sex & Patient sex is \texttt{[x]}. \\
  Smoking status & Smoking status is \texttt{[x]}. \\
  Primary tumour site & Primary tumour site is \texttt{[x]}. \\
  HPV association by p16 & HPV association by p16 is \texttt{[x]}. \\
\bottomrule
\end{tabular}}}
\caption{Clinical text segment templates for the head and neck keratinizing SCC grading (Hancock, 2 classes). The placeholder \texttt{[x]} represents the specific value of each clinical variable.}
\label{tab:clinical_text_templates_hancock_grading_scc_conventional_keratinizing_2class}
\end{table*}

\begin{table*}[htbp]
\centering
\resizebox{0.7\textwidth}{!}{
\renewcommand{\arraystretch}{1.2}{
\begin{tabular}{ll}
\toprule
  Clinical variable & Text segment template \\
\midrule
  Age at initial diagnosis & Patient age at initial diagnosis is \texttt{[x]} years. \\
  Sex & Patient sex is \texttt{[x]}. \\
  Smoking status & Smoking status is \texttt{[x]}. \\
  Primary tumour site & Primary tumour site is \texttt{[x]}. \\
  Histologic type & Histologic type is \texttt{[x]}. \\
  Histologic grade & Histologic grade is \texttt{[x]}. \\
  HPV association by p16 & HPV association by p16 is \texttt{[x]}. \\
\bottomrule
\end{tabular}}}
\caption{Clinical text segment templates for the head and neck lymphovascular invasion detection (Hancock, 2 classes) and vascular invasion detection (Hancock, 2 classes). The placeholder \texttt{[x]} represents the specific value of each clinical variable.}
\label{tab:clinical_text_templates_hancock_lymphovascular_invasion_l_2class}
\end{table*}

\begin{table*}[htbp]
\centering
\resizebox{0.7\textwidth}{!}{
\renewcommand{\arraystretch}{1.2}{
\begin{tabular}{ll}
\toprule
  Clinical variable & Text segment template \\
\midrule
  Age & Patient age is \texttt{[x]} years. \\
  Ethnicity & Patient ethnicity is \texttt{[x]}. \\
  Race & Patient race is \texttt{[x]}. \\
  Neo-adjuvant treatment status & Neo-adjuvant treatment status is \texttt{[x]}. \\
  Other cancer diagnosis & Other cancer diagnosis is \texttt{[x]}. \\
  Tumour type & Tumour type is \texttt{[x]}. \\
  Tumour location & Tumour location is \texttt{[x]}. \\
  Tumour location group & Tumour location group is \texttt{[x]}. \\
  Histological grade & Histological grade is \texttt{[x]}. \\
  Tumour stage & Tumour stage is \texttt{[x]}. \\
  Tumour substage & Tumour substage is \texttt{[x]}. \\
\bottomrule
\end{tabular}}}
\caption{Clinical text segment templates for the ovarian cancer treatment response prediction (PTRC-HGSOC, 2 classes). The placeholder \texttt{[x]} represents the specific value of each clinical variable.}
\label{tab:clinical_text_templates_ptrc_hgsoc_response_2class}
\end{table*}

\begin{table*}[htbp]
\centering
\resizebox{0.7\textwidth}{!}{
\renewcommand{\arraystretch}{1.2}{
\begin{tabular}{ll}
\toprule
  Clinical variable & Text segment template \\
\midrule
  Age & Patient age is \texttt{[x]} years. \\
  Ethnicity & Patient ethnicity is \texttt{[x]}. \\
  Race & Patient race is \texttt{[x]}. \\
  Neo-adjuvant treatment status & Neo-adjuvant treatment status is \texttt{[x]}. \\
  Other cancer diagnosis & Other cancer diagnosis is \texttt{[x]}. \\
  Histological grade & Histological grade is \texttt{[x]}. \\
  Tumour stage & Tumour stage is \texttt{[x]}. \\
  Tumour substage & Tumour substage is \texttt{[x]}. \\
\bottomrule
\end{tabular}}}
\caption{Clinical text segment templates for the ovarian cancer tumor type classification (PTRC-HGSOC, 2 classes). The placeholder \texttt{[x]} represents the specific value of each clinical variable.}
\label{tab:clinical_text_templates_ptrc_hgsoc_tumour_type_2class}
\end{table*}

\begin{table*}[htbp]
\centering
\resizebox{0.7\textwidth}{!}{
\renewcommand{\arraystretch}{1.2}{
\begin{tabular}{ll}
\toprule
  Clinical variable & Text segment template \\
\midrule
  Age at initial diagnosis interval & Patient age at initial diagnosis is \texttt{[x]} years. \\
  Sex & Patient sex is \texttt{[x]}. \\
  Breslow tumor thickness category & Breslow tumor thickness category is \texttt{[x]}. \\
  Epidermal ulceration & Epidermal ulceration is \texttt{[x]}. \\
  Histological melanoma type & Histological melanoma type is \texttt{[x]}. \\
  Primary melanoma body site & Primary melanoma body site is \texttt{[x]}. \\
\bottomrule
\end{tabular}}}
\caption{Clinical text segment templates for the melanoma relapse prediction without prior melanoma (Visiomel, 2 classes) and overall relapse prediction (Visiomel, 2 classes). The placeholder \texttt{[x]} represents the specific value of each clinical variable.}
\label{tab:clinical_text_templates_visiomel_relapse_noprev_melanoma_2class}
\end{table*}


\begin{table*}[htbp]
\centering
\resizebox{1\columnwidth}{!}{
\renewcommand{\arraystretch}{1.2}{
\begin{tabular}{lccccc}
\toprule
  Dataset & UNI-2 & Virchow-2 & H-Opt-1 & GPFM & \textbf{ALICE} \\
\midrule
  AGGC (5 classes) & 0.6565$\pm$0.0012 & 0.6573$\pm$0.0008 & 0.6549$\pm$0.0016 & \underline{0.6615$\pm$0.0008} & \textbf{0.6651$\pm$0.0015} \\
  BRACS (3 classes) & 0.7290$\pm$0.0504 & \underline{0.7384$\pm$0.0067} & 0.7084$\pm$0.0087 & 0.6611$\pm$0.0202 & \textbf{0.7544$\pm$0.0064} \\
  BRACS (7 classes) & \underline{0.5625$\pm$0.0055} & 0.5553$\pm$0.0044 & 0.4285$\pm$0.0058 & 0.4666$\pm$0.0078 & \textbf{0.6007$\pm$0.0035} \\
  BreakHis (2 classes) & \underline{0.8367$\pm$0.0352} & 0.7854$\pm$0.0550 & 0.7412$\pm$0.0701 & 0.5353$\pm$0.0488 & \textbf{0.8850$\pm$0.0026} \\
  CCRCC (6 classes) & 0.8407$\pm$0.0014 & \underline{0.8857$\pm$0.0009} & 0.8750$\pm$0.0009 & 0.8616$\pm$0.0011 & \textbf{0.9065$\pm$0.0009} \\
  Chaoyang (4 classes) & 0.6498$\pm$0.0301 & \textbf{0.7047$\pm$0.0039} & 0.6818$\pm$0.0234 & 0.5292$\pm$0.0967 & \underline{0.6856$\pm$0.0538} \\
  CRC-100K (9 classes) & \underline{0.9641$\pm$0.0005} & \textbf{0.9650$\pm$0.0005} & 0.9531$\pm$0.0004 & 0.9495$\pm$0.0001 & 0.9611$\pm$0.0008 \\
  CRC-MSI (2 classes) & \underline{0.7751$\pm$0.0083} & 0.7529$\pm$0.0030 & 0.7636$\pm$0.0006 & 0.7293$\pm$0.0005 & \textbf{0.7825$\pm$0.0007} \\
  ESCA-Tolkach (11 classes) & 0.8467$\pm$0.0011 & \underline{0.8632$\pm$0.0014} & 0.8371$\pm$0.0015 & 0.8124$\pm$0.0007 & \textbf{0.8648$\pm$0.0012} \\
  GCHTID (8 classes) & 0.6128$\pm$0.0017 & \underline{0.6342$\pm$0.0003} & 0.6158$\pm$0.0020 & 0.6329$\pm$0.0022 & \textbf{0.6586$\pm$0.0008} \\
  OTA (3 classes) & 0.6884$\pm$0.0562 & \underline{0.7503$\pm$0.0274} & 0.4964$\pm$0.1523 & 0.3519$\pm$0.0243 & \textbf{0.7782$\pm$0.0617} \\
  Unitopatho (6 classes) & \underline{0.2243$\pm$0.0715} & 0.2142$\pm$0.0117 & 0.1919$\pm$0.0178 & 0.1821$\pm$0.0147 & \textbf{0.3632$\pm$0.1806} \\
\midrule
  Overall & 0.6989 & \underline{0.7089} & 0.6623 & 0.6144 & \textbf{0.7422} \\
\bottomrule
\end{tabular}}}
\caption{\textbf{ROI-level classification} performance across twelve benchmarks under \textbf{linear probing}. Balanced accuracy was evaluated on the held-out test set for each of five train–validation splits and is reported as the mean ± standard deviation. \textbf{Bold} indicates the best-performing model, and \underline{underline} indicates the second-best-performing model for each dataset.}
\label{tab:roi_lp}
\end{table*}


\begin{table*}[htbp]
\centering
\resizebox{1\columnwidth}{!}{
\renewcommand{\arraystretch}{1.2}{
\begin{tabular}{lccccc}
\toprule
  Dataset & UNI-2 & Virchow-2 & H-Opt-1 & GPFM & \textbf{ALICE} \\
\midrule
  AGGC (5 classes) & \underline{0.6142$\pm$0.0038} & 0.6056$\pm$0.0075 & 0.6079$\pm$0.0032 & 0.6047$\pm$0.0069 & \textbf{0.6159$\pm$0.0028} \\
  BRACS (3 classes) & \underline{0.7633$\pm$0.0098} & 0.7493$\pm$0.0116 & 0.7423$\pm$0.0013 & 0.7533$\pm$0.0114 & \textbf{0.7633$\pm$0.0104} \\
  BRACS (7 classes) & 0.5888$\pm$0.0165 & 0.5888$\pm$0.0061 & 0.5453$\pm$0.0046 & \underline{0.6025$\pm$0.0088} & \textbf{0.6068$\pm$0.0087} \\
  BreakHis (2 classes) & \underline{0.9078$\pm$0.0003} & 0.8896$\pm$0.0060 & 0.8955$\pm$0.0024 & 0.8787$\pm$0.0014 & \textbf{0.9136$\pm$0.0029} \\
  CCRCC (6 classes) & 0.8333$\pm$0.0018 & \underline{0.8708$\pm$0.0030} & 0.8470$\pm$0.0028 & 0.8189$\pm$0.0036 & \textbf{0.8771$\pm$0.0005} \\
  Chaoyang (4 classes) & 0.7667$\pm$0.0084 & 0.7873$\pm$0.0044 & \underline{0.7874$\pm$0.0093} & 0.7642$\pm$0.0148 & \textbf{0.7924$\pm$0.0090} \\
  CRC-100K (9 classes) & \textbf{0.9552$\pm$0.0008} & \underline{0.9550$\pm$0.0004} & 0.9339$\pm$0.0003 & 0.9481$\pm$0.0012 & 0.9528$\pm$0.0013 \\
  CRC-MSI (2 classes) & 0.7139$\pm$0.0040 & 0.6963$\pm$0.0043 & \textbf{0.7189$\pm$0.0043} & 0.6749$\pm$0.0051 & \underline{0.7144$\pm$0.0045} \\
  ESCA-Tolkach (11 classes) & 0.8463$\pm$0.0007 & \textbf{0.8647$\pm$0.0015} & 0.8268$\pm$0.0017 & 0.8118$\pm$0.0016 & \underline{0.8611$\pm$0.0007} \\
  GCHTID (8 classes) & 0.6085$\pm$0.0126 & 0.6278$\pm$0.0108 & 0.6148$\pm$0.0129 & \textbf{0.6340$\pm$0.0111} & \underline{0.6333$\pm$0.0112} \\
  OTA (3 classes) & \underline{0.9470$\pm$0.0036} & 0.9378$\pm$0.0094 & \textbf{0.9531$\pm$0.0082} & 0.9422$\pm$0.0077 & 0.9466$\pm$0.0023 \\
  Unitopatho (6 classes) & 0.7981$\pm$0.0329 & \underline{0.8052$\pm$0.0398} & 0.7493$\pm$0.0486 & 0.7942$\pm$0.0453 & \textbf{0.8381$\pm$0.0450} \\
\midrule
  Overall & 0.7786 & \underline{0.7815} & 0.7685 & 0.7690 & \textbf{0.7930} \\
\bottomrule
\end{tabular}}}
\caption{\textbf{ROI-level classification} performance across twelve benchmarks under \textbf{KNN evaluation}. Balanced accuracy is reported as the mean ± standard deviation across six KNN settings, comprising cosine and Euclidean distance metrics with k = 5, 10 and 20. \textbf{Bold} indicates the best-performing model, and \underline{underline} indicates the second-best-performing model for each dataset.}
\label{tab:roi_knn_prism}
\end{table*}


\begin{table*}[htbp]
\centering
\resizebox{0.65\columnwidth}{!}{
\renewcommand{\arraystretch}{1.2}{
\begin{tabular}{lcccccc}
\toprule
  Dataset & UNI-2 & Virchow-2 & H-Opt-1 & GPFM & \textbf{ALICE} \\
\midrule
  AGGC (5 classes) & 0.8052 & 0.8017 & \underline{0.8068} & 0.7945 & \textbf{0.8122} \\
  BRACS (3 classes) & \underline{0.7579} & 0.7439 & 0.7561 & 0.7561 & \textbf{0.7596} \\
  BRACS (7 classes) & 0.5807 & 0.5667 & 0.5404 & \underline{0.5930} & \textbf{0.6018} \\
  BreakHis (2 classes) & \textbf{0.9099} & 0.9060 & 0.8932 & 0.8873 & \underline{0.9092} \\
  CCRCC (6 classes) & 0.9081 & \underline{0.9177} & 0.9116 & 0.8979 & \textbf{0.9227} \\
  Chaoyang (4 classes) & 0.8382 & 0.8471 & \underline{0.8537} & 0.8420 & \textbf{0.8560} \\
  CRC-100K (9 classes) & \textbf{0.9652} & 0.9596 & 0.9564 & 0.9628 & \underline{0.9631} \\
  CRC-MSI (2 classes) & 0.7616 & \underline{0.7621} & 0.7423 & 0.7307 & \textbf{0.7717} \\
  ESCA-Tolkach (11 classes) & 0.9402 & \underline{0.9513} & 0.9500 & 0.9320 & \textbf{0.9513} \\
  GCHTID (8 classes) & 0.5844 & \underline{0.6081} & 0.5890 & \textbf{0.6108} & 0.6073 \\
  OTA (3 classes) & \textbf{0.9543} & 0.9512 & 0.9421 & 0.9421 & \underline{0.9543} \\
  Unitopatho (6 classes) & 0.8675 & \underline{0.8744} & 0.8314 & 0.8656 & \textbf{0.8990} \\
\midrule
  \multicolumn{1}{c}{Overall} & 0.8228 & \underline{0.8242} & 0.8144 & 0.8179 & \textbf{0.8340} \\
\bottomrule
\end{tabular}}}
\caption{\textbf{ROI-level retrieval} performance across twelve benchmarks. Performance is reported as MVAcc@5. \textbf{Bold} indicates the best-performing model, and \underline{underline} indicates the second-best-performing model for each dataset.}
\label{tab:roi_retrieval_k5}
\end{table*}

\begin{table*}[htbp]
\centering
\resizebox{1\columnwidth}{!}{
\renewcommand{\arraystretch}{1.2}{
\begin{tabular}{llccccc}
\toprule
  Dataset & Metric & UNI-2 & Virchow-2 & H-Opt-1 & GPFM & \textbf{ALICE (Ours)} \\
\midrule
  CoNSeP & AP@50 & \textbf{0.2959$\pm$0.0060} & 0.2470$\pm$0.0079 & 0.2130$\pm$0.0020 & \underline{0.2604$\pm$0.0076} & 0.2580$\pm$0.0042 \\
  CRAG & AP@50 & \underline{0.8447$\pm$0.0113} & 0.8388$\pm$0.0152 & 0.8254$\pm$0.0172 & 0.8349$\pm$0.0124 & \textbf{0.8533$\pm$0.0108} \\
  GlaS & AP@50 & \underline{0.6803$\pm$0.0077} & 0.6617$\pm$0.0197 & 0.6640$\pm$0.0121 & 0.6728$\pm$0.0089 & \textbf{0.6857$\pm$0.0116} \\
  \midrule
  CoCaHis & DICE & \textbf{0.9464$\pm$0.0016} & 0.9438$\pm$0.0022 & 0.9399$\pm$0.0017 & 0.9444$\pm$0.0012 & \underline{0.9463$\pm$0.0013} \\
  COSAS24 & DICE & \underline{0.8988$\pm$0.0019} & 0.8899$\pm$0.0047 & 0.8900$\pm$0.0032 & 0.8959$\pm$0.0010 & \textbf{0.9004$\pm$0.0015} \\
  EBHI & DICE & \underline{0.8348$\pm$0.0125} & \textbf{0.8367$\pm$0.0146} & 0.8275$\pm$0.0284 & 0.8313$\pm$0.0244 & 0.8316$\pm$0.0147 \\
  Janowczyk & DICE & 0.6782$\pm$0.0156 & 0.6704$\pm$0.0092 & 0.6742$\pm$0.0199 & \underline{0.6824$\pm$0.0054} & \textbf{0.6886$\pm$0.0060} \\
  \midrule
  \multicolumn{1}{c}{Overall} & Instance (AP@50) & \textbf{0.6070} & 0.5825 & 0.5675 & 0.5894 & \underline{0.5990} \\
  \multicolumn{1}{c}{Overall} & Semantic (DICE) & \underline{0.8395} & 0.8352 & 0.8329 & 0.8385 & \textbf{0.8417} \\
\bottomrule
\end{tabular}}}
\caption{\textbf{ROI-level segmentation} performance across seven benchmarks using \textbf{Plain Mask Transformer (PMT)}. Instance segmentation is evaluated using AP@50, and semantic segmentation is evaluated using DICE. Results are reported as the mean ± standard deviation across five random seeds. \textbf{Bold} indicates the best-performing model, and \underline{underline} indicates the second-best-performing model for each task.}
\label{tab:roi_seg}
\end{table*}

\begin{table*}[htbp]
\centering
\resizebox{0.8\columnwidth}{!}{
\renewcommand{\arraystretch}{1.2}{
\begin{tabular}{lccccc}
\toprule
    Shot & UNI-2 & Virchow-2 & H-Opt-1 & GPFM & \textbf{ALICE} \\
\midrule
  1-shot & 0.4425$\pm$0.0949 & 0.3825$\pm$0.0798 & \underline{0.4573$\pm$0.0930} & 0.4518$\pm$0.0894 & \textbf{0.4719$\pm$0.0967} \\
  2-shot & 0.5090$\pm$0.0811 & 0.4232$\pm$0.0746 & \underline{0.5276$\pm$0.0792} & 0.5174$\pm$0.0786 & \textbf{0.5438$\pm$0.0803} \\
  4-shot & 0.5907$\pm$0.0673 & 0.4924$\pm$0.0674 & \underline{0.5950$\pm$0.0602} & 0.5882$\pm$0.0620 & \textbf{0.6196$\pm$0.0637} \\
  8-shot & \underline{0.6396$\pm$0.0606} & 0.5378$\pm$0.0636 & 0.6323$\pm$0.0528 & 0.6272$\pm$0.0556 & \textbf{0.6566$\pm$0.0560} \\
  16-shot & \underline{0.6754$\pm$0.0493} & 0.5681$\pm$0.0573 & 0.6556$\pm$0.0451 & 0.6549$\pm$0.0504 & \textbf{0.6846$\pm$0.0458} \\
  32-shot & \underline{0.6901$\pm$0.0431} & 0.5852$\pm$0.0497 & 0.6627$\pm$0.0442 & 0.6683$\pm$0.0468 & \textbf{0.6929$\pm$0.0444} \\
  64-shot & \underline{0.7004$\pm$0.0410} & 0.5931$\pm$0.0420 & 0.6662$\pm$0.0420 & 0.6784$\pm$0.0431 & \textbf{0.7021$\pm$0.0395} \\
\bottomrule
\end{tabular}}}
\caption{\textbf{AGGC (5 classes) ROI-level few-shot classification} performance under all-way \textbf{ProtoNet evaluation}. Results are reported as the mean ± standard deviation across 500 episodes. \textbf{Bold} indicates the best-performing model, and \underline{underline} indicates the second-best-performing model for each shot.}
\label{tab:roi_few_shot_protonet_aggc_5classes}
\end{table*}

\begin{table*}[htbp]
\centering
\resizebox{0.8\columnwidth}{!}{
\renewcommand{\arraystretch}{1.2}{
\begin{tabular}{lccccc}
\toprule
    Shot & UNI-2 & Virchow-2 & H-Opt-1 & GPFM & \textbf{ALICE} \\
\midrule
  1-shot & \textbf{0.5463$\pm$0.1066} & 0.4890$\pm$0.1009 & 0.4672$\pm$0.0984 & 0.4770$\pm$0.0941 & \underline{0.5409$\pm$0.1103} \\
  2-shot & \textbf{0.6176$\pm$0.0922} & 0.5508$\pm$0.0921 & 0.5076$\pm$0.0861 & 0.5195$\pm$0.0864 & \underline{0.6173$\pm$0.0907} \\
  4-shot & \textbf{0.6775$\pm$0.0714} & 0.6080$\pm$0.0820 & 0.5541$\pm$0.0749 & 0.5629$\pm$0.0740 & \underline{0.6751$\pm$0.0700} \\
  8-shot & \underline{0.7148$\pm$0.0619} & 0.6578$\pm$0.0756 & 0.5861$\pm$0.0677 & 0.5889$\pm$0.0692 & \textbf{0.7156$\pm$0.0630} \\
  16-shot & \underline{0.7334$\pm$0.0596} & 0.6929$\pm$0.0616 & 0.6068$\pm$0.0643 & 0.6084$\pm$0.0630 & \textbf{0.7366$\pm$0.0545} \\
  32-shot & \underline{0.7409$\pm$0.0545} & 0.7149$\pm$0.0545 & 0.6102$\pm$0.0616 & 0.6084$\pm$0.0596 & \textbf{0.7457$\pm$0.0523} \\
  64-shot & \underline{0.7401$\pm$0.0548} & 0.7210$\pm$0.0555 & 0.6089$\pm$0.0577 & 0.6064$\pm$0.0549 & \textbf{0.7463$\pm$0.0528} \\
  128-shot & \underline{0.7471$\pm$0.0518} & 0.7292$\pm$0.0526 & 0.6167$\pm$0.0538 & 0.6132$\pm$0.0598 & \textbf{0.7488$\pm$0.0540} \\
  256-shot & \textbf{0.7425$\pm$0.0548} & 0.7279$\pm$0.0532 & 0.6077$\pm$0.0584 & 0.6023$\pm$0.0599 & \underline{0.7411$\pm$0.0542} \\
\bottomrule
\end{tabular}}}
\caption{\textbf{BRACS (3 classes) ROI-level few-shot classification} performance under all-way \textbf{ProtoNet evaluation}. Results are reported as the mean ± standard deviation across 500 episodes. \textbf{Bold} indicates the best-performing model, and \underline{underline} indicates the second-best-performing model for each shot.}
\label{tab:roi_few_shot_protonet_bracs_3classes}
\end{table*}

\begin{table*}[htbp]
\centering
\resizebox{0.8\columnwidth}{!}{
\renewcommand{\arraystretch}{1.2}{
\begin{tabular}{lccccc}
\toprule
    Shot & UNI-2 & Virchow-2 & H-Opt-1 & GPFM & \textbf{ALICE} \\
\midrule
  1-shot & \textbf{0.3900$\pm$0.0625} & 0.3575$\pm$0.0616 & 0.3012$\pm$0.0609 & 0.3239$\pm$0.0593 & \underline{0.3845$\pm$0.0651} \\
  2-shot & \underline{0.4468$\pm$0.0529} & 0.4061$\pm$0.0551 & 0.3467$\pm$0.0574 & 0.3700$\pm$0.0583 & \textbf{0.4489$\pm$0.0562} \\
  4-shot & \underline{0.4961$\pm$0.0425} & 0.4578$\pm$0.0460 & 0.3969$\pm$0.0468 & 0.4223$\pm$0.0486 & \textbf{0.5056$\pm$0.0465} \\
  8-shot & \underline{0.5369$\pm$0.0403} & 0.4966$\pm$0.0421 & 0.4375$\pm$0.0393 & 0.4618$\pm$0.0436 & \textbf{0.5458$\pm$0.0378} \\
  16-shot & \underline{0.5608$\pm$0.0360} & 0.5219$\pm$0.0398 & 0.4586$\pm$0.0388 & 0.4836$\pm$0.0419 & \textbf{0.5688$\pm$0.0351} \\
  32-shot & \underline{0.5794$\pm$0.0330} & 0.5394$\pm$0.0367 & 0.4760$\pm$0.0405 & 0.4992$\pm$0.0388 & \textbf{0.5859$\pm$0.0354} \\
  64-shot & \underline{0.5881$\pm$0.0309} & 0.5475$\pm$0.0331 & 0.4787$\pm$0.0356 & 0.5025$\pm$0.0363 & \textbf{0.5914$\pm$0.0316} \\
  128-shot & \textbf{0.5954$\pm$0.0317} & 0.5517$\pm$0.0304 & 0.4805$\pm$0.0348 & 0.5076$\pm$0.0348 & \underline{0.5950$\pm$0.0309} \\
  256-shot & \textbf{0.5991$\pm$0.0311} & 0.5545$\pm$0.0314 & 0.4857$\pm$0.0341 & 0.5083$\pm$0.0331 & \underline{0.5932$\pm$0.0318} \\
\bottomrule
\end{tabular}}}
\caption{\textbf{BRACS (7 classes) ROI-level few-shot classification} performance under all-way \textbf{ProtoNet evaluation}. Results are reported as the mean ± standard deviation across 500 episodes. \textbf{Bold} indicates the best-performing model, and \underline{underline} indicates the second-best-performing model for each shot.}
\label{tab:roi_few_shot_protonet_bracs_7classes}
\end{table*}

\begin{table*}[htbp]
\centering
\resizebox{0.8\columnwidth}{!}{
\renewcommand{\arraystretch}{1.2}{
\begin{tabular}{lccccc}
\toprule
    Shot & UNI-2 & Virchow-2 & H-Opt-1 & GPFM & \textbf{ALICE} \\
\midrule
  1-shot & \textbf{0.6719$\pm$0.1243} & 0.5893$\pm$0.1346 & 0.5875$\pm$0.1074 & 0.5778$\pm$0.0986 & \underline{0.6705$\pm$0.1244} \\
  2-shot & \textbf{0.7356$\pm$0.1076} & 0.6255$\pm$0.1396 & 0.6336$\pm$0.1047 & 0.6206$\pm$0.0983 & \underline{0.7347$\pm$0.1033} \\
  4-shot & \underline{0.7833$\pm$0.0930} & 0.6737$\pm$0.1303 & 0.6759$\pm$0.0934 & 0.6602$\pm$0.0965 & \textbf{0.7860$\pm$0.0880} \\
  8-shot & \underline{0.8266$\pm$0.0747} & 0.7241$\pm$0.1155 & 0.7309$\pm$0.0943 & 0.7016$\pm$0.0912 & \textbf{0.8345$\pm$0.0702} \\
  16-shot & \underline{0.8448$\pm$0.0647} & 0.7645$\pm$0.1043 & 0.7780$\pm$0.0808 & 0.7406$\pm$0.0784 & \textbf{0.8544$\pm$0.0606} \\
  32-shot & \underline{0.8618$\pm$0.0599} & 0.8019$\pm$0.0808 & 0.8017$\pm$0.0705 & 0.7588$\pm$0.0703 & \textbf{0.8675$\pm$0.0546} \\
  64-shot & \underline{0.8692$\pm$0.0546} & 0.8232$\pm$0.0725 & 0.8283$\pm$0.0664 & 0.7783$\pm$0.0644 & \textbf{0.8749$\pm$0.0534} \\
  128-shot & \underline{0.8699$\pm$0.0539} & 0.8370$\pm$0.0590 & 0.8429$\pm$0.0570 & 0.7768$\pm$0.0624 & \textbf{0.8778$\pm$0.0512} \\
  256-shot & \underline{0.8696$\pm$0.0555} & 0.8423$\pm$0.0585 & 0.8530$\pm$0.0584 & 0.7827$\pm$0.0633 & \textbf{0.8765$\pm$0.0531} \\
\bottomrule
\end{tabular}}}
\caption{\textbf{BreakHis (2 classes ROI-level few-shot classification)} performance under all-way \textbf{ProtoNet evaluation}. Results are reported as the mean ± standard deviation across 500 episodes. \textbf{Bold} indicates the best-performing model, and \underline{underline} indicates the second-best-performing model for each shot.}
\label{tab:roi_few_shot_protonet_breakhis_2classes}
\end{table*}

\begin{table*}[htbp]
\centering
\resizebox{0.8\columnwidth}{!}{
\renewcommand{\arraystretch}{1.2}{
\begin{tabular}{lccccc}
\toprule
    Shot & UNI-2 & Virchow-2 & H-Opt-1 & GPFM & \textbf{ALICE} \\
\midrule
  1-shot & 0.5069$\pm$0.0896 & 0.5321$\pm$0.0866 & \underline{0.6183$\pm$0.0858} & 0.5301$\pm$0.0847 & \textbf{0.6324$\pm$0.0862} \\
  2-shot & 0.5935$\pm$0.0715 & 0.6079$\pm$0.0656 & \underline{0.6893$\pm$0.0641} & 0.6128$\pm$0.0752 & \textbf{0.7078$\pm$0.0667} \\
  4-shot & 0.6654$\pm$0.0577 & 0.6750$\pm$0.0538 & \underline{0.7416$\pm$0.0491} & 0.6875$\pm$0.0616 & \textbf{0.7674$\pm$0.0506} \\
  8-shot & 0.6995$\pm$0.0519 & 0.7155$\pm$0.0478 & \underline{0.7623$\pm$0.0457} & 0.7341$\pm$0.0552 & \textbf{0.7951$\pm$0.0452} \\
  16-shot & 0.7134$\pm$0.0480 & 0.7459$\pm$0.0469 & \underline{0.7744$\pm$0.0440} & 0.7576$\pm$0.0526 & \textbf{0.8132$\pm$0.0423} \\
  32-shot & 0.7158$\pm$0.0430 & 0.7666$\pm$0.0436 & \underline{0.7780$\pm$0.0427} & 0.7626$\pm$0.0478 & \textbf{0.8189$\pm$0.0380} \\
  64-shot & 0.7195$\pm$0.0415 & 0.7782$\pm$0.0386 & \underline{0.7815$\pm$0.0378} & 0.7670$\pm$0.0425 & \textbf{0.8241$\pm$0.0326} \\
  128-shot & 0.7135$\pm$0.0378 & \underline{0.7850$\pm$0.0353} & 0.7816$\pm$0.0371 & 0.7622$\pm$0.0367 & \textbf{0.8248$\pm$0.0326} \\
  256-shot & 0.7153$\pm$0.0325 & \underline{0.7931$\pm$0.0321} & 0.7845$\pm$0.0331 & 0.7641$\pm$0.0340 & \textbf{0.8289$\pm$0.0315} \\
\bottomrule
\end{tabular}}}
\caption{\textbf{CCRCC (6 classes) ROI-level few-shot classification} performance under all-way \textbf{ProtoNet evaluation}. Results are reported as the mean ± standard deviation across 500 episodes. \textbf{Bold} indicates the best-performing model, and \underline{underline} indicates the second-best-performing model for each shot.}
\label{tab:roi_few_shot_protonet_ccrcc_6classes}
\end{table*}

\begin{table*}[htbp]
\centering
\resizebox{0.8\columnwidth}{!}{
\renewcommand{\arraystretch}{1.2}{
\begin{tabular}{lccccc}
\toprule
    Shot & UNI-2 & Virchow-2 & H-Opt-1 & GPFM & \textbf{ALICE} \\
\midrule
  1-shot & \underline{0.5244$\pm$0.0953} & 0.5019$\pm$0.0896 & 0.5233$\pm$0.0937 & 0.4828$\pm$0.0933 & \textbf{0.5326$\pm$0.0944} \\
  2-shot & 0.5763$\pm$0.0791 & 0.5570$\pm$0.0750 & \textbf{0.5865$\pm$0.0768} & 0.5423$\pm$0.0829 & \underline{0.5827$\pm$0.0771} \\
  4-shot & 0.6098$\pm$0.0693 & 0.5992$\pm$0.0649 & \textbf{0.6245$\pm$0.0642} & 0.5909$\pm$0.0734 & \underline{0.6241$\pm$0.0655} \\
  8-shot & 0.6472$\pm$0.0585 & 0.6324$\pm$0.0589 & \textbf{0.6654$\pm$0.0562} & 0.6280$\pm$0.0603 & \underline{0.6594$\pm$0.0601} \\
  16-shot & 0.6726$\pm$0.0547 & 0.6661$\pm$0.0546 & \textbf{0.6928$\pm$0.0508} & 0.6566$\pm$0.0540 & \underline{0.6851$\pm$0.0537} \\
  32-shot & 0.6875$\pm$0.0480 & 0.6830$\pm$0.0494 & \textbf{0.7053$\pm$0.0452} & 0.6754$\pm$0.0488 & \underline{0.6995$\pm$0.0489} \\
  64-shot & 0.6958$\pm$0.0480 & 0.6961$\pm$0.0474 & \textbf{0.7128$\pm$0.0471} & 0.6780$\pm$0.0463 & \underline{0.7106$\pm$0.0462} \\
  128-shot & 0.7030$\pm$0.0457 & 0.7054$\pm$0.0484 & \textbf{0.7159$\pm$0.0471} & 0.6885$\pm$0.0483 & \underline{0.7141$\pm$0.0468} \\
  256-shot & 0.7118$\pm$0.0441 & 0.7114$\pm$0.0463 & \underline{0.7189$\pm$0.0454} & 0.6911$\pm$0.0465 & \textbf{0.7196$\pm$0.0446} \\
\bottomrule
\end{tabular}}}
\caption{\textbf{Chaoyang (4 classes) ROI-level few-shot classification} performance under all-way \textbf{ProtoNet evaluation}. Results are reported as the mean ± standard deviation across 500 episodes. \textbf{Bold} indicates the best-performing model, and \underline{underline} indicates the second-best-performing model for each shot.}
\label{tab:roi_few_shot_protonet_chaoyang_4classes}
\end{table*}

\begin{table*}[htbp]
\centering
\resizebox{0.8\columnwidth}{!}{
\renewcommand{\arraystretch}{1.2}{
\begin{tabular}{lccccc}
\toprule
    Shot & UNI-2 & Virchow-2 & H-Opt-1 & GPFM & \textbf{ALICE} \\
\midrule
  1-shot & \underline{0.8032$\pm$0.0643} & 0.7571$\pm$0.0820 & 0.7672$\pm$0.0567 & 0.8000$\pm$0.0576 & \textbf{0.8682$\pm$0.0504} \\
  2-shot & \underline{0.8710$\pm$0.0457} & 0.8323$\pm$0.0548 & 0.8281$\pm$0.0415 & 0.8691$\pm$0.0386 & \textbf{0.9167$\pm$0.0310} \\
  4-shot & \underline{0.9142$\pm$0.0289} & 0.8846$\pm$0.0394 & 0.8696$\pm$0.0345 & 0.9089$\pm$0.0261 & \textbf{0.9397$\pm$0.0206} \\
  8-shot & \underline{0.9348$\pm$0.0209} & 0.9087$\pm$0.0245 & 0.8911$\pm$0.0277 & 0.9268$\pm$0.0211 & \textbf{0.9502$\pm$0.0167} \\
  16-shot & \underline{0.9458$\pm$0.0171} & 0.9176$\pm$0.0218 & 0.9032$\pm$0.0226 & 0.9339$\pm$0.0182 & \textbf{0.9546$\pm$0.0143} \\
  32-shot & \underline{0.9522$\pm$0.0157} & 0.9232$\pm$0.0189 & 0.9118$\pm$0.0202 & 0.9376$\pm$0.0173 & \textbf{0.9579$\pm$0.0146} \\
  64-shot & \underline{0.9543$\pm$0.0149} & 0.9254$\pm$0.0181 & 0.9145$\pm$0.0204 & 0.9392$\pm$0.0171 & \textbf{0.9581$\pm$0.0139} \\
  128-shot & \underline{0.9551$\pm$0.0133} & 0.9256$\pm$0.0174 & 0.9163$\pm$0.0175 & 0.9409$\pm$0.0152 & \textbf{0.9588$\pm$0.0127} \\
  256-shot & \underline{0.9560$\pm$0.0137} & 0.9264$\pm$0.0180 & 0.9179$\pm$0.0177 & 0.9410$\pm$0.0157 & \textbf{0.9591$\pm$0.0137} \\
\bottomrule
\end{tabular}}}
\caption{\textbf{CRC-100K (9 classes) ROI-level few-shot classification} performance under all-way \textbf{ProtoNet evaluation}. Results are reported as the mean ± standard deviation across 500 episodes. \textbf{Bold} indicates the best-performing model, and \underline{underline} indicates the second-best-performing model for each shot.}
\label{tab:roi_few_shot_protonet_crc_100k_9classes}
\end{table*}

\begin{table*}[htbp]
\centering
\resizebox{0.8\columnwidth}{!}{
\renewcommand{\arraystretch}{1.2}{
\begin{tabular}{lccccc}
\toprule
    Shot & UNI-2 & Virchow-2 & H-Opt-1 & GPFM & \textbf{ALICE} \\
\midrule
  1-shot & \underline{0.5466$\pm$0.0902} & 0.5188$\pm$0.0725 & 0.5399$\pm$0.0859 & 0.5326$\pm$0.0840 & \textbf{0.5490$\pm$0.0866} \\
  2-shot & \underline{0.5746$\pm$0.0838} & 0.5231$\pm$0.0818 & 0.5698$\pm$0.0841 & 0.5466$\pm$0.0827 & \textbf{0.5764$\pm$0.0850} \\
  4-shot & \underline{0.6017$\pm$0.0893} & 0.5445$\pm$0.0798 & 0.5940$\pm$0.0882 & 0.5665$\pm$0.0877 & \textbf{0.6071$\pm$0.0895} \\
  8-shot & \underline{0.6467$\pm$0.0833} & 0.5678$\pm$0.0865 & 0.6315$\pm$0.0882 & 0.6010$\pm$0.0852 & \textbf{0.6483$\pm$0.0826} \\
  16-shot & \underline{0.6735$\pm$0.0807} & 0.5906$\pm$0.0835 & 0.6545$\pm$0.0824 & 0.6303$\pm$0.0824 & \textbf{0.6773$\pm$0.0797} \\
  32-shot & \underline{0.6884$\pm$0.0788} & 0.6161$\pm$0.0871 & 0.6773$\pm$0.0797 & 0.6435$\pm$0.0803 & \textbf{0.7013$\pm$0.0780} \\
  64-shot & \underline{0.7043$\pm$0.0736} & 0.6383$\pm$0.0816 & 0.6882$\pm$0.0710 & 0.6697$\pm$0.0787 & \textbf{0.7152$\pm$0.0709} \\
  128-shot & \underline{0.7060$\pm$0.0726} & 0.6488$\pm$0.0797 & 0.6879$\pm$0.0741 & 0.6680$\pm$0.0774 & \textbf{0.7176$\pm$0.0727} \\
  256-shot & \underline{0.7055$\pm$0.0706} & 0.6534$\pm$0.0747 & 0.6881$\pm$0.0732 & 0.6728$\pm$0.0714 & \textbf{0.7203$\pm$0.0682} \\
\bottomrule
\end{tabular}}}
\caption{\textbf{CRC-MSI (2 classes) ROI-level few-shot classification} performance under all-way \textbf{ProtoNet evaluation}. Results are reported as the mean ± standard deviation across 500 episodes. \textbf{Bold} indicates the best-performing model, and \underline{underline} indicates the second-best-performing model for each shot.}
\label{tab:roi_few_shot_protonet_crc_msi_2classes}
\end{table*}

\begin{table*}[htbp]
\centering
\resizebox{0.8\columnwidth}{!}{
\renewcommand{\arraystretch}{1.2}{
\begin{tabular}{lccccc}
\toprule
    Shot & UNI-2 & Virchow-2 & H-Opt-1 & GPFM & \textbf{ALICE} \\
\midrule
  1-shot & 0.6939$\pm$0.0494 & 0.5947$\pm$0.0568 & \underline{0.6998$\pm$0.0528} & 0.6582$\pm$0.0513 & \textbf{0.7394$\pm$0.0468} \\
  2-shot & 0.7625$\pm$0.0423 & 0.6797$\pm$0.0497 & \underline{0.7685$\pm$0.0400} & 0.7302$\pm$0.0427 & \textbf{0.7971$\pm$0.0385} \\
  4-shot & 0.8103$\pm$0.0317 & 0.7495$\pm$0.0373 & \underline{0.8167$\pm$0.0300} & 0.7825$\pm$0.0343 & \textbf{0.8363$\pm$0.0288} \\
  8-shot & 0.8432$\pm$0.0260 & 0.7918$\pm$0.0311 & \underline{0.8470$\pm$0.0236} & 0.8148$\pm$0.0286 & \textbf{0.8586$\pm$0.0240} \\
  16-shot & \underline{0.8633$\pm$0.0227} & 0.8164$\pm$0.0264 & 0.8628$\pm$0.0218 & 0.8331$\pm$0.0261 & \textbf{0.8711$\pm$0.0216} \\
  32-shot & 0.8729$\pm$0.0211 & 0.8229$\pm$0.0250 & \underline{0.8730$\pm$0.0205} & 0.8428$\pm$0.0257 & \textbf{0.8772$\pm$0.0207} \\
\bottomrule
\end{tabular}}}
\caption{\textbf{ESCA (11 classes) ROI-level few-shot classification} performance under all-way \textbf{ProtoNet evaluation}. Results are reported as the mean ± standard deviation across 500 episodes. \textbf{Bold} indicates the best-performing model, and \underline{underline} indicates the second-best-performing model for each shot.}
\label{tab:roi_few_shot_protonet_esca_11classes}
\end{table*}

\begin{table*}[htbp]
\centering
\resizebox{0.8\columnwidth}{!}{
\renewcommand{\arraystretch}{1.2}{
\begin{tabular}{lccccc}
\toprule
    Shot & UNI-2 & Virchow-2 & H-Opt-1 & GPFM & \textbf{ALICE} \\
\midrule
  1-shot & 0.2763$\pm$0.0511 & 0.2536$\pm$0.0514 & 0.2881$\pm$0.0541 & \underline{0.2947$\pm$0.0546} & \textbf{0.3124$\pm$0.0585} \\
  2-shot & 0.3364$\pm$0.0478 & 0.3029$\pm$0.0517 & 0.3469$\pm$0.0491 & \underline{0.3541$\pm$0.0487} & \textbf{0.3805$\pm$0.0526} \\
  4-shot & 0.4028$\pm$0.0490 & 0.3581$\pm$0.0507 & 0.4084$\pm$0.0466 & \underline{0.4200$\pm$0.0473} & \textbf{0.4422$\pm$0.0506} \\
  8-shot & 0.4647$\pm$0.0467 & 0.4085$\pm$0.0448 & 0.4692$\pm$0.0419 & \underline{0.4839$\pm$0.0434} & \textbf{0.5030$\pm$0.0441} \\
  16-shot & 0.5153$\pm$0.0413 & 0.4526$\pm$0.0414 & 0.5182$\pm$0.0395 & \underline{0.5356$\pm$0.0398} & \textbf{0.5485$\pm$0.0405} \\
  32-shot & 0.5411$\pm$0.0371 & 0.4774$\pm$0.0408 & 0.5480$\pm$0.0367 & \underline{0.5665$\pm$0.0370} & \textbf{0.5699$\pm$0.0363} \\
  64-shot & 0.5537$\pm$0.0380 & 0.4960$\pm$0.0401 & 0.5600$\pm$0.0378 & \textbf{0.5819$\pm$0.0383} & \underline{0.5785$\pm$0.0390} \\
  128-shot & 0.5602$\pm$0.0381 & 0.5060$\pm$0.0394 & 0.5694$\pm$0.0376 & \textbf{0.5903$\pm$0.0371} & \underline{0.5825$\pm$0.0381} \\
  256-shot & 0.5657$\pm$0.0384 & 0.5181$\pm$0.0393 & 0.5749$\pm$0.0401 & \textbf{0.5948$\pm$0.0392} & \underline{0.5878$\pm$0.0387} \\
\bottomrule
\end{tabular}}}
\caption{\textbf{GCHTID (8 classes) ROI-level few-shot classification} performance under all-way \textbf{ProtoNet evaluation}. Results are reported as the mean ± standard deviation across 500 episodes. \textbf{Bold} indicates the best-performing model, and \underline{underline} indicates the second-best-performing model for each shot.}
\label{tab:roi_few_shot_protonet_gchtid_8classes}
\end{table*}

\begin{table*}[htbp]
\centering
\resizebox{0.8\columnwidth}{!}{
\renewcommand{\arraystretch}{1.2}{
\begin{tabular}{lccccc}
\toprule
    Shot & UNI-2 & Virchow-2 & H-Opt-1 & GPFM & \textbf{ALICE} \\
\midrule
  1-shot & \underline{0.6323$\pm$0.1186} & 0.5490$\pm$0.1023 & 0.6212$\pm$0.1172 & 0.6147$\pm$0.1103 & \textbf{0.6466$\pm$0.1188} \\
  2-shot & \underline{0.7345$\pm$0.0992} & 0.6360$\pm$0.1045 & 0.7192$\pm$0.0963 & 0.7133$\pm$0.0921 & \textbf{0.7459$\pm$0.0989} \\
  4-shot & \underline{0.8118$\pm$0.0704} & 0.6964$\pm$0.1048 & 0.7977$\pm$0.0693 & 0.7980$\pm$0.0725 & \textbf{0.8244$\pm$0.0722} \\
  8-shot & \underline{0.8630$\pm$0.0560} & 0.7414$\pm$0.0994 & 0.8399$\pm$0.0587 & 0.8472$\pm$0.0592 & \textbf{0.8725$\pm$0.0534} \\
  16-shot & \underline{0.8825$\pm$0.0430} & 0.7652$\pm$0.0913 & 0.8609$\pm$0.0475 & 0.8766$\pm$0.0457 & \textbf{0.8986$\pm$0.0412} \\
  32-shot & \underline{0.8958$\pm$0.0411} & 0.7726$\pm$0.0823 & 0.8755$\pm$0.0454 & 0.8911$\pm$0.0399 & \textbf{0.9103$\pm$0.0382} \\
  64-shot & \underline{0.8982$\pm$0.0344} & 0.7666$\pm$0.0702 & 0.8767$\pm$0.0409 & 0.8939$\pm$0.0376 & \textbf{0.9150$\pm$0.0332} \\
  128-shot & \underline{0.8965$\pm$0.0379} & 0.7621$\pm$0.0537 & 0.8785$\pm$0.0381 & 0.8959$\pm$0.0363 & \textbf{0.9179$\pm$0.0328} \\
\bottomrule
\end{tabular}}}
\caption{\textbf{OTA (3 classes) ROI-level few-shot classification} performance under all-way \textbf{ProtoNet evaluation}. Results are reported as the mean ± standard deviation across 500 episodes. \textbf{Bold} indicates the best-performing model, and \underline{underline} indicates the second-best-performing model for each shot.}
\label{tab:roi_few_shot_protonet_ota_3classes}
\end{table*}

\begin{table*}[htbp]
\centering
\resizebox{0.8\columnwidth}{!}{
\renewcommand{\arraystretch}{1.2}{
\begin{tabular}{lccccc}
\toprule
    Shot & UNI-2 & Virchow-2 & H-Opt-1 & GPFM & \textbf{ALICE} \\
\midrule
  1-shot & 0.3449$\pm$0.0673 & \underline{0.3684$\pm$0.0683} & 0.3097$\pm$0.0642 & 0.2991$\pm$0.0584 & \textbf{0.3887$\pm$0.0722} \\
  2-shot & 0.4039$\pm$0.0702 & \underline{0.4262$\pm$0.0627} & 0.3627$\pm$0.0638 & 0.3446$\pm$0.0614 & \textbf{0.4561$\pm$0.0690} \\
  4-shot & 0.4766$\pm$0.0621 & \underline{0.4794$\pm$0.0538} & 0.4238$\pm$0.0565 & 0.4016$\pm$0.0554 & \textbf{0.5286$\pm$0.0564} \\
  8-shot & \underline{0.5377$\pm$0.0544} & 0.5135$\pm$0.0473 & 0.4755$\pm$0.0515 & 0.4482$\pm$0.0508 & \textbf{0.5793$\pm$0.0485} \\
  16-shot & \underline{0.5816$\pm$0.0478} & 0.5346$\pm$0.0480 & 0.5126$\pm$0.0452 & 0.4901$\pm$0.0456 & \textbf{0.6117$\pm$0.0418} \\
  32-shot & \underline{0.6114$\pm$0.0436} & 0.5432$\pm$0.0429 & 0.5336$\pm$0.0432 & 0.5156$\pm$0.0443 & \textbf{0.6308$\pm$0.0414} \\
  64-shot & \underline{0.6212$\pm$0.0409} & 0.5489$\pm$0.0421 & 0.5475$\pm$0.0385 & 0.5275$\pm$0.0420 & \textbf{0.6378$\pm$0.0391} \\
  128-shot & \underline{0.6308$\pm$0.0392} & 0.5544$\pm$0.0417 & 0.5534$\pm$0.0429 & 0.5342$\pm$0.0416 & \textbf{0.6389$\pm$0.0380} \\
  256-shot & \underline{0.6314$\pm$0.0404} & 0.5565$\pm$0.0384 & 0.5574$\pm$0.0402 & 0.5350$\pm$0.0423 & \textbf{0.6393$\pm$0.0384} \\
\bottomrule
\end{tabular}}}
\caption{\textbf{Unitopatho (6 classes) ROI-level few-shot classification} performance under all-way \textbf{ProtoNet evaluation}. Results are reported as the mean ± standard deviation across 500 episodes. \textbf{Bold} indicates the best-performing model, and \underline{underline} indicates the second-best-performing model for each shot.}
\label{tab:roi_few_shot_protonet_unitopatho_6classes}
\end{table*}


\begin{table*}[htbp]
\centering
\resizebox{1\columnwidth}{!}{
\renewcommand{\arraystretch}{1.2}{
\begin{tabular}{lccccc}
\toprule
  Dataset & UNI-2 & Virchow-2 & H-Opt-1 & GPFM & \textbf{ALICE} \\
\midrule
  \makecell[l]{Prostate cancer screening\\(PANDA, 2 classes)} & 0.9756$\pm$0.0037 & 0.9820$\pm$0.0017 & 0.9786$\pm$0.0028 & \underline{0.9825$\pm$0.0018} & \textbf{0.9835$\pm$0.0016} \\
  \makecell[l]{Lymph node metastasis screening\\(CAMELYON+, 2 classes)} & 0.9385$\pm$0.0093 & \underline{0.9458$\pm$0.0044} & 0.9438$\pm$0.0077 & 0.9446$\pm$0.0063 & \textbf{0.9488$\pm$0.0066} \\
  \makecell[l]{Lymph node metastatic burden classification\\(CAMELYON+, 4 classes)} & 0.6332$\pm$0.0062 & \underline{0.6448$\pm$0.0119} & 0.6136$\pm$0.0263 & 0.6351$\pm$0.0399 & \textbf{0.7046$\pm$0.0101} \\
  \makecell[l]{Colorectal neoplasia screening\\(HunCRC, 4 classes)} & 0.3701$\pm$0.0422 & \textbf{0.4909$\pm$0.0198} & \underline{0.4686$\pm$0.0233} & 0.4269$\pm$0.0280 & 0.4622$\pm$0.0732 \\
  \makecell[l]{Bone metastasis primary site prediction\\(SYSBM, 10 classes)} & 0.7007$\pm$0.0352 & \underline{0.7624$\pm$0.0169} & 0.7409$\pm$0.0320 & 0.6508$\pm$0.0380 & \textbf{0.8014$\pm$0.0335} \\
  \makecell[l]{Lung lesion triage\\(SYSFL+, 4 classes)} & 0.9464$\pm$0.0127 & 0.9397$\pm$0.0194 & \underline{0.9632$\pm$0.0203} & 0.9308$\pm$0.0126 & \textbf{0.9715$\pm$0.0202} \\
  \makecell[l]{Cervical epithelial lesion classification\\(TissueNet, 4 classes)} & 0.7518$\pm$0.0097 & \underline{0.7542$\pm$0.0139} & 0.7529$\pm$0.0324 & 0.7035$\pm$0.0217 & \textbf{0.7738$\pm$0.0230} \\
\bottomrule
\end{tabular}}}
\caption{\textbf{WSI-level classification performance} across 19 benchmarks using \textbf{vision-only PFM features with ABMIL}. Balanced accuracy is reported as the mean ± standard deviation. \textbf{Bold} indicates the best-performing model, and \underline{underline} indicates the second-best-performing model for each dataset.}
\label{tab:wsi_mil}
\end{table*}


\begin{table*}[htbp]
\centering
\resizebox{0.8\columnwidth}{!}{
\renewcommand{\arraystretch}{1.2}{
\begin{tabular}{lcccc}
\toprule
  Dataset & MUSK & KEEP & CONCH & \textbf{ALICE} \\
\midrule
  AGGC (5 classes) & \underline{0.4682$\pm$0.0564} & 0.4298$\pm$0.0771 & 0.3514$\pm$0.0597 & \textbf{0.5657$\pm$0.0635} \\
  BRACS (3 classes) & \textbf{0.7054$\pm$0.0454} & 0.5204$\pm$0.0521 & 0.4729$\pm$0.0676 & \underline{0.6014$\pm$0.0649} \\
  BRACS (7 classes) & \underline{0.4795$\pm$0.0296} & 0.4158$\pm$0.0582 & 0.4288$\pm$0.0739 & \textbf{0.5211$\pm$0.0471} \\
  BreakHis (2 classes) & 0.6595$\pm$0.0626 & \textbf{0.7824$\pm$0.0471} & \underline{0.7790$\pm$0.0690} & 0.7773$\pm$0.0657 \\
  CCRCC (6 classes) & \underline{0.5993$\pm$0.0458} & 0.5753$\pm$0.0528 & 0.5383$\pm$0.0772 & \textbf{0.6866$\pm$0.0838} \\
  Chaoyang (4 classes) & 0.4823$\pm$0.0426 & \underline{0.5644$\pm$0.0311} & 0.5287$\pm$0.0771 & \textbf{0.5685$\pm$0.0568} \\
  CRC-100K (9 classes) & 0.7403$\pm$0.0913 & \underline{0.8690$\pm$0.0869} & 0.7663$\pm$0.0719 & \textbf{0.9103$\pm$0.0445} \\
  CRC-MSI (2 classes) & \textbf{0.5582$\pm$0.0185} & 0.5081$\pm$0.0234 & 0.4995$\pm$0.0290 & \underline{0.5207$\pm$0.0273} \\
  ESCA-Tolkach (11 classes) & 0.5620$\pm$0.0294 & \underline{0.6438$\pm$0.0598} & \textbf{0.6494$\pm$0.0687} & 0.6274$\pm$0.0497 \\
  GCHTID (8 classes) & 0.4373$\pm$0.0325 & 0.3714$\pm$0.0320 & \underline{0.4504$\pm$0.0397} & \textbf{0.4538$\pm$0.0214} \\
  OTA (3 classes) & 0.5445$\pm$0.1118 & 0.6943$\pm$0.1369 & \underline{0.7239$\pm$0.1498} & \textbf{0.7625$\pm$0.1448} \\
  Unitopatho (6 classes) & \underline{0.4494$\pm$0.0273} & 0.4215$\pm$0.0399 & 0.2947$\pm$0.0585 & \textbf{0.4850$\pm$0.0544} \\
\midrule
  Overall & 0.5572 & \underline{0.5664} & 0.5403 & \textbf{0.6234} \\
\bottomrule
\end{tabular}}}
\caption{\textbf{ROI-level zero-shot classification} performance across twelve benchmarks. Balanced accuracy is reported with the per-prompt standard deviation. \textbf{Bold} indicates the best-performing model, and \underline{underline} indicates the second-best-performing model for each dataset.}
\label{tab:roi_zero_shot}
\end{table*}


\begin{table*}[htbp]
\centering
\resizebox{0.8\columnwidth}{!}{
\renewcommand{\arraystretch}{1.2}{
\begin{tabular}{llcccccccc}
\toprule
  Source & Recall@\textit{K} & \multicolumn{2}{c}{MUSK} & \multicolumn{2}{c}{KEEP} & \multicolumn{2}{c}{CONCH} & \multicolumn{2}{c}{\textbf{ALICE}} \\ \cline{3-10}
         &   & t2i         & i2t  & t2i         & i2t  & t2i         & i2t  & t2i         & i2t  \\
\midrule
  BookSet & \textit{K}=1 & \textbf{0.1805} & \textbf{0.1784} & 0.0921 & 0.0759 & \underline{0.1583} & \underline{0.1598} & 0.1448 & 0.0966 \\
   & \textit{K}=5 & \textbf{0.3970} & \textbf{0.4024} & 0.2771 & 0.2135 & 0.3550 & \underline{0.3298} & \underline{0.3793} & 0.2870 \\
   & \textit{K}=10 & \textbf{0.5112} & \textbf{0.5244} & 0.4036 & 0.3031 & 0.4522 & \underline{0.4420} & \underline{0.5031} & 0.4030 \\
   & \textit{K}=20 & \underline{0.6318} & \textbf{0.6411} & 0.5271 & 0.4039 & 0.5610 & \underline{0.5475} & \textbf{0.6366} & 0.5460 \\
   & \textit{K}=25 & \underline{0.6717} & \textbf{0.6720} & 0.5712 & 0.4462 & 0.5985 & 0.5832 & \textbf{0.6765} & \underline{0.5877} \\
   & \textit{K}=50 & \underline{0.7778} & \textbf{0.7754} & 0.7043 & 0.5718 & 0.7106 & 0.7019 & \textbf{0.7949} & \underline{0.7172} \\
\bottomrule
\end{tabular}}}
\caption{\textbf{ROI cross-modal retrieval} performance across three benchmarks. \textbf{Bold} indicates the best-performing model, and \underline{underline} indicates the second-best-performing model for each dataset, K value and retrieval direction.}
\label{tab:cross_modal_retrieval}
\end{table*}


\begin{table*}[htbp]
\centering
\resizebox{1\columnwidth}{!}{
\renewcommand{\arraystretch}{1.2}{
\begin{tabular}{lcccc@{\hspace{8pt}}cccc@{\hspace{8pt}}cccc}
\toprule
    Source & \multicolumn{4}{c}{Test} & \multicolumn{4}{c}{Test-tiny} & \multicolumn{4}{c}{Val} \\ \cmidrule[0.7pt](lr){2-5} \cmidrule[0.7pt](lr){6-9} \cmidrule[0.7pt](lr){10-13}
   & MUSK & KEEP & CONCH & \textbf{ALICE} & MUSK & KEEP & CONCH & \textbf{ALICE} & MUSK & KEEP & CONCH & \textbf{ALICE} \\
\midrule
  Atlas & \underline{0.5919} & 0.5221 & 0.5732 & \textbf{0.5987} & \underline{0.6720} & 0.5820 & \textbf{0.6720} & 0.6614 & 0.5584 & 0.5584 & \underline{0.6104} & \textbf{0.6364} \\
  EduContent & \textbf{0.5387} & 0.4471 & 0.4287 & \underline{0.4533} & \textbf{0.5900} & 0.5272 & \underline{0.5439} & 0.5356 & \textbf{0.5455} & \underline{0.4196} & 0.3706 & 0.3916 \\
  CRC100K & 0.6961 & \textbf{0.7778} & 0.6209 & \underline{0.7582} & 0.6471 & 0.7353 & \underline{0.7647} & \textbf{0.8235} & \underline{0.7222} & 0.7222 & 0.6667 & \textbf{0.8333} \\
  LC25000 & 0.9588 & 0.8941 & \underline{0.9588} & \textbf{0.9941} & 1.0000 & 1.0000 & \underline{1.0000} & \textbf{1.0000} & 0.9000 & \textbf{1.0000} & 0.9000 & \underline{0.9000} \\
  Osteo & \underline{0.6373} & 0.5588 & 0.5980 & \textbf{0.6863} & 0.3333 & \underline{0.5833} & 0.5000 & \textbf{0.8333} & 0.6667 & 0.6667 & \underline{0.8333} & \textbf{0.8333} \\
  SICAPv2 & \textbf{0.5956} & 0.3235 & 0.5588 & \underline{0.5735} & \textbf{0.7500} & 0.2500 & 0.5625 & \underline{0.5625} & \textbf{0.7500} & 0.2500 & 0.6250 & \underline{0.6250} \\
\bottomrule
\end{tabular}}}
\caption{\textbf{PathMMU VQA} accuracy across ten data sources. \textbf{Bold} indicates the best-performing model, and \underline{underline} indicates the second-best-performing model for each source and cohort.}
\label{tab:pathmmu_vqa}
\end{table*}


\begin{table*}[htbp]
\centering
\resizebox{0.7\columnwidth}{!}{
\renewcommand{\arraystretch}{1.2}{
\begin{tabular}{llccccc}
\toprule
Task & Top\textit{K} pooling & \multicolumn{1}{c}{MUSK} & \multicolumn{1}{c}{KEEP} & \multicolumn{1}{c}{CONCH} & \multicolumn{1}{c}{ALICE} \\
\midrule
             \multirow{4}{*}{\makecell[l]{Lymph node metastasis screening\\(CAMELYON+, 2 classes)}} &       \textit{K}=1 & \underline{0.6549} & 0.6471 & 0.5444 & \textbf{0.8180} \\
                                   &       \textit{K}=5 & 0.6415 & \underline{0.6627} & 0.5278 & \textbf{0.7827} \\
                                   &      \textit{K}=10 & 0.6304 & \underline{0.6497} & 0.5222 & \textbf{0.7634} \\
                                   &      \textit{K}=50 & 0.5804 & \underline{0.6438} & 0.5222 & \textbf{0.6971} \\ \midrule
               \multirow{4}{*}{\makecell[l]{CNS tumor-family classification\\(EBRAINS, 9 classes)}} &       \textit{K}=1 & \underline{0.4415} & 0.4388 & 0.3272 & \textbf{0.5471} \\
                                   &       \textit{K}=5 & 0.4464 & \underline{0.5074} & 0.3476 & \textbf{0.5744} \\
                                   &      \textit{K}=10 & 0.4549 & \underline{0.5531} & 0.3597 & \textbf{0.5809} \\
\bottomrule
\end{tabular}}}
\caption{\textbf{Zero-shot WSI-level classification} performance across six benchmarks using \textbf{MI-Zero} TopK pooling. \textbf{Bold} indicates the best-performing model, and \underline{underline} indicates the second-best-performing model for each task and Top\textit{K} setting.}
\label{tab:zero_shot_comparison}
\end{table*}


\begin{table*}[htbp]
\centering
\resizebox{0.7\columnwidth}{!}{
\renewcommand{\arraystretch}{1.2}{
\begin{tabular}{lcccc}
\toprule
    Shot & MUSK & KEEP & CONCH & \textbf{ALICE} \\
\midrule
  1-shot & 0.4341$\pm$0.0181 & \underline{0.4940$\pm$0.0239} & 0.4577$\pm$0.0209 & \textbf{0.5237$\pm$0.0221} \\
  5-shot & 0.5266$\pm$0.0128 & \underline{0.6024$\pm$0.0163} & 0.5741$\pm$0.0109 & \textbf{0.6079$\pm$0.0135} \\
  10-shot & 0.5658$\pm$0.0193 & \underline{0.6296$\pm$0.0089} & 0.6135$\pm$0.0158 & \textbf{0.6525$\pm$0.0147} \\
  15-shot & 0.5917$\pm$0.0171 & \underline{0.6693$\pm$0.0183} & 0.6512$\pm$0.0122 & \textbf{0.6940$\pm$0.0149} \\
  20-shot & 0.6153$\pm$0.0201 & \underline{0.6769$\pm$0.0145} & 0.6507$\pm$0.0099 & \textbf{0.7081$\pm$0.0200} \\
\bottomrule
\end{tabular}}}
\caption{\textbf{EBRAINS (30 classes) WSI-level few-shot classification} performance using \textbf{PathPT}. Results are reported as the mean ± standard deviation across five random seeds. \textbf{Bold} indicates the best-performing model, and \underline{underline} indicates the second-best-performing model for each shot.}
\label{tab:wsi_few_shot_pathpt_ebrains}
\end{table*}

\begin{table*}[htbp]
\centering
\resizebox{0.7\columnwidth}{!}{
\renewcommand{\arraystretch}{1.2}{
\begin{tabular}{lcccc}
\toprule
    Shot & MUSK & KEEP & CONCH & \textbf{ALICE} \\
\midrule
  1-shot & 0.5661$\pm$0.0776 & 0.5077$\pm$0.0729 & \underline{0.5841$\pm$0.0637} & \textbf{0.6712$\pm$0.0210} \\
  5-shot & 0.7166$\pm$0.0320 & \underline{0.7966$\pm$0.0293} & 0.7231$\pm$0.0174 & \textbf{0.8012$\pm$0.0290} \\
  10-shot & 0.7596$\pm$0.0318 & \underline{0.8258$\pm$0.0370} & 0.7877$\pm$0.0480 & \textbf{0.8699$\pm$0.0198} \\
  15-shot & 0.7862$\pm$0.0305 & \underline{0.8671$\pm$0.0099} & 0.8177$\pm$0.0148 & \textbf{0.8832$\pm$0.0163} \\
  20-shot & 0.8118$\pm$0.0137 & \underline{0.8690$\pm$0.0304} & 0.8277$\pm$0.0317 & \textbf{0.8905$\pm$0.0059} \\
\bottomrule
\end{tabular}}}
\caption{\textbf{KidRare (4 classes) WSI-level few-shot classification} performance using \textbf{PathPT}. Results are reported as the mean ± standard deviation across five random seeds. \textbf{Bold} indicates the best-performing model, and \underline{underline} indicates the second-best-performing model for each shot.}
\label{tab:wsi_few_shot_pathpt_kidrare}
\end{table*}

\begin{table*}[htbp]
\centering
\resizebox{0.7\columnwidth}{!}{
\renewcommand{\arraystretch}{1.2}{
\begin{tabular}{lcccc}
\toprule
    Shot & MUSK & KEEP & CONCH & \textbf{ALICE} \\
\midrule
  1-shot & 0.2661$\pm$0.0197 & \underline{0.3283$\pm$0.0391} & 0.2954$\pm$0.0209 & \textbf{0.3347$\pm$0.0258} \\
  5-shot & 0.3131$\pm$0.0351 & 0.3567$\pm$0.0208 & \underline{0.3593$\pm$0.0314} & \textbf{0.3731$\pm$0.0232} \\
  10-shot & 0.3855$\pm$0.0123 & 0.3980$\pm$0.0163 & \underline{0.3987$\pm$0.0216} & \textbf{0.4057$\pm$0.0160} \\
  15-shot & 0.3904$\pm$0.0304 & \underline{0.4185$\pm$0.0194} & 0.4175$\pm$0.0388 & \textbf{0.4229$\pm$0.0306} \\
  20-shot & 0.3984$\pm$0.0290 & \underline{0.4265$\pm$0.0205} & 0.4053$\pm$0.0365 & \textbf{0.4292$\pm$0.0380} \\
\bottomrule
\end{tabular}}}
\caption{\textbf{SYSFL+ (7 classes) WSI-level few-shot classification} performance using \textbf{PathPT}. Results are reported as the mean ± standard deviation across five random seeds. \textbf{Bold} indicates the best-performing model, and \underline{underline} indicates the second-best-performing model for each shot.}
\label{tab:wsi_few_shot_pathpt_sys_fl_plus}
\end{table*}

\begin{table*}[htbp]
\centering
\resizebox{0.7\columnwidth}{!}{
\renewcommand{\arraystretch}{1.2}{
\begin{tabular}{lcccc}
\toprule
    Shot & MUSK & KEEP & CONCH & \textbf{ALICE} \\
\midrule
  1-shot & \underline{0.5568$\pm$0.0536} & 0.4902$\pm$0.0599 & 0.5143$\pm$0.0900 & \textbf{0.5914$\pm$0.0720} \\
  5-shot & 0.7089$\pm$0.0403 & \underline{0.7974$\pm$0.0216} & 0.7263$\pm$0.0705 & \textbf{0.8156$\pm$0.0342} \\
  10-shot & 0.7383$\pm$0.0402 & 0.7928$\pm$0.0384 & \underline{0.8054$\pm$0.0290} & \textbf{0.8339$\pm$0.0377} \\
  15-shot & 0.7781$\pm$0.0506 & 0.8199$\pm$0.0272 & \textbf{0.8285$\pm$0.0200} & \underline{0.8273$\pm$0.0177} \\
  20-shot & 0.8413$\pm$0.0230 & \underline{0.8425$\pm$0.0373} & 0.8363$\pm$0.0189 & \textbf{0.8524$\pm$0.0176} \\
\bottomrule
\end{tabular}}}
\caption{\textbf{UBC-OCEAN (5 classes) WSI-level few-shot classification} performance using \textbf{PathPT}. Results are reported as the mean ± standard deviation across five random seeds. \textbf{Bold} indicates the best-performing model, and \underline{underline} indicates the second-best-performing model for each shot.}
\label{tab:wsi_few_shot_pathpt_ubc_ocean}
\end{table*}

\begin{table*}[htbp]
\centering
\resizebox{0.7\columnwidth}{!}{
\renewcommand{\arraystretch}{1.2}{
\begin{tabular}{lcccc}
\toprule
    Shot & MUSK & KEEP & CONCH & \textbf{ALICE} \\
\midrule
  1-shot & \underline{0.5133$\pm$0.0179} & \textbf{0.5941$\pm$0.0492} & 0.5051$\pm$0.0212 & 0.5089$\pm$0.0182 \\
  5-shot & 0.5995$\pm$0.0864 & \underline{0.6476$\pm$0.0430} & 0.6201$\pm$0.1020 & \textbf{0.6667$\pm$0.0993} \\
  10-shot & 0.7376$\pm$0.1038 & 0.6948$\pm$0.0394 & \underline{0.7765$\pm$0.0935} & \textbf{0.8274$\pm$0.0484} \\
  15-shot & 0.8465$\pm$0.0637 & 0.7941$\pm$0.0695 & \underline{0.8620$\pm$0.0518} & \textbf{0.8770$\pm$0.0531} \\
  20-shot & 0.8171$\pm$0.0681 & 0.8433$\pm$0.0492 & \underline{0.8875$\pm$0.0428} & \textbf{0.9101$\pm$0.0300} \\
\bottomrule
\end{tabular}}}
\caption{\textbf{Camelyon+ (2 classes) WSI-level few-shot classification} performance using \textbf{PathPT}. Results are reported as the mean ± standard deviation across five random seeds. \textbf{Bold} indicates the best-performing model, and \underline{underline} indicates the second-best-performing model for each shot.}
\label{tab:wsi_few_shot_pathpt_camelyon_plus}
\end{table*}

\begin{table*}[htbp]
\centering
\resizebox{0.7\columnwidth}{!}{
\renewcommand{\arraystretch}{1.2}{
\begin{tabular}{lcccc}
\toprule
    Shot & MUSK & KEEP & CONCH & \textbf{ALICE} \\
\midrule
  1-shot & 0.8348$\pm$0.0286 & 0.7679$\pm$0.1063 & \textbf{0.9161$\pm$0.0201} & \underline{0.9056$\pm$0.0139} \\
  5-shot & 0.8727$\pm$0.0354 & 0.8994$\pm$0.0357 & \underline{0.9090$\pm$0.0145} & \textbf{0.9181$\pm$0.0063} \\
  10-shot & 0.9076$\pm$0.0203 & \textbf{0.9435$\pm$0.0088} & 0.9136$\pm$0.0284 & \underline{0.9395$\pm$0.0131} \\
  15-shot & 0.9139$\pm$0.0212 & \underline{0.9455$\pm$0.0242} & 0.9395$\pm$0.0103 & \textbf{0.9476$\pm$0.0079} \\
  20-shot & 0.9213$\pm$0.0316 & \underline{0.9557$\pm$0.0200} & 0.9541$\pm$0.0202 & \textbf{0.9557$\pm$0.0119} \\
\bottomrule
\end{tabular}}}
\caption{\textbf{CPTAC-NSCLC (2 classes) WSI-level few-shot classification} performance using PathPT. Results are reported as the mean ± standard deviation across five random seeds. \textbf{Bold} indicates the best-performing model, and \underline{underline} indicates the second-best-performing model for each shot.}
\label{tab:wsi_few_shot_pathpt_cptac_nsclc}
\end{table*}


\begin{table*}[htbp]
\centering
\resizebox{\textwidth}{!}{
\renewcommand{\arraystretch}{1.2}{
\begin{tabular}{llcccc}
\toprule
    Task & Modality & MUSK & KEEP & CONCH & \textbf{ALICE} \\
\midrule
  \multirow{3}{*}{\makecell[l]{Breast cancer axillary lymph node metastasis\\(BCNB, 2 classes)}} & WSI-only & \textbf{0.7040$\pm$0.0134} & 0.6354$\pm$0.0191 & 0.6610$\pm$0.0329 & \underline{0.6658$\pm$0.0303} \\
   & Text-only & 0.5490$\pm$0.0286 & \textbf{0.6071$\pm$0.0409} & \underline{0.5839$\pm$0.0475} & 0.5667$\pm$0.0353 \\
   & Multimodal & 0.6003$\pm$0.0856 & 0.6624$\pm$0.0116 & \underline{0.6636$\pm$0.0330} & \textbf{0.6693$\pm$0.0293} \\
\midrule
  \multirow{3}{*}{\makecell[l]{Breast cancer axillary lymph node metastasis\\(BCNB, 3 classes)}} & WSI-only & \textbf{0.5010$\pm$0.0318} & 0.4401$\pm$0.0245 & 0.4297$\pm$0.0265 & \underline{0.4534$\pm$0.0186} \\
   & Text-only & 0.3613$\pm$0.0302 & \underline{0.3749$\pm$0.0342} & \textbf{0.4283$\pm$0.0315} & 0.3675$\pm$0.0290 \\
   & Multimodal & 0.3934$\pm$0.0458 & 0.3920$\pm$0.0209 & \underline{0.4022$\pm$0.0289} & \textbf{0.4530$\pm$0.0132} \\
\midrule
  \multirow{3}{*}{\makecell[l]{Head and neck keratinizing SCC grading\\(Hancock, 2 classes)}} & WSI-only & 0.6347$\pm$0.0163 & \textbf{0.6607$\pm$0.0404} & \underline{0.6583$\pm$0.0368} & 0.6385$\pm$0.0142 \\
   & Text-only & 0.5200$\pm$0.0240 & 0.5303$\pm$0.0258 & \underline{0.5509$\pm$0.0111} & \textbf{0.5567$\pm$0.0273} \\
   & Multimodal & 0.6105$\pm$0.0680 & 0.6446$\pm$0.0129 & \underline{0.6483$\pm$0.0211} & \textbf{0.6978$\pm$0.0326} \\
\midrule
  \multirow{3}{*}{\makecell[l]{Head and neck lymphovascular invasion detection\\(Hancock, 2 classes)}} & WSI-only & 0.6185$\pm$0.0358 & \textbf{0.6760$\pm$0.0381} & \underline{0.6586$\pm$0.0201} & 0.6410$\pm$0.0196 \\
   & Text-only & \underline{0.6705$\pm$0.0045} & 0.6522$\pm$0.0310 & 0.6681$\pm$0.0069 & \textbf{0.6725$\pm$0.0171} \\
   & Multimodal & 0.5970$\pm$0.0512 & \underline{0.6441$\pm$0.0145} & 0.6360$\pm$0.0482 & \textbf{0.7261$\pm$0.0332} \\
\midrule
  \multirow{3}{*}{\makecell[l]{Head and neck vascular invasion detection\\(Hancock, 2 classes)}} & WSI-only & 0.6131$\pm$0.0288 & 0.5819$\pm$0.1122 & \underline{0.6512$\pm$0.0905} & \textbf{0.6577$\pm$0.0923} \\
   & Text-only & 0.5661$\pm$0.0474 & 0.5466$\pm$0.0266 & \underline{0.6327$\pm$0.0534} & \textbf{0.6405$\pm$0.0153} \\
   & Multimodal & 0.5381$\pm$0.0700 & \underline{0.7171$\pm$0.1264} & 0.7141$\pm$0.0586 & \textbf{0.9032$\pm$0.0243} \\
\midrule
  \multirow{3}{*}{\makecell[l]{Ovarian cancer treatment response prediction\\(PTRC-HGSOC, 2 classes)}} & WSI-only & 0.5108$\pm$0.0453 & \underline{0.5775$\pm$0.0240} & 0.5308$\pm$0.0295 & \textbf{0.6058$\pm$0.0278} \\
   & Text-only & 0.5750$\pm$0.0736 & 0.6566$\pm$0.0037 & \textbf{0.6767$\pm$0.0354} & \underline{0.6583$\pm$0.0000} \\
   & Multimodal & 0.4800$\pm$0.0471 & \underline{0.5817$\pm$0.0309} & 0.5625$\pm$0.0466 & \textbf{0.6617$\pm$0.0173} \\
\midrule
  \multirow{3}{*}{\makecell[l]{Ovarian cancer tumor type classification\\(PTRC-HGSOC, 2 classes)}} & WSI-only & 0.8056$\pm$0.0099 & \textbf{0.8660$\pm$0.0219} & 0.8163$\pm$0.0225 & \underline{0.8320$\pm$0.0119} \\
   & Text-only & 0.5000$\pm$0.0000 & 0.5111$\pm$0.0249 & \textbf{0.5167$\pm$0.0373} & \underline{0.5167$\pm$0.0373} \\
   & Multimodal & 0.8254$\pm$0.0334 & 0.8212$\pm$0.0379 & \underline{0.8358$\pm$0.0388} & \textbf{0.8674$\pm$0.0294} \\
\midrule
  \multirow{3}{*}{\makecell[l]{Melanoma relapse prediction without prior melanoma\\(Visiomel, 2 classes)}} & WSI-only & \underline{0.6856$\pm$0.0183} & \textbf{0.6981$\pm$0.0354} & 0.6396$\pm$0.0131 & 0.6743$\pm$0.0226 \\
   & Text-only & 0.7327$\pm$0.0172 & \textbf{0.7426$\pm$0.0000} & \underline{0.7383$\pm$0.0199} & 0.7150$\pm$0.0179 \\
   & Multimodal & \underline{0.7139$\pm$0.0234} & 0.6908$\pm$0.0422 & 0.6684$\pm$0.0418 & \textbf{0.7215$\pm$0.0318} \\
\midrule
  \multirow{3}{*}{\makecell[l]{Melanoma overall relapse prediction\\(Visiomel, 2 classes)}} & WSI-only & 0.7005$\pm$0.0262 & \underline{0.7208$\pm$0.0187} & 0.7174$\pm$0.0092 & \textbf{0.7322$\pm$0.0175} \\
   & Text-only & 0.7717$\pm$0.0100 & \textbf{0.7792$\pm$0.0179} & 0.7641$\pm$0.0254 & \underline{0.7741$\pm$0.0076} \\
   & Multimodal & 0.7423$\pm$0.0162 & \underline{0.7468$\pm$0.0240} & 0.7465$\pm$0.0217 & \textbf{0.7539$\pm$0.0174} \\
\bottomrule
\end{tabular}}}
\caption{\textbf{Modality comparison on MICA multimodal WSI-text tasks.} Results are reported as the mean ± standard deviation across five random seeds. \textbf{Bold} indicates the best-performing model, and \underline{underline} indicates the second-best-performing model for each task and modality.}
\label{tab:wsi_multimodal_modality_comparison_appendix}
\end{table*}

\begin{table*}[htbp]
\centering
\resizebox{0.9\textwidth}{!}{
\renewcommand{\arraystretch}{1.2}{
\begin{tabular}{lccc}
\toprule
    Task & TITAN & CARE & \textbf{ALICE} \\
\midrule
  \makecell[l]{Breast carcinoma coarse-grained subtyping\\(BRACS, 3 classes)} & \textbf{0.6798$\pm$0.0343} & \underline{0.6164$\pm$0.0237} & 0.6082$\pm$0.0141 \\
  \makecell[l]{Breast carcinoma fine-grained subtyping\\(BRACS, 7 classes)} & 0.3206$\pm$0.0228 & \underline{0.3297$\pm$0.0274} & \textbf{0.3455$\pm$0.0259} \\
  \makecell[l]{Lymph node metastasis screening\\(CAMELYON+, 2 classes)} & 0.7234$\pm$0.0229 & \underline{0.7339$\pm$0.0172} & \textbf{0.7595$\pm$0.0020} \\
  \makecell[l]{Lymph node metastatic burden classification\\(CAMELYON+, 4 classes)} & 0.4114$\pm$0.0146 & \underline{0.4382$\pm$0.0313} & \textbf{0.4745$\pm$0.0179} \\
  \makecell[l]{Non-small cell lung cancer subtyping\\(CPTAC-NSCLC, 2 classes)} & \underline{0.9605$\pm$0.0080} & 0.9330$\pm$0.0067 & \textbf{0.9774$\pm$0.0040} \\
  \makecell[l]{Renal cell carcinoma subtyping\\(DHMC-RCC, 5 classes)} & \underline{0.7449$\pm$0.0070} & \textbf{0.7790$\pm$0.0023} & 0.7288$\pm$0.0150 \\
  \makecell[l]{CNS fine-grained tumor diagnosis\\(EBRAINS, 30 classes)} & \textbf{0.6525$\pm$0.0133} & 0.5943$\pm$0.0185 & \underline{0.6489$\pm$0.0069} \\
  \makecell[l]{CNS tumor-family classification\\(EBRAINS, 9 classes)} & \textbf{0.8677$\pm$0.0152} & 0.8173$\pm$0.0181 & \underline{0.8609$\pm$0.0176} \\
  \makecell[l]{CNS tumor WHO grading\\(EBRAINS, 4 classes)} & \textbf{0.7298$\pm$0.0078} & 0.6940$\pm$0.0041 & \underline{0.7155$\pm$0.0092} \\
  \makecell[l]{Colorectal neoplasia screening\\(HunCRC, 4 classes)} & \underline{0.6320$\pm$0.0198} & \textbf{0.6375$\pm$0.0138} & 0.4830$\pm$0.0151 \\
  \makecell[l]{Pediatric rare tumor classification\\(KidRare, 4 classes)} & \textbf{0.9032$\pm$0.0043} & 0.8629$\pm$0.0095 & \underline{0.9030$\pm$0.0123} \\
  \makecell[l]{Pediatric rare tumor fine-grained subtyping\\(KidRare, 13 classes)} & \textbf{0.5876$\pm$0.0354} & \underline{0.5242$\pm$0.0552} & 0.4657$\pm$0.0467 \\
  \makecell[l]{Prostate cancer Gleason grading\\(PANDA, 6 classes)} & 0.5717$\pm$0.0214 & \underline{0.6185$\pm$0.0316} & \textbf{0.6318$\pm$0.0263} \\
  \makecell[l]{Prostate cancer screening\\(PANDA, 2 classes)} & 0.8786$\pm$0.0093 & \underline{0.8972$\pm$0.0077} & \textbf{0.9340$\pm$0.0060} \\
  \makecell[l]{Bone metastasis primary site prediction\\(SYSBM, 10 classes)} & \textbf{0.6725$\pm$0.0416} & 0.5593$\pm$0.0631 & \underline{0.6332$\pm$0.0511} \\
  \makecell[l]{Lung lesion triage\\(SYSFL+, 4 classes)} & \underline{0.6080$\pm$0.0183} & \textbf{0.6176$\pm$0.0144} & 0.6080$\pm$0.0027 \\
  \makecell[l]{Lung cancer subtyping and invasion grading\\(SYSFL+, 7 classes)} & \underline{0.4211$\pm$0.0123} & \textbf{0.4238$\pm$0.0094} & 0.4200$\pm$0.0113 \\
  \makecell[l]{Cervical epithelial lesion classification\\(TissueNet, 4 classes)} & 0.6068$\pm$0.0293 & \underline{0.6230$\pm$0.0222} & \textbf{0.6772$\pm$0.0113} \\
  \makecell[l]{Ovarian carcinoma subtyping\\(UBC-OCEAN, 5 classes)} & \underline{0.8536$\pm$0.0232} & 0.8344$\pm$0.0297 & \textbf{0.8612$\pm$0.0171} \\
\bottomrule
\end{tabular}}}
\caption{\textbf{Slide-level clinical diagnosis KNN evaluation.} Balanced accuracy is reported as the mean ± standard deviation across five random seeds. \textbf{Bold} indicates the best-performing model, and \underline{underline} indicates the second-best-performing model for each task and modality.}
\label{tab:knn_balacc_clinical_diagnosis_classcount_4dec}
\end{table*}

\begin{table*}[htbp]
\centering
\resizebox{0.9\textwidth}{!}{
\renewcommand{\arraystretch}{1.2}{
\begin{tabular}{lccc}
\toprule
    Task & TITAN & CARE & \textbf{ALICE} \\
\midrule
  \makecell[l]{Breast carcinoma coarse-grained subtyping\\(BRACS, 3 classes)} & \underline{0.3800$\pm$0.0554} & 0.3278$\pm$0.0442 & \textbf{0.4458$\pm$0.0462} \\
  \makecell[l]{Breast carcinoma fine-grained subtyping\\(BRACS, 7 classes)} & 0.2064$\pm$0.0464 & \textbf{0.2568$\pm$0.0303} & \underline{0.2333$\pm$0.0533} \\
  \makecell[l]{Lymph node metastasis screening\\(CAMELYON+, 2 classes)} & 0.5354$\pm$0.0510 & \underline{0.5674$\pm$0.0652} & \textbf{0.7399$\pm$0.0904} \\
  \makecell[l]{Lymph node metastatic burden classification\\(CAMELYON+, 4 classes)} & 0.2602$\pm$0.0172 & \underline{0.2753$\pm$0.0426} & \textbf{0.3065$\pm$0.0621} \\
  \makecell[l]{Non-small cell lung cancer subtyping\\(CPTAC-NSCLC, 2 classes)} & \underline{0.9528$\pm$0.0205} & 0.8991$\pm$0.0340 & \textbf{0.9915$\pm$0.0060} \\
  \makecell[l]{Renal cell carcinoma subtyping\\(DHMC-RCC, 5 classes)} & \underline{0.2657$\pm$0.0549} & \textbf{0.3558$\pm$0.1012} & 0.2290$\pm$0.0584 \\
  \makecell[l]{CNS fine-grained tumor diagnosis\\(EBRAINS, 30 classes)} & \underline{0.1276$\pm$0.0332} & 0.0704$\pm$0.0238 & \textbf{0.2901$\pm$0.2013} \\
  \makecell[l]{CNS tumor-family classification\\(EBRAINS, 9 classes)} & \underline{0.3727$\pm$0.2099} & 0.1560$\pm$0.0410 & \textbf{0.4451$\pm$0.2614} \\
  \makecell[l]{CNS tumor WHO grading\\(EBRAINS, 4 classes)} & \underline{0.4728$\pm$0.0034} & 0.3848$\pm$0.0736 & \textbf{0.6019$\pm$0.0677} \\
  \makecell[l]{Colorectal neoplasia screening\\(HunCRC, 4 classes)} & \textbf{0.3388$\pm$0.0907} & \underline{0.2542$\pm$0.0505} & 0.2475$\pm$0.0481 \\
  \makecell[l]{Pediatric rare tumor classification\\(KidRare, 4 classes)} & \underline{0.6210$\pm$0.0678} & 0.4318$\pm$0.0155 & \textbf{0.6268$\pm$0.2081} \\
  \makecell[l]{Pediatric rare tumor fine-grained subtyping\\(KidRare, 13 classes)} & \textbf{0.2178$\pm$0.0025} & 0.1305$\pm$0.0305 & \underline{0.2109$\pm$0.0056} \\
  \makecell[l]{Prostate cancer Gleason grading\\(PANDA, 6 classes)} & \underline{0.4655$\pm$0.0076} & 0.3535$\pm$0.1181 & \textbf{0.5593$\pm$0.0156} \\
  \makecell[l]{Prostate cancer screening\\(PANDA, 2 classes)} & 0.5909$\pm$0.0980 & \underline{0.5932$\pm$0.1210} & \textbf{0.8998$\pm$0.0187} \\
  \makecell[l]{Bone metastasis primary site prediction\\(SYSBM, 10 classes)} & \underline{0.2322$\pm$0.0582} & 0.1500$\pm$0.0457 & \textbf{0.2414$\pm$0.0491} \\
  \makecell[l]{Lung lesion triage\\(SYSFL+, 4 classes)} & \textbf{0.3133$\pm$0.0644} & 0.2707$\pm$0.0300 & \underline{0.2830$\pm$0.0280} \\
  \makecell[l]{Lung cancer subtyping and invasion grading\\(SYSFL+, 7 classes)} & \textbf{0.2925$\pm$0.0885} & 0.1707$\pm$0.0491 & \underline{0.2430$\pm$0.1048} \\
  \makecell[l]{Cervical epithelial lesion classification\\(TissueNet, 4 classes)} & \underline{0.4696$\pm$0.0126} & 0.3983$\pm$0.0799 & \textbf{0.6167$\pm$0.0637} \\
  \makecell[l]{Ovarian carcinoma subtyping\\(UBC-OCEAN, 5 classes)} & \textbf{0.3396$\pm$0.0949} & 0.2432$\pm$0.0558 & \underline{0.3225$\pm$0.0863} \\
\bottomrule
\end{tabular}}}
\caption{\textbf{Slide-level clinical diagnosis linear probe evaluation.} Balanced accuracy is reported as the mean $\pm$ sample standard deviation across five random seeds. \textbf{Bold} indicates the best-performing model, and \underline{underline} indicates the second-best-performing model for each row.}
\label{tab:lp_ft_aligned_s5_pat30_balacc_clinical_diagnosis_classcount_4dec}
\end{table*}


\begin{table*}[htbp]
\centering
\resizebox{0.6\columnwidth}{!}{
\renewcommand{\arraystretch}{1.2}{
\begin{tabular}{lccc}
\toprule
  Dataset & TITAN & CARE & \textbf{ALICE} \\
\midrule
  \makecell[l]{Breast carcinoma coarse-grained subtyping\\(BRACS, 3 classes)} & \underline{0.6782} & 0.6782 & \textbf{0.7126} \\
  \makecell[l]{Breast carcinoma fine-grained subtyping\\(BRACS, 7 classes)} & 0.3793 & \underline{0.4138} & \textbf{0.4368} \\
  \makecell[l]{Lymph node metastasis screening\\(CAMELYON+, 2 classes)} & \underline{0.8000} & 0.7885 & \textbf{0.8192} \\
  \makecell[l]{Lymph node metastatic burden classification\\(CAMELYON+, 4 classes)} & \underline{0.7348} & 0.7348 & \textbf{0.7879} \\
  \makecell[l]{Pan-cancer primary site prediction\\(CMB, 9 classes)} & \textbf{0.8112} & 0.7551 & \underline{0.7959} \\
  \makecell[l]{Non-small cell lung cancer subtyping\\(CPTAC-NSCLC, 2 classes)} & \underline{0.9417} & 0.9250 & \textbf{0.9750} \\
  \makecell[l]{Pan-cancer primary site prediction\\(CPTAC, 9 classes)} & \textbf{0.9581} & \underline{0.9532} & 0.9458 \\
  \makecell[l]{Renal cell carcinoma subtyping\\(DHMC-RCC, 5 classes)} & \underline{0.8344} & \textbf{0.8535} & 0.8025 \\
  \makecell[l]{CNS fine-grained tumor diagnosis\\(EBRAINS, 30 classes)} & \underline{0.6883} & 0.6497 & \textbf{0.7075} \\
  \makecell[l]{CNS tumor-family classification\\(EBRAINS, 9 classes)} & \underline{0.9194} & 0.9037 & \textbf{0.9317} \\
  \makecell[l]{CNS tumor WHO grading\\(EBRAINS, 4 classes)} & \underline{0.7911} & 0.7748 & \textbf{0.7992} \\
  \makecell[l]{Colorectal neoplasia screening\\(HunCRC, 4 classes)} & \underline{0.8095} & \textbf{0.8333} & 0.6429 \\
  \makecell[l]{Pediatric rare tumor classification\\(KidRare, 4 classes)} & \textbf{0.9141} & 0.8750 & \underline{0.8984} \\
  \makecell[l]{Pediatric rare tumor fine-grained subtyping\\(KidRare, 13 classes)} & \textbf{0.7148} & \underline{0.6641} & 0.6562 \\
  \makecell[l]{Prostate cancer Gleason grading\\(PANDA, 6 classes)} & 0.6024 & \underline{0.6413} & \textbf{0.6623} \\
  \makecell[l]{Prostate cancer screening\\(PANDA, 2 classes)} & 0.8859 & \underline{0.9010} & \textbf{0.9288} \\
  \makecell[l]{Bone metastasis primary site prediction\\(SYSBM, 10 classes)} & \textbf{0.7544} & \underline{0.7310} & 0.7310 \\
  \makecell[l]{Lung lesion triage\\(SYSFL+, 4 classes)} & 0.6510 & \underline{0.6577} & \textbf{0.6779} \\
  \makecell[l]{Lung cancer subtyping and invasion grading\\(SYSFL+, 7 classes)} & 0.4128 & \textbf{0.4329} & \underline{0.4329} \\
  \makecell[l]{Cervical epithelial lesion classification\\(TissueNet, 4 classes)} & \underline{0.5864} & 0.5288 & \textbf{0.5969} \\
  \makecell[l]{Ovarian carcinoma subtyping\\(UBC-OCEAN, 5 classes)} & \textbf{0.8727} & \underline{0.8636} & 0.8455 \\
\midrule
  \multicolumn{1}{c}{Overall} & \underline{0.7495} & 0.7409 & \textbf{0.7518} \\
\bottomrule
\end{tabular}}}
\caption{\textbf{Slide-level retrieval} performance across 21 benchmarks using \textbf{WSI features}. Performance is reported as MVAcc@3. \textbf{Bold} indicates the best-performing model, and \underline{underline} indicates the second-best-performing model for each dataset.}
\label{tab:wsi_retrieval_mvacc_at_3}
\end{table*}


\begin{table*}[htbp]
\centering
\resizebox{0.8\textwidth}{!}{
\renewcommand{\arraystretch}{1.2}{
\begin{tabular}{lccc}
\toprule
    Task & TITAN & CARE & \textbf{ALICE} \\
\midrule
  \makecell[l]{ER status prediction\\(BCNB, 2 classes)} & \underline{0.5728$\pm$0.1189} & \textbf{0.6608$\pm$0.0657} & 0.5019$\pm$0.0856 \\
  \makecell[l]{HER2 status prediction\\(BCNB, 2 classes)} & \underline{0.5118$\pm$0.0548} & 0.4682$\pm$0.0509 & \textbf{0.5667$\pm$0.0735} \\
  \makecell[l]{PR status prediction\\(BCNB, 2 classes)} & \underline{0.5726$\pm$0.0820} & \textbf{0.6095$\pm$0.0405} & 0.5329$\pm$0.0792 \\
  \makecell[l]{Glioma IDH status prediction\\(EBRAINS, 2 classes)} & \underline{0.6515$\pm$0.1593} & 0.5633$\pm$0.0782 & \textbf{0.9259$\pm$0.0658} \\
  \makecell[l]{BAP1 mutation prediction\\(MUT-HET-RCC, 2 classes)} & 0.4621$\pm$0.1252 & \underline{0.5171$\pm$0.1226} & \textbf{0.5925$\pm$0.1256} \\
  \makecell[l]{PBRM1 mutation prediction\\(MUT-HET-RCC, 2 classes)} & \underline{0.6392$\pm$0.0464} & 0.6070$\pm$0.0669 & \textbf{0.7001$\pm$0.0191} \\
  \makecell[l]{SETD2 mutation prediction\\(MUT-HET-RCC, 2 classes)} & 0.5182$\pm$0.0682 & \textbf{0.5519$\pm$0.0696} & \underline{0.5412$\pm$0.0588} \\
\bottomrule
\end{tabular}}}
\caption{\textbf{Slide-level biomarker prediction linear probe evaluation.} AUC is reported as the mean ± standard deviation across five random seeds. \textbf{Bold} indicates the best-performing model, and \underline{underline} indicates the second-best-performing model for each task.}
\label{tab:lp_ft_aligned_s5_pat30_auc_biomarker_prediction_classcount_4dec}
\end{table*}


\begin{table*}[htbp]
\centering
\resizebox{0.7\textwidth}{!}{
\renewcommand{\arraystretch}{1.2}{
\begin{tabular}{lccc}
\toprule
    Task & TITAN & CARE & \textbf{ALICE} \\
\midrule
  CPTAC-CCRCC (OS) & \underline{0.5953$\pm$0.1875} & 0.5835$\pm$0.1015 & \textbf{0.5989$\pm$0.2200} \\
  CPTAC-HNSC (OS) & 0.5549$\pm$0.1531 & \textbf{0.6498$\pm$0.1064} & \underline{0.6304$\pm$0.0776} \\
  CPTAC-LUAD (OS) & \textbf{0.5987$\pm$0.1215} & 0.5367$\pm$0.0884 & \underline{0.5707$\pm$0.0918} \\
  CPTAC-PDAC (OS) & \underline{0.5019$\pm$0.0543} & \textbf{0.5308$\pm$0.0728} & 0.4949$\pm$0.1000 \\
  Hancock (OS) & \underline{0.5022$\pm$0.0814} & 0.4558$\pm$0.0563 & \textbf{0.5586$\pm$0.0744} \\
  SURGEN (OS) & 0.5368$\pm$0.0731 & \underline{0.5435$\pm$0.0384} & \textbf{0.5537$\pm$0.0919} \\
\bottomrule
\end{tabular}}}
\caption{\textbf{Slide-level survival prediction evaluation.} C-index is reported as the mean ± standard deviation across five random seeds. \textbf{Bold} indicates the best-performing model, and \underline{underline} indicates the second-best-performing model for each task.}
\label{tab:wsi_surv_cindex}
\end{table*}


\begin{table*}[htbp]
\centering
\resizebox{0.9\columnwidth}{!}{
\renewcommand{\arraystretch}{1.2}{
\begin{tabular}{llccc}
\toprule
Task & Shot & \multicolumn{1}{c}{TITAN} & \multicolumn{1}{c}{CARE} & \multicolumn{1}{c}{ALICE} \\
\midrule
\multirow{6}{*}{\makecell[l]{Pan-cancer classification\\(CPTAC, 9 classes)}} 
& \textit{K}=1  & \textbf{0.7266$\pm$0.0754} & 0.5469$\pm$0.0303 & \underline{0.7255$\pm$0.0706} \\
& \textit{K}=2  & \textbf{0.8049$\pm$0.0468} & 0.5401$\pm$0.0622 & \underline{0.7993$\pm$0.0283} \\
& \textit{K}=4  & \textbf{0.8718$\pm$0.0205} & 0.7522$\pm$0.0245 & \underline{0.8696$\pm$0.0218} \\
& \textit{K}=8  & \underline{0.8818$\pm$0.0292} & 0.8011$\pm$0.0232 & \textbf{0.8960$\pm$0.0087} \\
& \textit{K}=16 & \underline{0.8993$\pm$0.0047} & 0.8461$\pm$0.0139 & \textbf{0.9153$\pm$0.0136} \\
& \textit{K}=32 & \underline{0.9037$\pm$0.0073} & 0.8521$\pm$0.0140 & \textbf{0.9288$\pm$0.0058} \\ \midrule

\multirow{6}{*}{\makecell[l]{Bone metastasis primary site prediction\\(SYSBM, 10 classes)}} 
& \textit{K}=1  & \textbf{0.3474$\pm$0.0884} & 0.2683$\pm$0.0786 & \underline{0.3083$\pm$0.1255} \\
& \textit{K}=2  & \textbf{0.4435$\pm$0.0935} & 0.4217$\pm$0.0319 & \underline{0.4337$\pm$0.0614} \\
& \textit{K}=4  & \textbf{0.5322$\pm$0.0440} & 0.4877$\pm$0.0342 & \underline{0.5310$\pm$0.0779} \\
& \textit{K}=8  & \textbf{0.6176$\pm$0.0495} & 0.5273$\pm$0.0213 & \underline{0.5766$\pm$0.0225} \\
& \textit{K}=16 & \underline{0.5992$\pm$0.0282} & 0.5163$\pm$0.0309 & \textbf{0.6087$\pm$0.0575} \\
& \textit{K}=32 & \textbf{0.4983$\pm$0.0186} & 0.4525$\pm$0.0191 & \underline{0.4852$\pm$0.0318} \\ \midrule

\multirow{6}{*}{\makecell[l]{Pan-cancer classification\\(CMB, 9 classes)}} 
& \textit{K}=1  & \underline{0.4647$\pm$0.0656} & 0.4220$\pm$0.0348 & \textbf{0.4718$\pm$0.0861} \\
& \textit{K}=2  & \underline{0.5040$\pm$0.0492} & 0.4011$\pm$0.0812 & \textbf{0.5139$\pm$0.0640} \\
& \textit{K}=4  & \underline{0.6207$\pm$0.0582} & 0.4726$\pm$0.0485 & \textbf{0.6676$\pm$0.0813} \\
& \textit{K}=8  & \underline{0.6388$\pm$0.0359} & 0.5384$\pm$0.0360 & \textbf{0.6757$\pm$0.0270} \\
& \textit{K}=16 & \underline{0.6786$\pm$0.0330} & 0.5468$\pm$0.0612 & \textbf{0.6995$\pm$0.0168} \\
& \textit{K}=32 & \underline{0.5653$\pm$0.0419} & 0.4964$\pm$0.0232 & \textbf{0.6613$\pm$0.0257} \\
\bottomrule
\end{tabular}}}
\caption{\textbf{Comparison of few-shot slide-level classification} performance using \textbf{WSI features} under linear probing. Results are reported as the mean ± standard deviation. \textbf{Bold} indicates the best-performing model, and \underline{underline} indicates the second-best-performing model for each task and shot.}
\label{tab:slide_few_shot_wsi_comparison}
\end{table*}


\begin{table*}[htbp]
\centering
\resizebox{1.0\textwidth}{!}{
\renewcommand{\arraystretch}{1.2}{
\begin{tabular}{llccc}
\toprule
    Task & Setting & TITAN & CARE & \textbf{ALICE} \\
\midrule
  \multirow{2}{*}{\makecell[l]{Breast carcinoma coarse-grained subtyping\\(BRACS, 3 classes)}} & Linear probe & \underline{0.3800$\pm$0.0554} & 0.3278$\pm$0.0442 & \textbf{0.4458$\pm$0.0462} \\
   & Fine-tuning & 0.6704$\pm$0.0362 & \underline{0.6913$\pm$0.0473} & \textbf{0.7258$\pm$0.0340} \\
\midrule
  \multirow{2}{*}{\makecell[l]{Breast carcinoma fine-grained subtyping\\(BRACS, 7 classes)}} & Linear probe & 0.2064$\pm$0.0464 & \textbf{0.2568$\pm$0.0303} & \underline{0.2333$\pm$0.0533} \\
   & Fine-tuning & 0.3727$\pm$0.0198 & \underline{0.4397$\pm$0.0340} & \textbf{0.4447$\pm$0.0200} \\
\midrule
  \multirow{2}{*}{\makecell[l]{Lymph node metastasis coarse-grained classification\\(CAMELYON+, 2 classes)}} & Linear probe & 0.5354$\pm$0.0510 & \underline{0.5675$\pm$0.0652} & \textbf{0.7399$\pm$0.0904} \\
   & Fine-tuning & 0.7508$\pm$0.0202 & \underline{0.9013$\pm$0.0125} & \textbf{0.9229$\pm$0.0085} \\
\midrule
  \multirow{2}{*}{\makecell[l]{Lymph node metastasis fine-grained classification\\(CAMELYON+, 4 classes)}} & Linear probe & 0.2602$\pm$0.0172 & \underline{0.2753$\pm$0.0426} & \textbf{0.3065$\pm$0.0621} \\
   & Fine-tuning & 0.4454$\pm$0.0112 & \underline{0.5738$\pm$0.0246} & \textbf{0.6155$\pm$0.0220} \\
\midrule
  \multirow{2}{*}{\makecell[l]{Pediatric rare tumor classification\\(KidRare, 4 classes)}} & Linear probe & \underline{0.6210$\pm$0.0678} & 0.4318$\pm$0.0155 & \textbf{0.6268$\pm$0.2081} \\
   & Fine-tuning & \underline{0.9013$\pm$0.0072} & 0.8977$\pm$0.0247 & \textbf{0.9298$\pm$0.0197} \\
\midrule
  \multirow{2}{*}{\makecell[l]{Pediatric rare tumor fine-grained subtyping\\(KidRare, 13 classes)}} & Linear probe & \textbf{0.2178$\pm$0.0025} & 0.1305$\pm$0.0305 & \underline{0.2109$\pm$0.0056} \\
   & Fine-tuning & \underline{0.5374$\pm$0.0436} & 0.4955$\pm$0.0630 & \textbf{0.5774$\pm$0.0544} \\
\bottomrule
\end{tabular}}}
\caption{\textbf{Slide-level clinical diagnosis comparison between linear probing and fine-tuning.} Balanced accuracy is reported as the mean ± standard deviation across five random seeds. \textbf{Bold} indicates the best-performing model, and \underline{underline} indicates the second-best-performing model for each setting.}
\label{tab:lp_vs_ft_diagnosis_results_setting_rows}
\end{table*}

\begin{table*}[htbp]
\centering
\resizebox{0.8\textwidth}{!}{
\renewcommand{\arraystretch}{1.2}{
\begin{tabular}{llccc}
\toprule
    Task & Setting & TITAN & CARE & \textbf{ALICE} \\
\midrule
  \multirow{2}{*}{\makecell[l]{ER status prediction\\(BCNB, 2 classes)}} & Linear probe & \underline{0.5728$\pm$0.1329} & \textbf{0.6608$\pm$0.0734} & 0.5019$\pm$0.0957 \\
   & Fine-tuning & \underline{0.8557$\pm$0.0172} & 0.8491$\pm$0.0200 & \textbf{0.8900$\pm$0.0260} \\
\midrule
  \multirow{2}{*}{\makecell[l]{PR status prediction\\(BCNB, 2 classes)}} & Linear probe & \underline{0.5726$\pm$0.0917} & \textbf{0.6095$\pm$0.0453} & 0.5329$\pm$0.0885 \\
   & Fine-tuning & \underline{0.8032$\pm$0.0076} & 0.7813$\pm$0.0219 & \textbf{0.8175$\pm$0.0085} \\
\midrule
  \multirow{2}{*}{\makecell[l]{PBRM1 mutation prediction\\(MUT-HET-RCC, 2 classes)}} & Linear probe & \underline{0.6392$\pm$0.0518} & 0.6070$\pm$0.0748 & \textbf{0.7001$\pm$0.0213} \\
   & Fine-tuning & \underline{0.7391$\pm$0.0202} & 0.7369$\pm$0.0220 & \textbf{0.8225$\pm$0.0081} \\
\midrule
  \multirow{2}{*}{\makecell[l]{BAP1 mutation prediction\\(MUT-HET-RCC, 2 classes)}} & Linear probe & 0.4621$\pm$0.1399 & \underline{0.5171$\pm$0.1370} & \textbf{0.5925$\pm$0.1404} \\
   & Fine-tuning & 0.8254$\pm$0.0297 & \underline{0.8413$\pm$0.0103} & \textbf{0.8997$\pm$0.0213} \\
\midrule
  \multirow{2}{*}{\makecell[l]{SETD2 mutation prediction\\(MUT-HET-RCC, 2 classes)}} & Linear probe & 0.5182$\pm$0.0763 & \textbf{0.5519$\pm$0.0778} & \underline{0.5412$\pm$0.0657} \\
   & Fine-tuning & 0.6663$\pm$0.0163 & \underline{0.7080$\pm$0.0213} & \textbf{0.7305$\pm$0.0256} \\
\bottomrule
\end{tabular}}}
\caption{\textbf{Slide-level biomarker prediction comparison between linear probing and fine-tuning.} AUC is reported as the mean ± standard deviation across five random seeds. \textbf{Bold} indicates the best-performing model, and \underline{underline} indicates the second-best-performing model for each setting.}
\label{tab:lp_vs_ft_biomarker_results_setting_rows}
\end{table*}


\end{document}